%% file: main.tex
\documentclass[nohyperref]{article}

\usepackage{microtype}
\usepackage{graphicx}
\usepackage{subfigure}
\usepackage{booktabs} 
\usepackage{float}

\usepackage{hyperref}

\usepackage[accepted]{icml2022}

\usepackage{amsmath}
\usepackage{amssymb}
\usepackage{mathtools}
\usepackage{amsthm}
\usepackage{eucal}
\usepackage[capitalize,noabbrev]{cleveref}

\theoremstyle{plain}

\theoremstyle{definition}

\theoremstyle{remark}

\usepackage{amsmath}

\newcommand{\cA}{\mathcal{A}}
\newcommand{\cD}{\mathcal{D}}

\newcommand{\cL}{\mathcal{L}}

\newcommand{\cP}{\mathcal{P}}
\newcommand{\cR}{\mathcal{R}}

\newcommand{\cX}{\mathcal{X}}

\newcommand{\rE}{\mathbb{E}}

\newcommand{\rR}{\mathbb{R}}
\usepackage{tikz}
\usepackage{scalerel}
\usepackage{xcolor}

\definecolor{dense}{HTML}{648FFF}
\definecolor{prune}{HTML}{DC267F}
\definecolor{rigl}{HTML}{362682}
\definecolor{set}{HTML}{FE6100}
\definecolor{static}{HTML}{FFB000}
\definecolor{last}{HTML}{134E0C}

\newcommand{\riglshp}[1][rigl,fill=rigl]{\scalerel*{\tikz \draw[#1] (0,0) circle (2pt);}{\circ}}

\newcommand{\pruneshp}[1][prune,fill=prune]{\scalerel*{\tikz \draw[rounded corners=0.06pt,#1] (-3pt,0)--++(45:3pt)--++(-45:3pt)--++(-180+45:3pt)--cycle;}{\diamond}}

\newcommand{\staticshp}[1][static,fill=static]{\scalerel*{\tikz \draw[rounded corners=0.1pt,#1] (0,-2.5pt)--++(0,5pt)--++(-30:5pt)--cycle;}{\triangleright}}

\newcommand{\setshp}[1][set,fill=set]{\scalerel*{\tikz \draw[rounded corners=0.1pt,#1] (0,-2.5pt)--++(0,5pt)--++(-180+30:5pt)--cycle;}{\triangleleft}}

\newcommand{\denseshp}[1][dense,fill=dense]{\tikz \draw[rounded corners=0.06pt,#1] (-1.5pt,0)--++(0:3pt)--++(90:3pt)--++(180:3pt)--cycle;}

\usepackage[acronym]{glossaries}
\newacronym{drl}{DRL}{Deep Reinforcement Learning}
\newacronym{rigl}{RigL}{Rigged Lottery}
\newacronym{set}{SET}{Sparse Evolutionary Training}
\newacronym{erk}{ERK}{Erdos Renyi Kernel}
\newacronym{snr}{SNR}{signal-to-noise ratio}
\newacronym{dst}{DST}{Dynamic sparse training}

\usepackage[textsize=tiny]{todonotes}

\icmltitlerunning{Th{\color{white}e} State of Spars{\color{white}e} Train{\color{white}i}ng in D{\color{white}e}ep Re{\color{white}i}nforc{\color{white}e}m{\color{white}e}nt Le{\color{white}a}rn{\color{white}i}ng}

\begin{document}

\twocolumn[
\icmltitle{Th{\color{white}e} State of Spars{\color{white}e} Train{\color{white}i}ng in D{\color{white}e}ep Re{\color{white}i}nforc{\color{white}e}m{\color{white}e}nt Le{\color{white}a}rn{\color{white}i}ng}



\icmlsetsymbol{equal}{*}

\begin{icmlauthorlist}
\icmlauthor{Laura Graesser}{equal,robo-goog,goog}
\icmlauthor{Utku Evci}{equal,goog}
\icmlauthor{Erich Elsen}{adept}
\icmlauthor{Pablo Samuel Castro}{goog}
\end{icmlauthorlist}

\icmlaffiliation{robo-goog}{Robotics at Google}
\icmlaffiliation{goog}{Google Research, Canada}
\icmlaffiliation{adept}{Adept AI}

\icmlcorrespondingauthor{Laura Graesser}{lauragraesser@google.com}
\icmlcorrespondingauthor{Utku Evci}{evcu@google.com}
\icmlcorrespondingauthor{Pablo Samuel Castro}{psc@google.com}

\icmlkeywords{Machine Learning, ICML}

\vskip 0.3in
]



\printAffiliationsAndNotice{\icmlEqualContribution} 

\begin{abstract}
The use of sparse neural networks has seen rapid growth in recent years, particularly in computer vision. Their appeal stems largely from the reduced number of parameters required to train and store, as well as in an increase in learning efficiency. Somewhat surprisingly, there have been very few efforts exploring their use in \gls{drl}. In this work we perform a systematic investigation into applying a number of existing sparse training techniques on a variety of \gls{drl} agents and environments. Our results corroborate the findings from sparse training in the computer vision domain -- \textit{sparse networks perform better than dense networks for the same parameter count} -- in the \gls{drl} domain. We provide detailed analyses on how the various components in \gls{drl} are affected by the use of sparse networks and conclude by suggesting promising avenues for improving the effectiveness of sparse training methods, as well as for advancing their use in \gls{drl}\footnote{Code for reproducing our results can be found at \hyperlink{https://github.com/google-research/rigl/tree/master/rigl/rl}{github.com/google-research/rigl/tree/master/rigl/rl}}.
\end{abstract}

\section{Introduction}

Deep neural networks are typically organized as a stack of layers. Each layer consists of multiple neurons, where each neuron is connected to all neurons in the next layer; this is often referred to as a {\em dense} network. Alternatively, each neuron can be wired to a subset of the neurons in the next layer, resulting in a sparse, and smaller, network. Such sparse neural networks have been shown to match the performance of their dense counterparts while requiring only 10\%-to-20\% of the connections in most cases \citep{han2015learning,gale2019state,Blalock2020stateofpruning} providing significant memory, storage, and latency gains.

Deep networks have become a mainstay of scalable reinforcement learning (RL), key to recent successes such as playing -- at superhuman levels -- Atari games \cite{mnih15human}, Go \cite{silver16mastering}, Dota 2 \cite{dota2} and as well as controlling complex dynamical systems such as stratospheric balloons \cite{bellemare20autonomous} and plasma in real-time \citep{plasmaRL2022}.
Despite their importance, most deep reinforcement learning (DRL) research focuses on improving the {\em algorithmic} aspect of \gls{drl}, and less on the {\em architecture} aspect. Sparse networks in particular have received very little attention, likely due to the belief that network over-parameterization helps with learning. However, recent work suggests that RL agents may suffer from {\em implicit under-parameterization} when training deep networks with gradient descent \cite{kumar2021implicit}, suggesting that the network's expressivity is in fact underused. In addition to this, \citet{nikishin22primacy} suggests deep RL agents may have a tendency to overfit to early training data. Given this, one might expect there is substantial opportunity to compress RL agents. Further, sparse networks might benefit \gls{drl} by reducing the cost of training or aid running them in latency-constrained settings such as controlling plasma \citep{plasmaRL2022}. 

One limitation of current research on training sparse neural networks is that it almost solely focuses on image classification benchmarks \citep{Blalock2020stateofpruning,Hoefler2021SparsityID} creating the risk of over-fitting to a specific domain. Do advances observed in computer vision (CV) transfer to \gls{drl}? A few recent works \citep{Sokar2021DynamicST,Arnob2021SingleShotPF} attempt to address this by applying individual sparse training algorithms to \gls{drl} agents. However, it is still unknown if the key observation made in CV, that \textit{sparse models perform better than dense ones for the same parameter count}, transfers to \gls{drl}. 

In this work we focus on answering that question and systematically explore the effectiveness of different sparse learning algorithms in the online \gls{drl} setting. In order to achieve this, we benchmark four different sparse training algorithms using value-based (DQN~\citep{mnih15human}) and actor-critic, (SAC~\citep{haarnoja18sac} and PPO~\citep{schulman17ppo}) agents.
Our results also include a broad analysis of various components that play a role in the training of these sparse networks: sparsity distribution strategies, weight decay, layer initialization, signal-to-noise ratio for gradients, as well as batch size, topology update strategy and frequency. We summarize our main findings below:

\begin{itemize}
    \item In almost all cases, sparse neural networks perform better than their dense counterparts for a given parameter count, demonstrating their potential for \gls{drl}.
    \item It is possible to train up to 80 - 90\% sparse networks with minimal loss in performance compared to the standard dense networks.
    \item Pruning often obtains the best results, and dynamic sparse training improves over static sparse training significantly. However gradient based growth \citep{evci2019rigl} seems to have a limited effect on performance. We argue this is due to low signal-to-noise ratio in gradients.
    \item The distribution of parameters among the actor and critic networks, as well as among different layers, impact training greatly. We observe that the best performance is obtained by allocating the majority of parameters to the critic network and using \gls{erk} sparsity distributions.
    \item We observe robust performance over various hyper-parameter variations. Somewhat surprisingly, when adding noise to the observations, sparse methods achieve better robustness in most cases.
\end{itemize}

\section{Background}

\subsection{Sparse training}
Removing connections from neural networks was suggested at least as early as \citet{mozer1989skeletonization}, which coined the name ``Skeleton'' networks for what we today call sparse networks. Techniques for finding sparse neural networks can be grouped under two main categories.\\{\bf(1)} \pruneshp{} \textbf{Dense-to-sparse training} approaches \citep{Han2016,variational-dropout,Wortsman2019neuralwirings,Kusupati2020str,Peste2021ACDCAC} start with a dense neural network and gradually reduce the network size by pruning its weights. This approach often achieves state-of-the-art performance amongst sparse networks, however it requires the same (or more) computation as training a large dense network. An alternative to pruning is {\bf(2)} \staticshp~\setshp~\riglshp~ \textbf{sparse training} \citep{Mocanu2018}. This family of methods sparsifies the network at initialization and maintains this sparsity throughout training, thus reducing the training cost proportional to the sparsity of the network. However, training sparse neural networks from scratch is known to be difficult, leading to sub-optimal solutions \citep{Frankle2018lottery,Liu2018,Evci2019difficulty}. 

DRL training is notoriously resource hungry, hence we focus on the second family of methods (i.e. sparse training) in this work. There are various approaches to sparse training. One line of work \citep{snip,grasp,tanaka2020pruning}, attempts to prune a dense network \emph{immediately} on iteration 0. The resulting networks are used as an initialization for sparse training and kept fixed throughout. These techniques have been shown to have marginal gains over random pruning \citep{Frankle2020pruningatinit}, especially when used in modern training pipelines. Furthermore they may not generalize well in the RL setting as the non-stationarity of the data make it less clear that any decision made at iteration 0 will remain optimal throughout training.

Another line of work starts with randomly initialized sparse neural networks (both weights and masks) and focuses on improving sparse training by changing the sparse connectivity among neurons \citep{Mocanu2018,Bellec2017} throughout the optimization. Known as \gls{dst}, such approaches have been shown to match pruning results, making it possible to train sparse networks efficiently without sacrificing performance \citep{dettmers2019,evci2019rigl}. 

In this work we benchmark one dense-to-sparse and three sparse training methods, which we briefly describe below:

{\pruneshp{} \bf Pruning \cite{gupta2018}:} uses a simple procedure to slowly make a dense network sparse over the course of one training run using weight magnitudes. We start pruning the network from 20\% of the training steps and stop when we reach 80\%, keeping the final sparse network fixed for the remaining of the training. This simple pruning algorithm is shown to exceed or match more complex pruning algorithms \citep{gale2019state}. Despite the fact it requires the same order of magnitude resources as training a dense network, we included this method since it serves as an upper bound on the sparse training performance. 

{\staticshp{} \bf Static:} prunes a given dense network randomly at initialization and the resulting sparse network is trained with a fixed structure. This is an important baseline to show the effectiveness of \gls{dst} algorithms explained below. 

{\setshp{} \bf \gls{set} \citep{Mocanu2018}:} Similar to \textit{Static}, \gls{set} starts training with a random sparse network. During training, a portion of the connections are changed every N steps (the \textit{update interval}) by replacing the lowest magnitude connections with new random ones. The fraction (\textit{drop fraction}) of updated weights are decayed over the course of training to help the network converge to a minima. We use cosine decay as proposed by \citet{dettmers2019}.

{\riglshp{} \bf \gls{rigl} \citep{evci2019rigl}:} is the same as \gls{set}, except the new connections are activated using the gradient signal (highest magnitude) instead of at random. This criteria has been shown to improve results significantly in image classification and with enough training iterations matches or exceed accuracies obtained by pruning.

\subsection{Reinforcement learning}
Reinforcement learning (RL) aims to design learning algorithms for solving sequential decision-making problems. Typically these are framed as an {\em agent} interacting with an {\em environment} at discrete time-steps by making action choices from a set of possible agent states; the environment in turn responds to the action selection by (possibly) changing the agent's state and/or providing a numerical {\em reward} (or cost); the agent's objective is to find a {\em policy} mapping states to actions so as to maximize (minimize) the sum of rewards (costs). This is formalized as a Markov decision process \cite{puterman94mdp} defined as a tuple $\langle\cX, \cA, \cP, \cR, \gamma\rangle$, where $\cX$ is the state space, $\cA$ is the action space, $\cP:\cX\times\cA\rightarrow\Delta(\cX)$ defines the transition dynamics\footnote{$\Delta(X)$ denotes the set of probability distributions over a finite set $X$.}, $\cR:\cX\times\cA\rightarrow\rR$ is the reward function, and $\gamma\in [0, 1)$ is a discount factor. A policy $\pi:\cX\rightarrow\Delta(\cA)$ formalizes an agent's behaviour and induces a value function $V^{\pi}:\cX\rightarrow\rR$ defined via the well-known Bellman recurrence:
\begin{align}\label{eqn:bellmanRecurrence}
    V^{\pi}(x) & := \rE_{a\sim\pi(x)}\left[ \cR(x, a) + \gamma \rE_{x'\sim\cP(x, a)}V^{\pi}(x') \right]
\end{align}

It is convenient to define state-action value functions $Q^{\pi}:\cX\times\cA\rightarrow\rR$ as: $Q^{\pi}(x, a)  := \cR(x, a) + \gamma \rE_{x'\sim\cP(x, a)}V^{\pi}(x')$.

The goal of an RL agent is to find a policy $\pi^* := \max_{\pi} V^{\pi}$ (which is guaranteed to exist); for notational convenience we denote $V^* := V^{\pi^*}$ and $Q^* := Q^{\pi^*}$. In online RL the agent achieves this by iteratively improving an initial policy $\pi_0$: $\{\pi_0,\pi_1,\cdots,\pi_t,\cdots\}$ and using these intermediate policies to collect new experience from the environment in the form of {\em transitions} $(x, a, r, x')$, where $a\sim\pi_t(x)$, $r=\cR(x, a)$, and $x'\sim\cP(x, a)$. These transitions constitute the {\em dataset} the agent uses to improve its policies. In other words, the learning proocess is a type of {\em closed feedback loop}: an agent's policy directly affects the data gathered from the environment, which in turn directly affects how the agent updates its policy.

When $\cX$ is very large, it is impractical to store $V^\pi$ and $Q^\pi$ in a table, so a function approximator $V_{\theta}\approx V^\pi$ (where $\theta$ are the approximator's parameters) is employed instead. This function approximator is usually one or more deep networks, and this type of RL is known as deep RL (DRL). \gls{drl} algorithms can be broadly categorized into two groups:

{\bf Value-based:} The function $Q^\pi$ is approximated by a deep network $Q_\theta$. The policy is directly induced from the value estimate via $\pi_t(x) = \arg\max_{a\in\cA}Q_{\theta_t}(x, a)$\footnote{Although there are other mechanisms for defining a policy, such as using a softmax, and there are exploration strategies to consider, we present only the argmax setup for simplicity, as the other variants are mostly orthogonal to our analyses.}.
The parameters $\theta$ are trained using a temporal difference loss (based on \autoref{eqn:bellmanRecurrence}) from transitions sampled from $\cD$:
\begin{align}
    \cL(\theta) = \rE_{(x, a, r, x')\sim\cD}\left[ Q_{\theta}(x, a) - (r + \gamma \max_{a'\in\cA}Q_{\bar{\theta}}(x', a')) \right]
\end{align}
Here, $\bar{\theta}$ s a copy of $\theta$ that is infrequently synced with $\theta$ for more stable training \cite{mnih15human}. These methods are typically employed for {\em discrete control} environments, where there is a finite (and relatively small) set of actions (e.g. Atari games \citep{bellemare13arcade}).

{\bf Policy-gradient:} In contrast to value-based methods where the policy is implicitly improved by virtue of improving $Q_{\theta}$, policy-gradient methods maintain and directly improve upon a policy $\pi_{\psi}$ parameterized by $\psi$. These methods typically still make use of a value estimate $Q_{\theta}$ as part of their learning process, and are thus often referred to as actor-critic methods (where $\pi_{\psi}$ is the actor and $Q_{\theta}$ the critic). Two potential advantages of these methods is that they can be more forgiving of errors in the $Q_{\theta}$ estimates, and they can handle continuous action spaces (for instance, by having $\pi_{\psi}(x)$ output mean and variance parameters from which actions may be sampled).
These methods are typically employed for {\em continuous control} environments, where the action space is continuous (e.g. MuJoCo \citep{todorov12mujoco}).

\section{Experimental setup}
\paragraph{\gls{drl} algorithms} We investigate both value-based and policy-gradient methods. We chose DQN \citep{mnih15human} as the value-based algorithm, as it is the algorithm that first spurred the field of \gls{drl}, and has thus been extensively studied and extended. We chose two actor-critic algorithms for our investigations: an {\em on-policy} algorithm (PPO \citep{schulman17ppo}) and an {\em off-policy} one (SAC \citep{haarnoja18sac}); both are generally considered to be state-of-the-art for many domains.

\begin{figure*}[!t]
    \centering
   \includegraphics[width=0.45\textwidth]{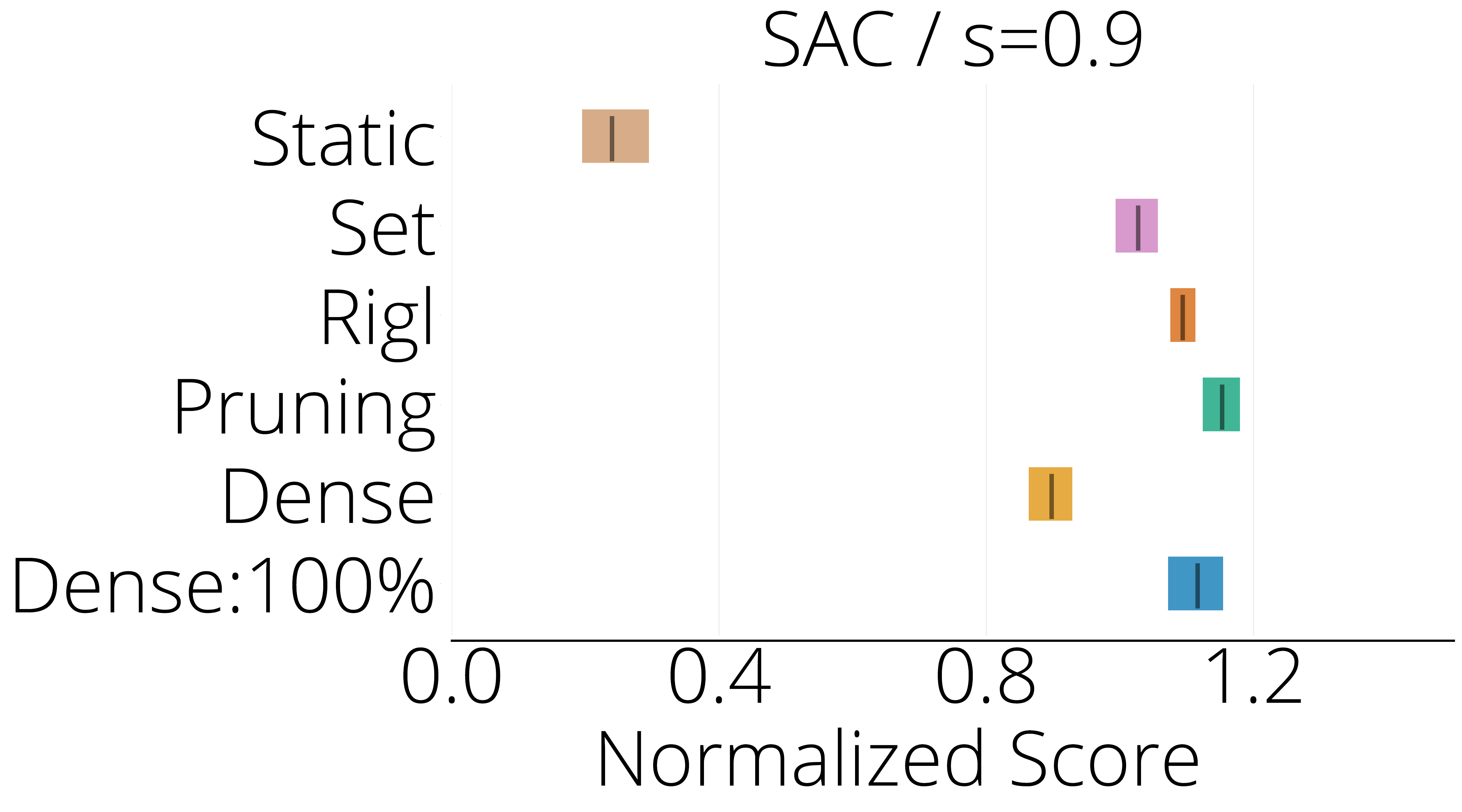}
    \includegraphics[width=0.45\textwidth]{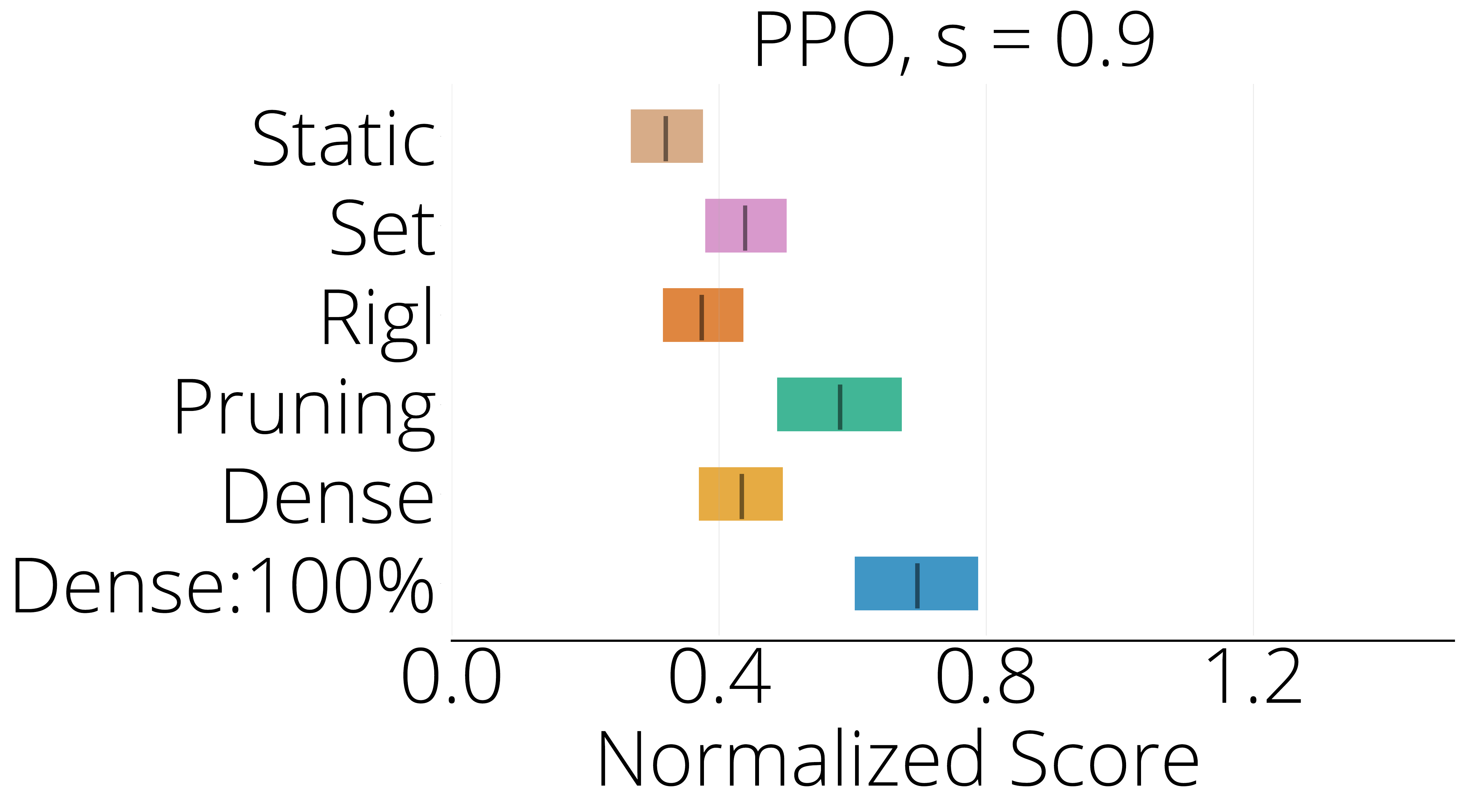}
   \includegraphics[width=0.45\textwidth]{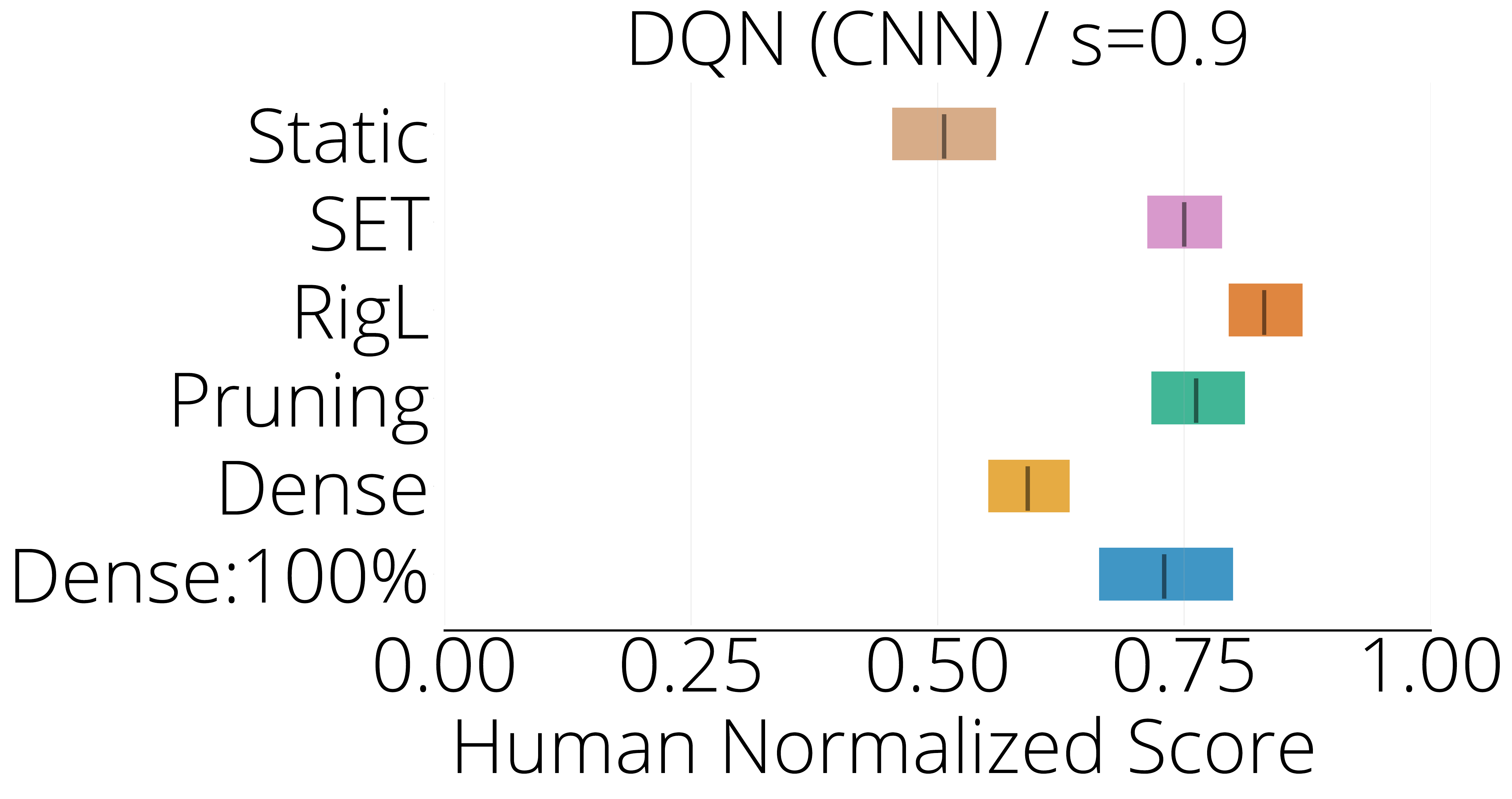}
   \includegraphics[width=0.45\textwidth]{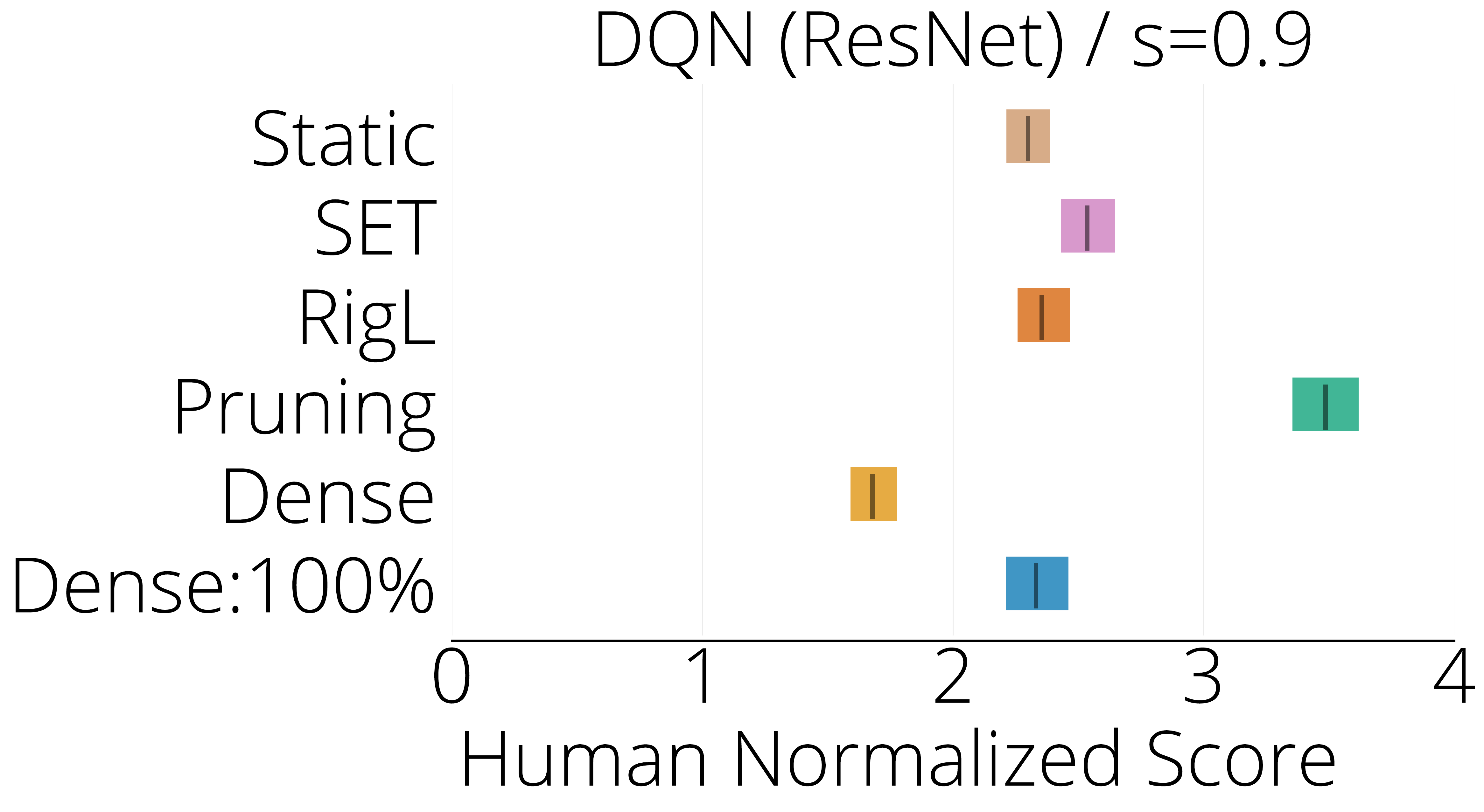}
    \caption{IQM plots for networks at 90\% sparsity for various architecture and algorithm combinations. SAC and PPO are averaged over 5 MuJoCo environments, whereas DQN is averaged over 15 Atari environments. Results at different sparsities can be found at \autoref{sec:iqm_extended}. "Dense: 100\%" corresponds to the standard dense model. Atari scores were normalized using human performance per game. MuJoCo scores were normalized using the average returns obtained by the Dense: 100\% SAC agent per game.}
    \label{fig:highLevelIQM}
\end{figure*}

\begin{figure*}[!t]
    \centering
    \includegraphics[width=0.31\textwidth]{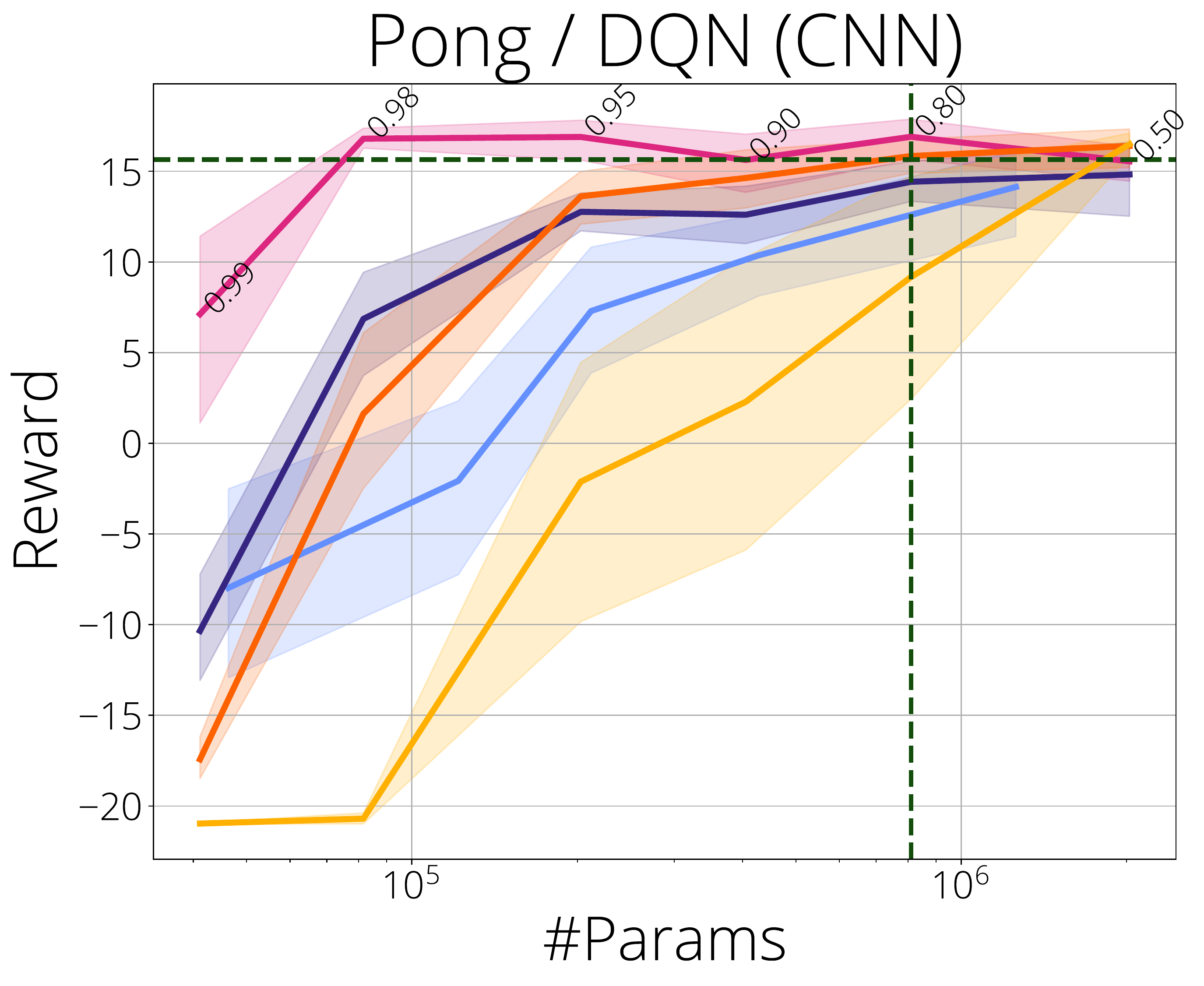}
    \includegraphics[width=0.31\textwidth]{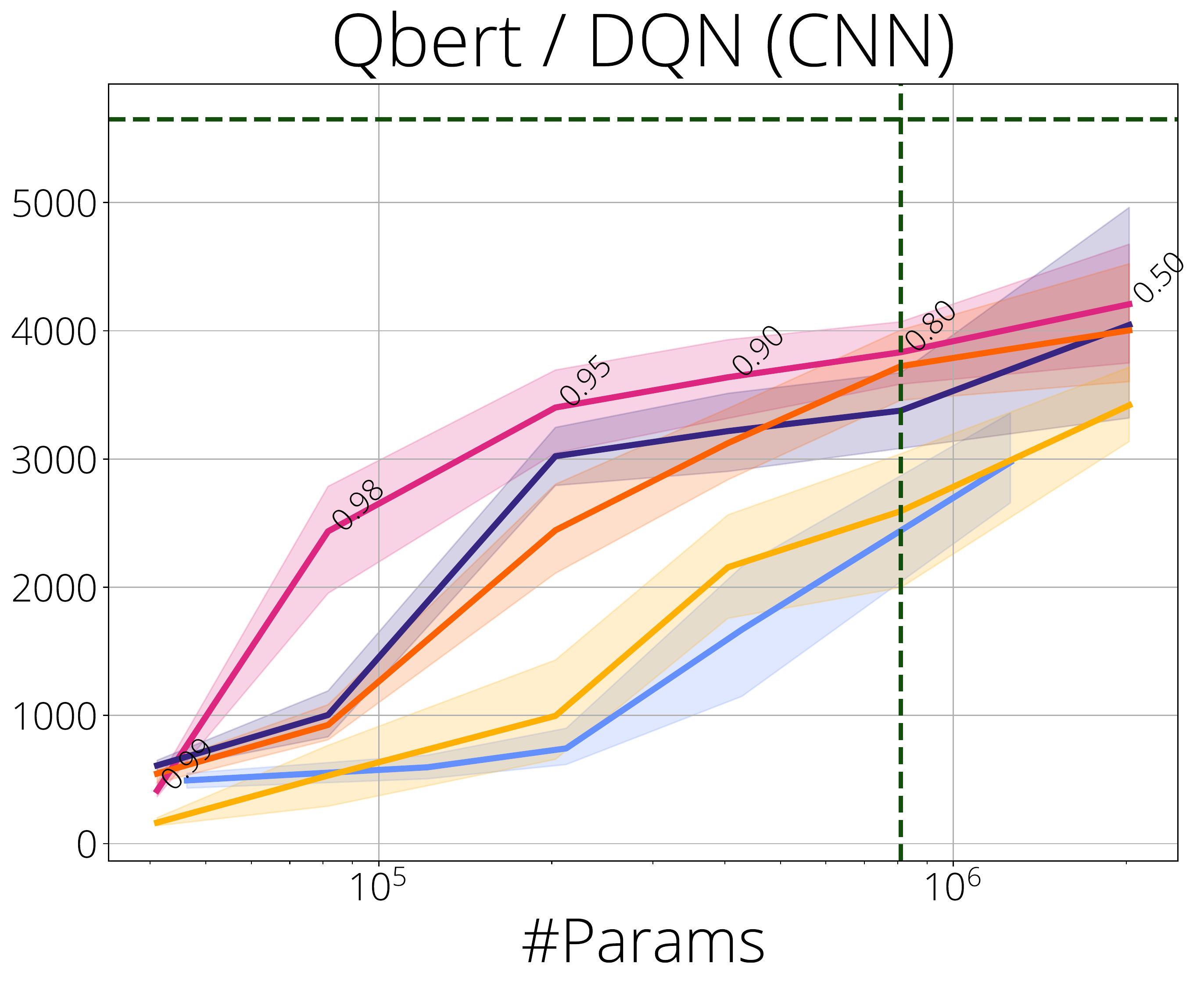}
    \includegraphics[width=0.31\textwidth]{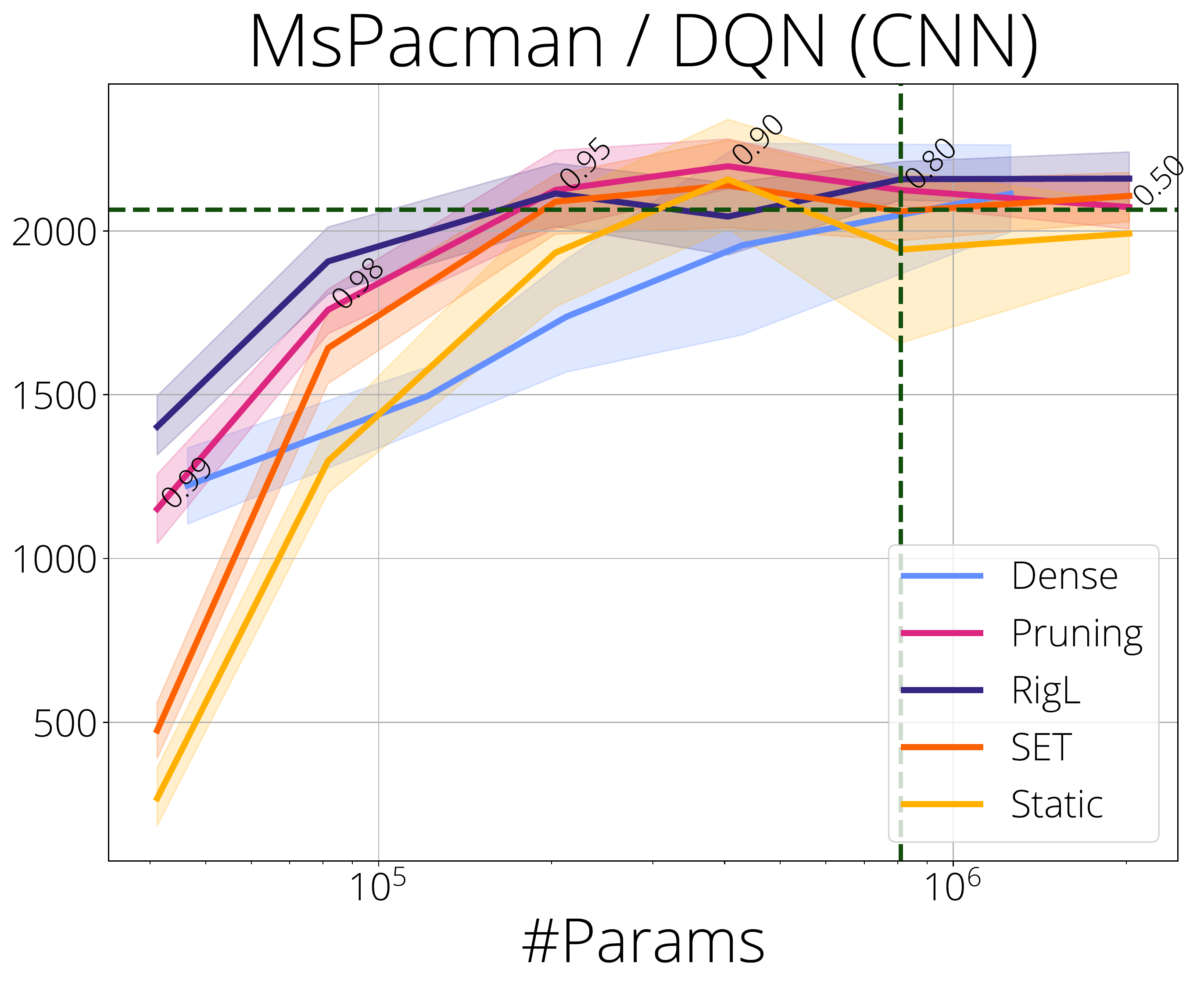}
     \includegraphics[width=0.31\textwidth]{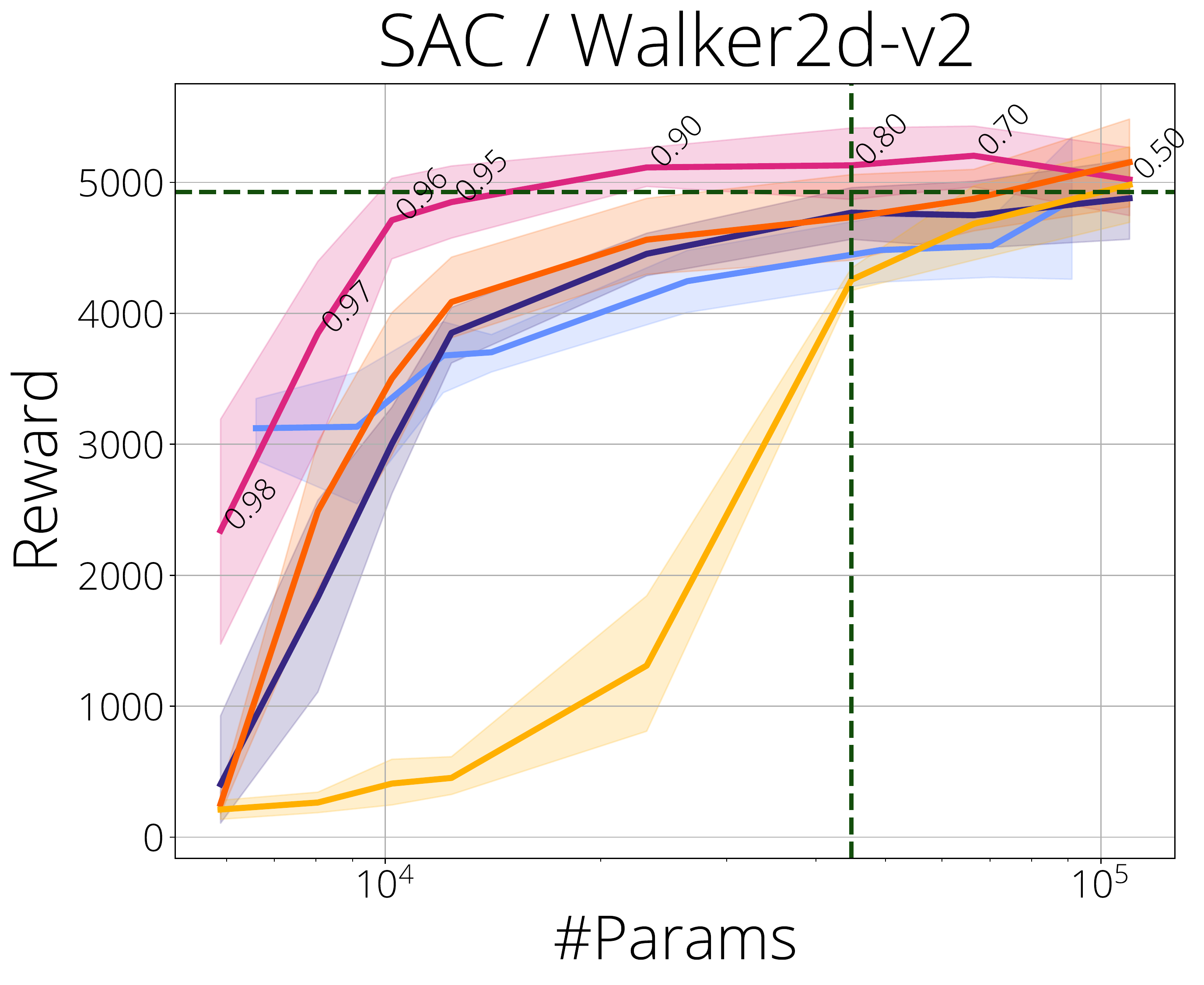}
    \includegraphics[width=0.31\textwidth]{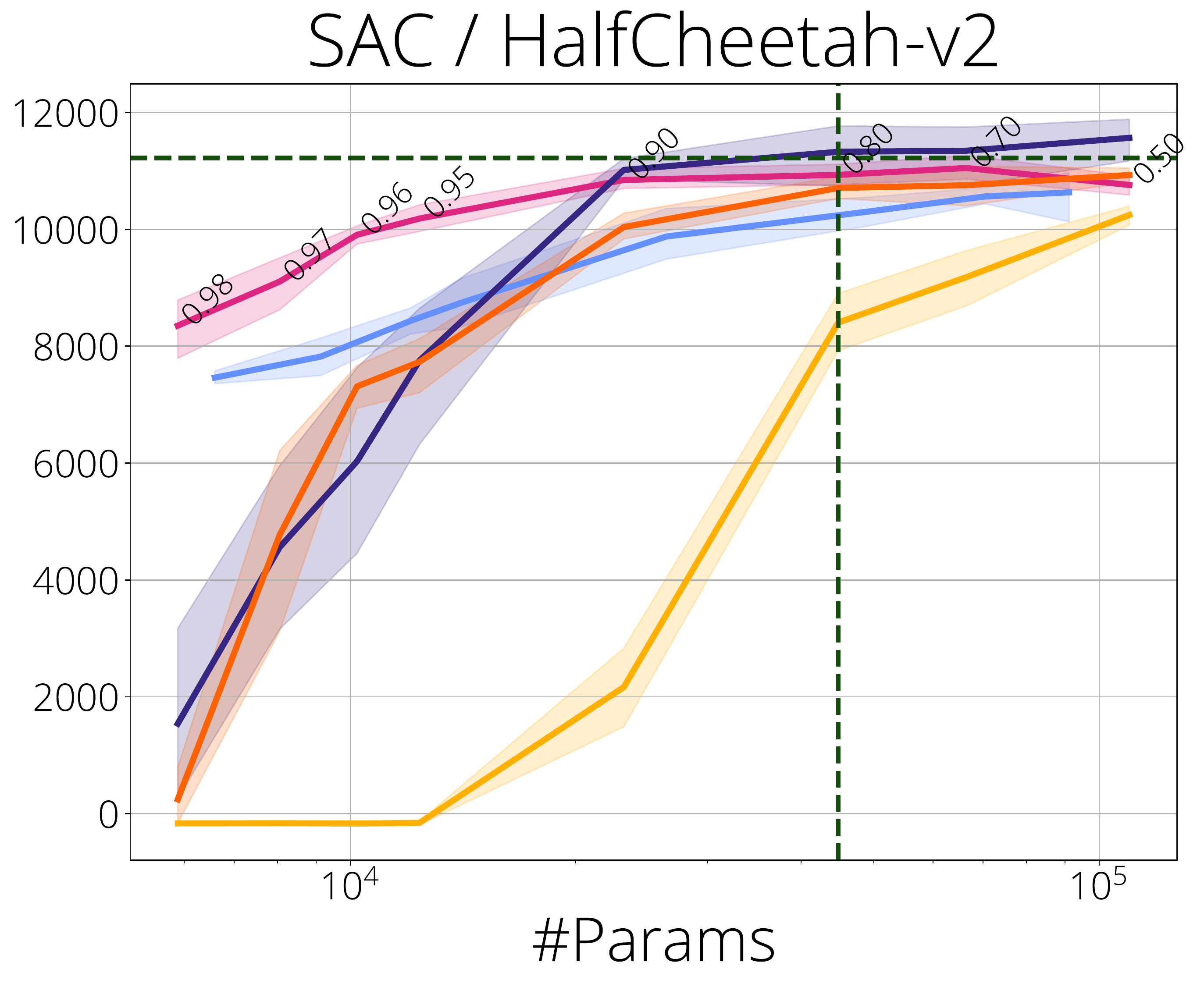}
    \includegraphics[width=0.31\textwidth]{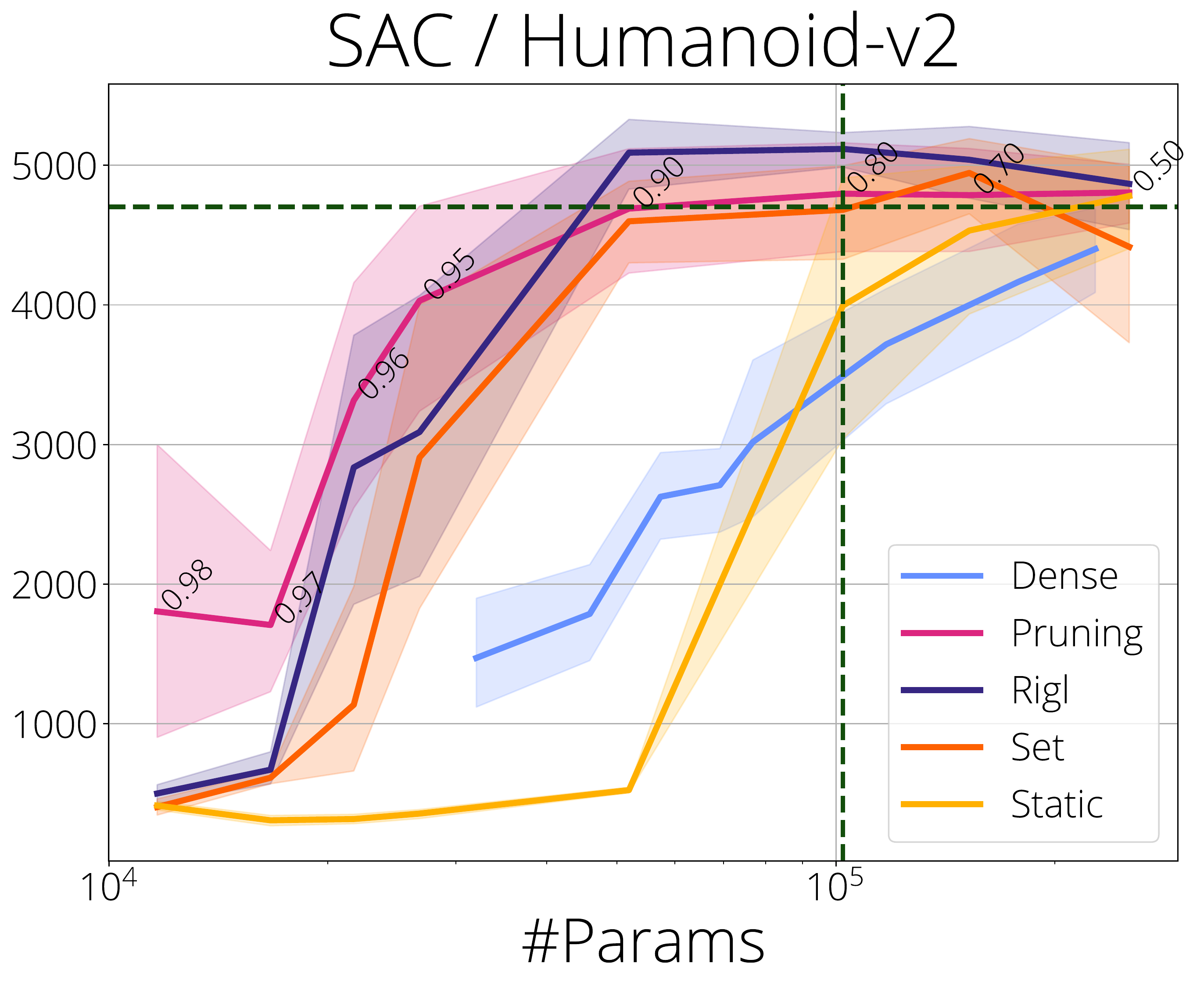}
    \includegraphics[width=0.31\textwidth]{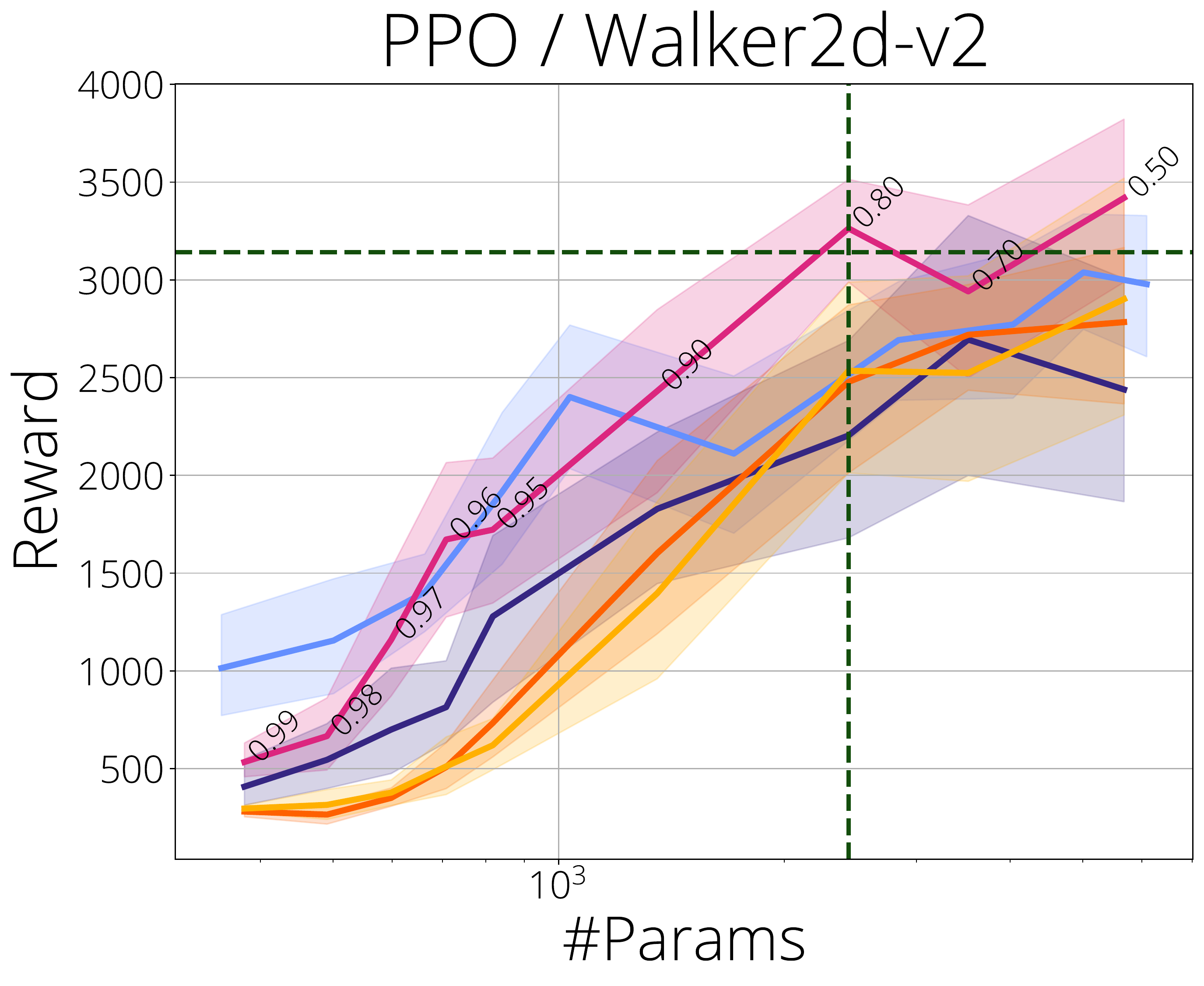}
    \includegraphics[width=0.31\textwidth]{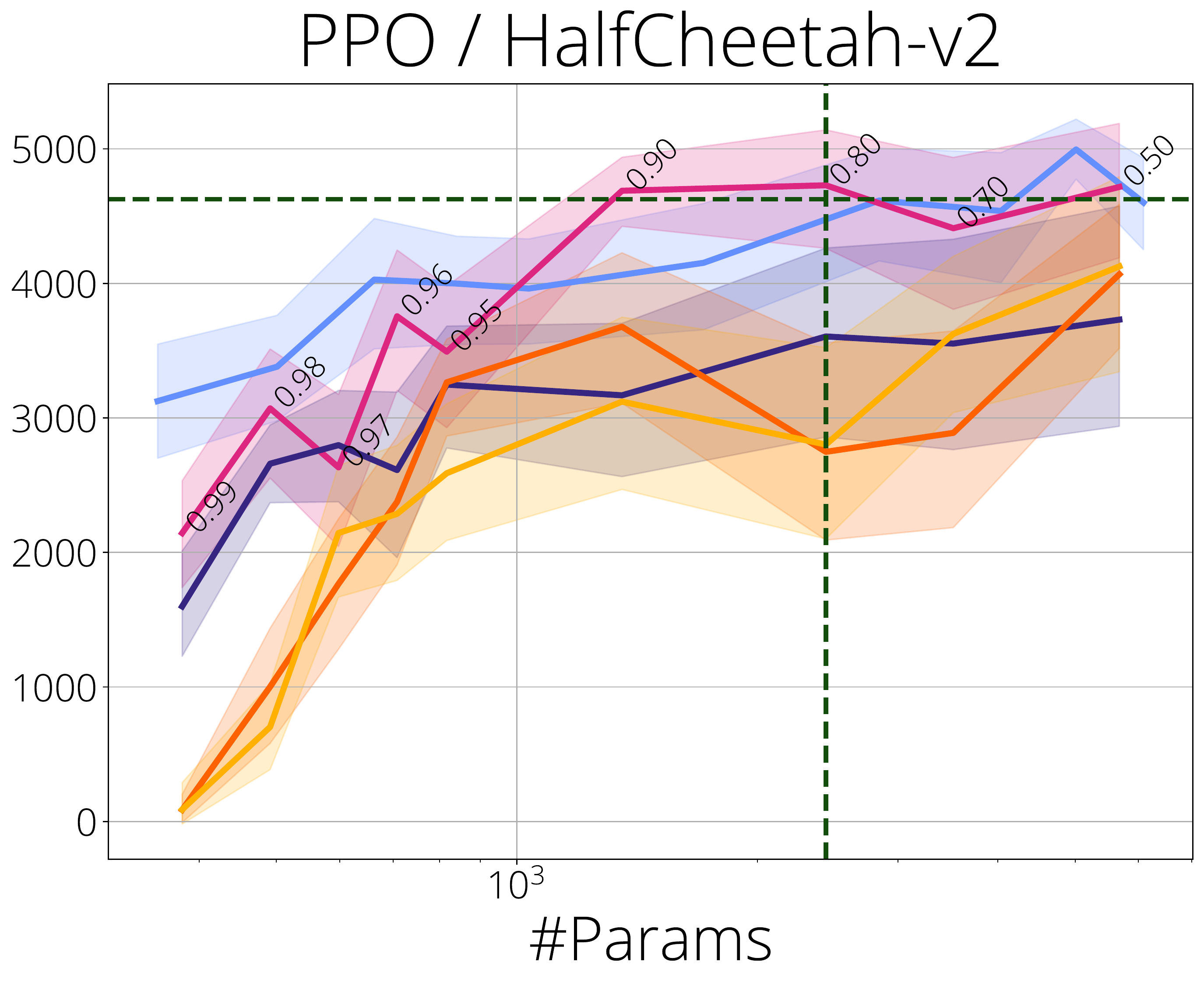}
    \includegraphics[width=0.31\textwidth]{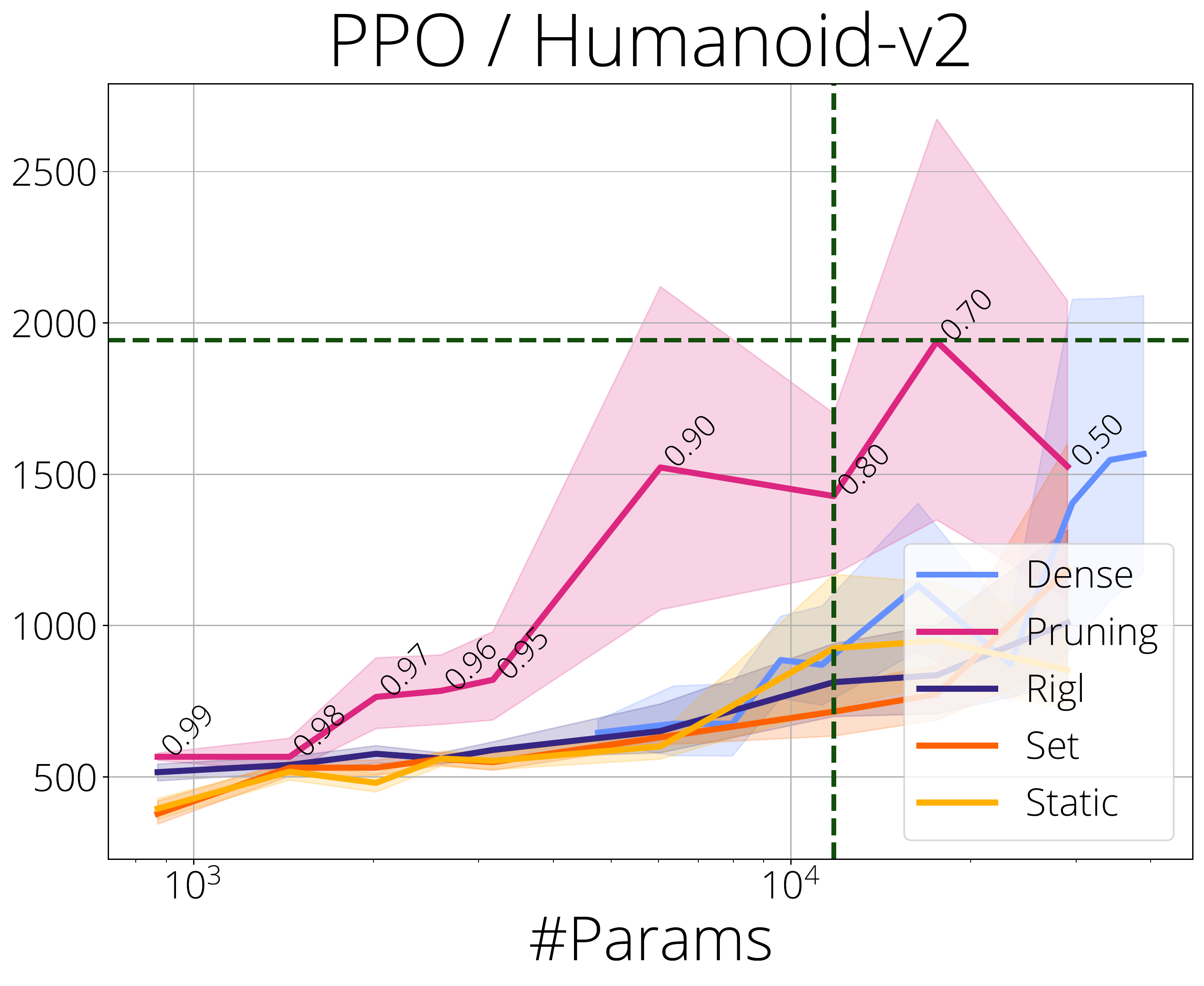}
    
    \caption{Comparison of the final reward relative to parameter count for the various methods considered: (row-1) DQN on Atari (CNN) (row-2) SAC on MuJoCo, (row-3) PPO on MuJoCo. We consider sparsities from 50\% to 95\% (annotated on the pruning curve) for sparse training methods and pruning. Parameter count for networks with 80\% sparsity and the reward obtained by the dense baseline are highlighted with vertical and horizontal lines. Shaded areas represent 95\% confidence intervals. See \autoref{sec:additional_results_erk} for results on additional environments.}
    \label{fig:highLevelComparisonERK}
\end{figure*}

\paragraph{Environments} For discrete-control we focus on three classic control environments (CartPole, Acrobot, and MountainCar) as well as 15 games from the ALE Atari suite \citep{bellemare13arcade} (see \autoref{sec:atari-game-selection} for game selection details). 
For continuous-control we use five environments of varying difficulty from the MuJoCo suite~\citep{todorov12mujoco} (HalfCheetah, Hopper, Walker2d, Ant, and Humanoid). Rewards obtained by \gls{drl} algorithms have notoriously high variance \citep{rliable}. Therefore we repeat each experiment with at least 10 different seeds and report the average reward obtained over the last 10\% of evaluations. We also provide 95\% confidence intervals in all plots. See \autoref{app:exp-details} for additional details.

\vspace{-0.5em}
\paragraph{Training}
For each sparse training algorithm considered (\pruneshp{} Pruning , \staticshp{} Static, \riglshp{} \gls{rigl}, and \setshp{} \gls{set}) we train policies ranging between 50\% to 99\% sparsity. To ensure a fair comparison between algorithms, we performed a hyper parameter sweep for each algorithm separately. The exception is DQN experiments on Atari for which it was too computationally expensive to do a full hyper-parameter sweep and we used values found in previous experiments instead. Sparse results in these environments may therefore be conservative compared to the well tuned dense baseline.

In addition to training the standard dense networks used in the literature, we also train smaller dense networks by scaling down layer widths to approximately match the parameter counts of the sparse networks, thereby providing a "parameter-equivalent" dense baseline. We share details of the hyper parameter sweeps and hyper parameters used for each algorithm in \autoref{app:hyper-details}.

\vspace{-0.5em}
\paragraph{Code} Our code is built upon the TF-Agents \citep{tfagents}, Dopamine \citep{dopamine}, and RigL \citep{evci2019rigl} codebases. We use rliable \citep{rliable} to calculate the interquartile mean (IQM) and plot the results. The IQM is calculated by discarding the bottom and top 25\% of normalized scores aggregated from multiple runs and environments, then calculating the mean (reported with 95\% confidence intervals) over the remaining 50\% runs. Code for reproducing our results can be found at \hyperlink{https://github.com/google-research/rigl/tree/master/rigl/rl}{github.com/google-research/rigl/tree/master/rigl/rl}

\begin{figure*}[!t]
    \centering
    \includegraphics[width=0.31\textwidth]{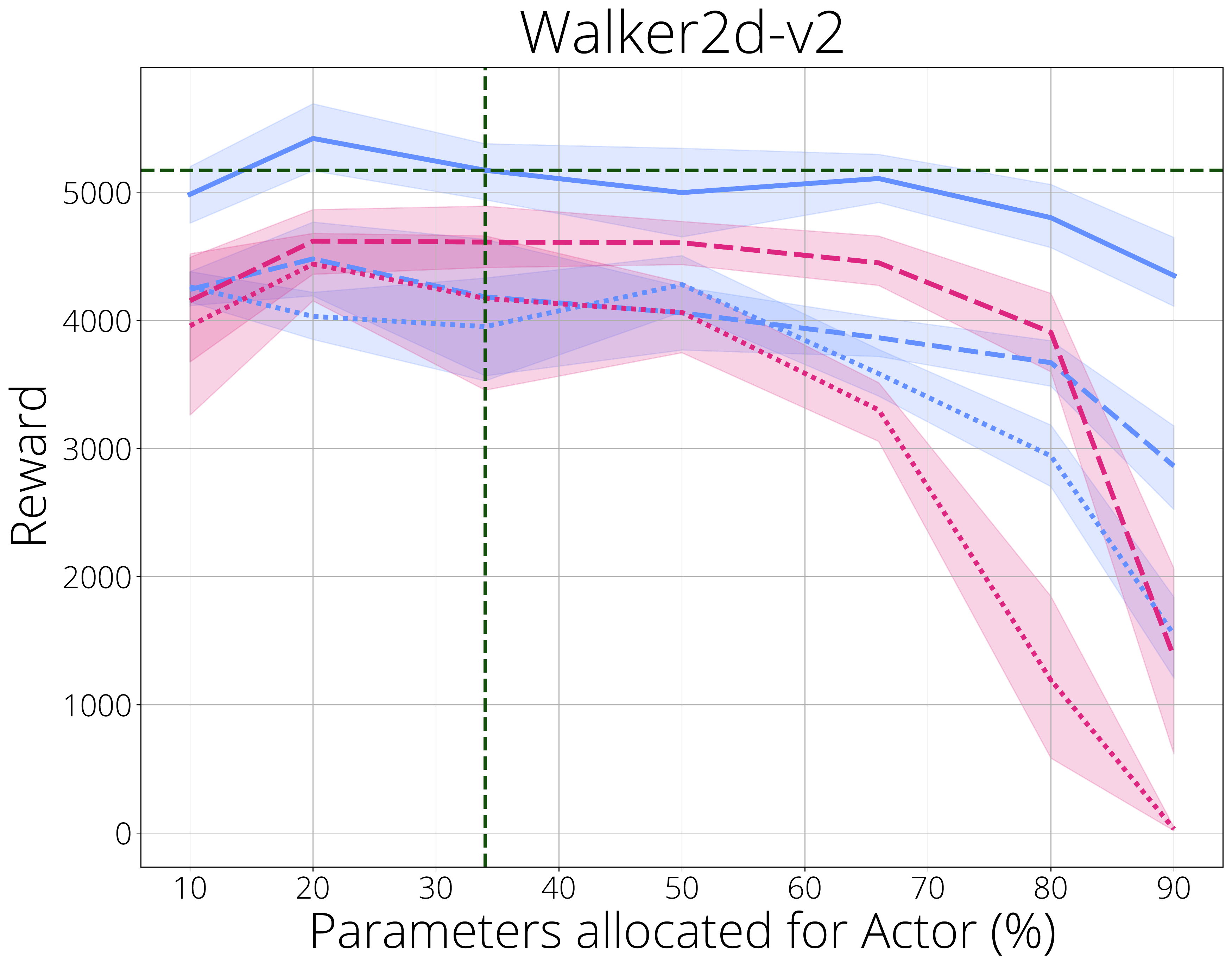}
    \includegraphics[width=0.31\textwidth]{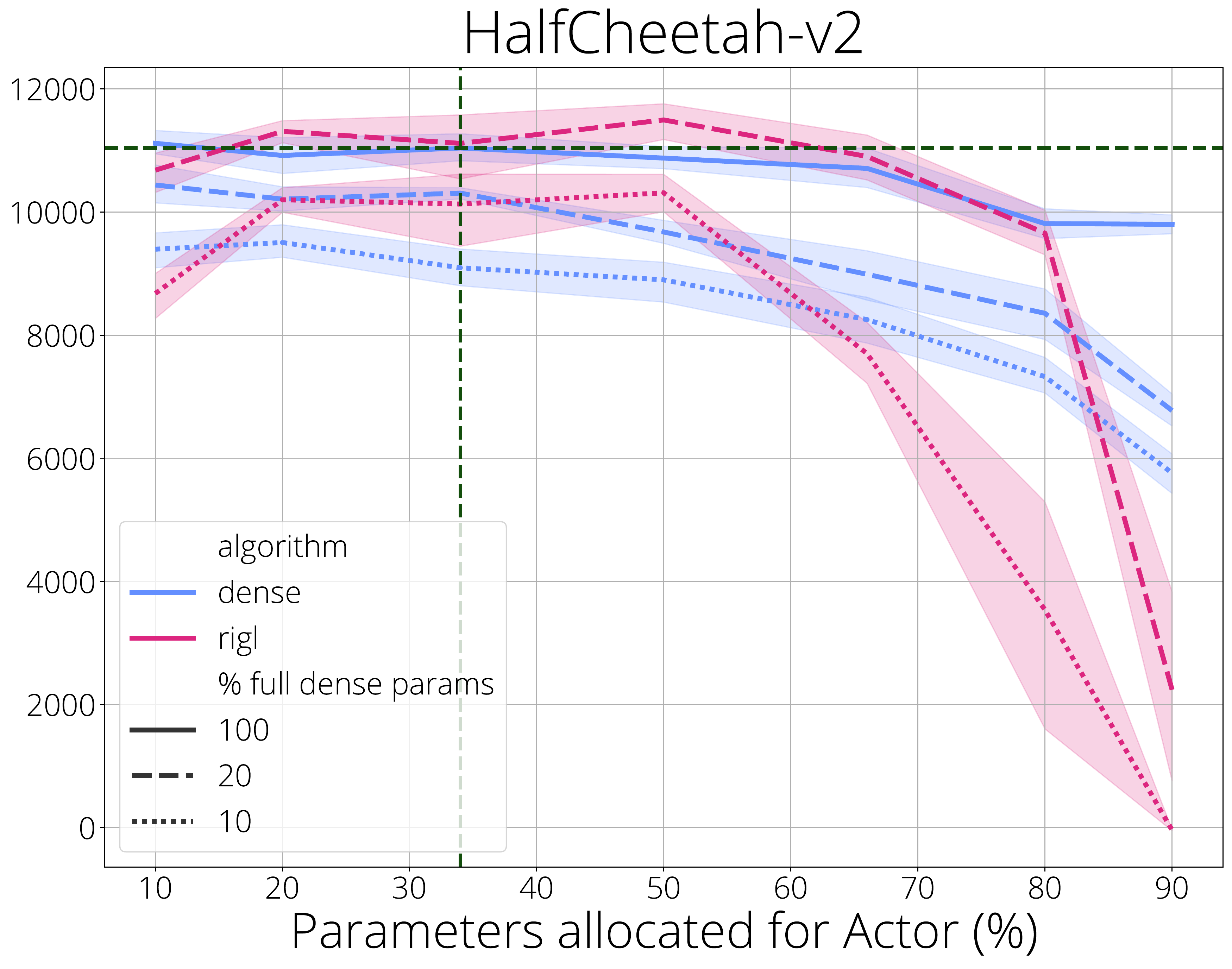}
    \includegraphics[width=0.31\textwidth]{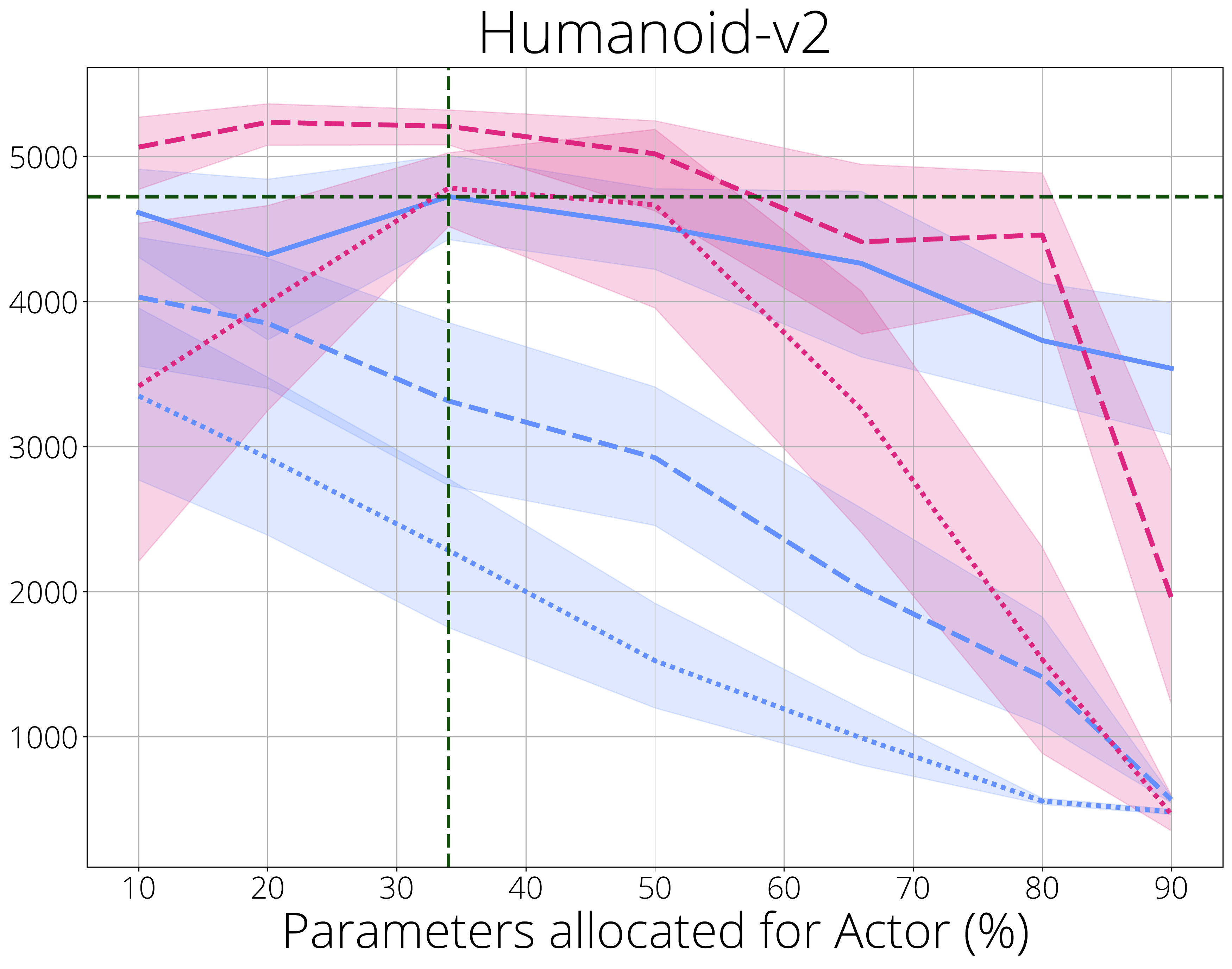}
    \caption{Evaluating how varying the actor-critic parameter ratio affects performance for a given parameter budget on different environment with policies trained using SAC. \% of parameters allocated to the actor network is reported on the x axis. In SAC the parameter count of both critic networks is summed to give the overall critic parameter count. The vertical line corresponds the the standard parameter split and the horizontal line to the full dense training reward.}
    \label{fig:betweenNetDist}
\end{figure*}

\section{The State of Sparse Networks in Deep RL}
We begin by presenting the outcome of our analyses in \autoref{fig:highLevelIQM} and
\autoref{fig:highLevelComparisonERK} for DQN (Atari), PPO and SAC (MuJoCo). \autoref{fig:highLevelIQM} presents the IQM at 90\% sparsity, whilst in \autoref{fig:highLevelComparisonERK} we evaluate final performance relative to the number of parameters. We share results for classic control, 2 additional MuJoCo and 12 additional Atari environments in \autoref{sec:additional_results_erk}. Three main conclusions emerge; {\bf(1)} In most cases performance obtained by sparse networks significantly exceeds that of their dense counterparts with a comparable number of parameters. Critically, in more difficult environments requiring larger networks (e.g. Humanoid, Atari), sparse networks can be obtained with efficient sparse training methods. {\bf(2)} It is possible to train sparse networks with up to 80-90\% fewer parameters and without loss in performance compared to the standard dense model. {\bf(3)} Gradient based growing (i.e. \gls{rigl}) seems to have limited impact on the performance of sparse networks. Next, we discuss each of these points in detail.

{\bf Sparse networks perform better.} Inline with previous observations made in speech \citep{kalchbrenner2018}, natural language modelling \citep{Li2020TrainLT} and computer vision \citep{evci2019rigl}, in almost all environments, sparse networks found by pruning achieve significantly higher rewards than the dense baseline. However training these sparse networks from scratch (\textit{static}) performs poorly. \gls{dst} algorithms (\gls{rigl} and \gls{set}) improve over \textit{static} significantly, however often fall short of matching the pruning performance.

Critically, we observe that for more difficult environment requiring larger networks such as Humanoid, MsPacman, Qbert and Pong, sparse networks found by efficient \gls{dst} algorithms exceed the performance of the dense baseline.

{\bf How sparse?} Next we asked how much sparsity is possible without loss in performance relative to that of the standard dense model (denoted by Dense:100\% in \autoref{fig:highLevelIQM} and by the horizontal lines in \autoref{fig:highLevelComparisonERK}). We find that on average \gls{dst} algorithms maintain performance up to 90\% sparsity using SAC (\autoref{fig:highLevelIQM} top left) or DQN ( \autoref{fig:highLevelIQM} bottom row), after which performance drops.

However performance is variable. For example, \gls{dst} algorithms maintain performance especially well in MsPacman and Humanoid. Whereas in Qbert none of the methods are able to match the performance of the standard dense model at any of the examined levels of sparsity.

In the Atari environments, training a ResNet \citep{resnetHe} following the architecture from \citet{impala} instead of the standard CNN alone provided about 3x improvement in IQM scores. We were also surprised to see that pruning at 90\% sparsity exceeds the performance of the standard ResNet model.

These observations indicate that while sparse training can bring very significant efficiency gains in some environments, it is not a guaranteed benefit. Unlike supervised learning, expected gains likely depend on both task and network, and merits further inquiry.

{\bf \gls{rigl} and \gls{set}:} For most sparsities (50\% - 95\%) we observe little difference between these two sparse training algorithms. At very high sparsities, \gls{rigl} may outperform \gls{set}. The difference can be large (e.g. MsPacman), but is more often moderate (e.g. Pong) or negligible (e.g. Humanoid, Qbert) with overlapping confidence intervals. This suggests that the gradient signal used by \gls{rigl} may be less informative in the \gls{drl} setting compared to image classification, where it obtains state-of-the-art performance and consistently outperforms \gls{set}. Understanding this phenomenon could be a promising direction for improving sparse training methods for \gls{drl}.

Perhaps unsurprisingly, the clarity of the differences between sparse and dense training is affected by the stability of the underlying RL algorithm. Our results using SAC, designed for stability, were the clearest, as were the DQN results. In contrast, our results using PPO which has much higher variance, were less stark. For this reason, we used SAC and DQN when studying the different aspects of sparse agents in the rest of this work.

\begin{figure*}[!t]
    \centering
    \includegraphics[width=0.3\textwidth]{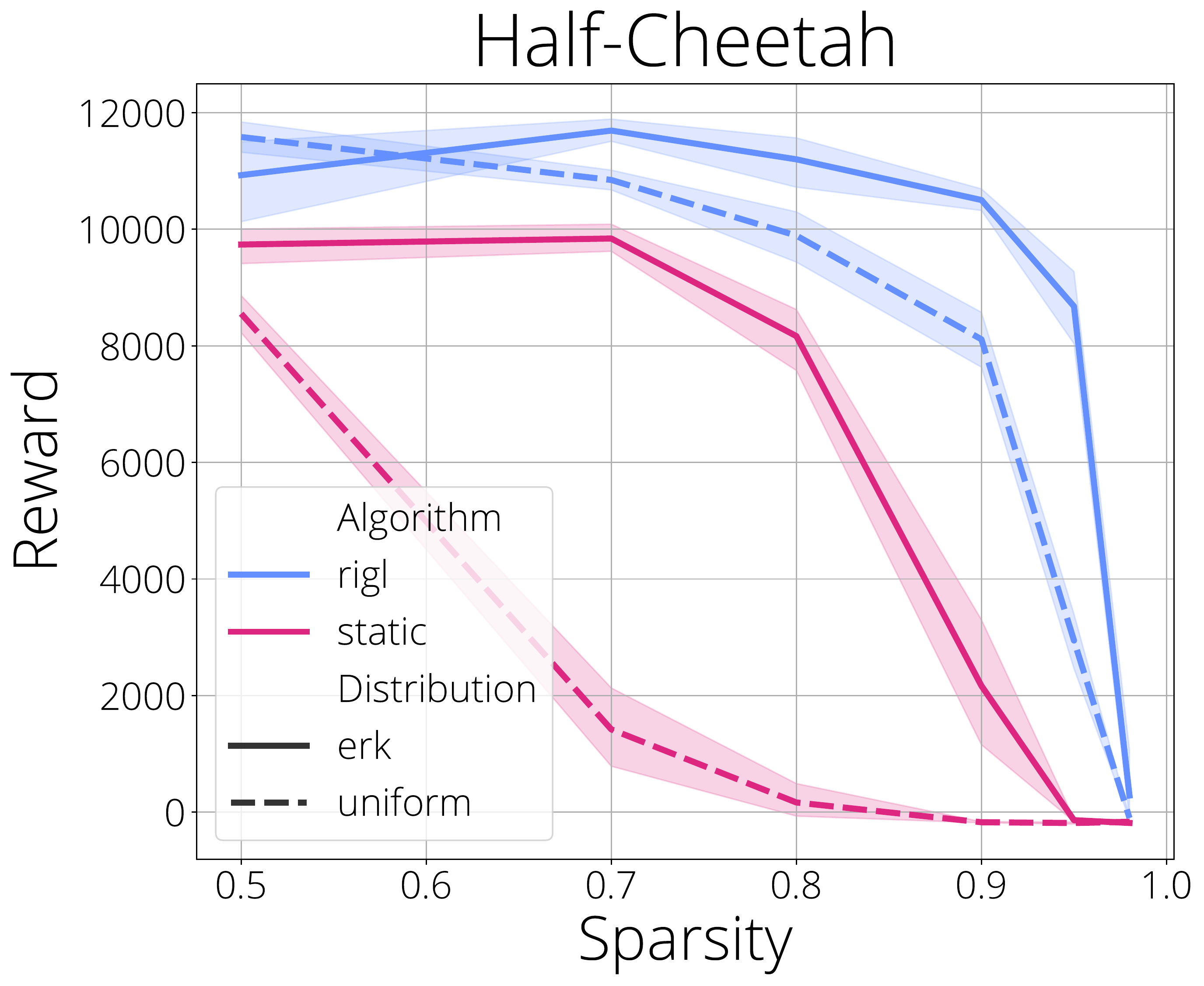}
    \includegraphics[width=0.3\textwidth]{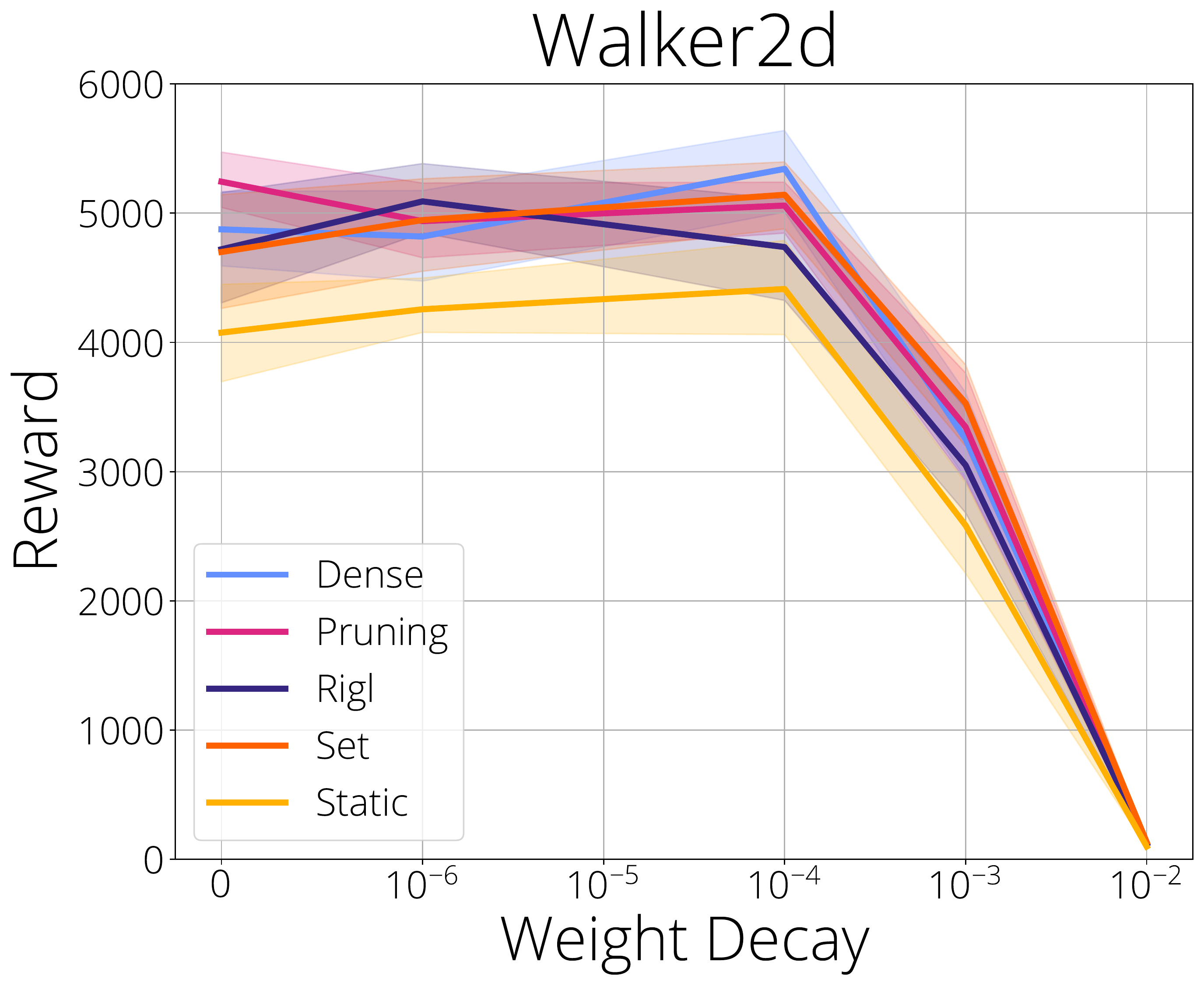}
    \includegraphics[width=0.3\textwidth]{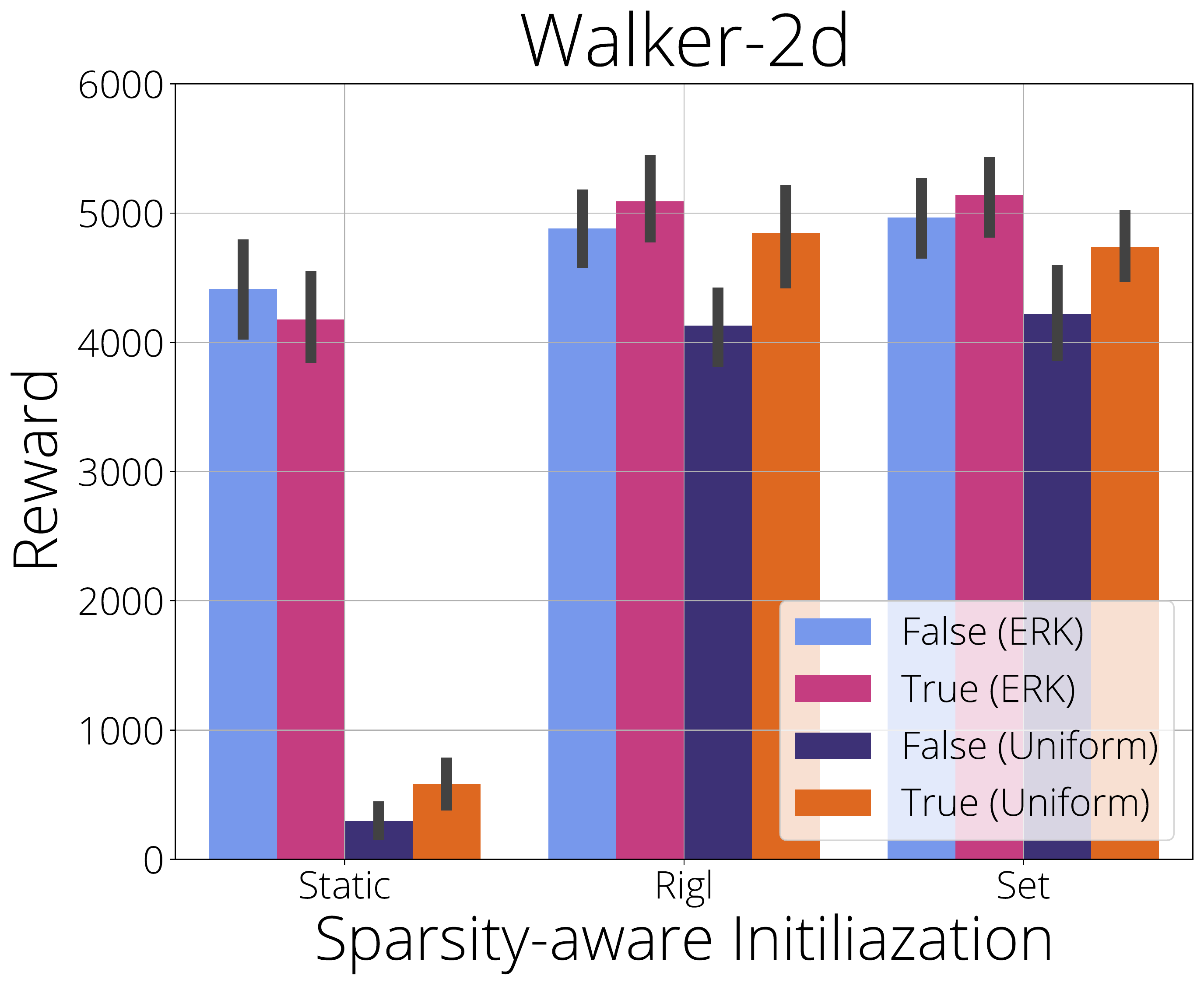}
    \caption{Sensitivity analysis on policies trained with SAC: (left) uniform vs \gls{erk} sparsity distributions, (center) weight decay, (right) sparsity-aware vs dense initialization. We use 80\% (ERK) sparse networks in all plots unless noted otherwise. Plots for the remaining hyper-parameters are shared in \autoref{app:additional_hypers}.}
    \label{fig:hyper-ablations}
    \vspace{-0.5em}
\end{figure*}
\vspace{-0.5em}
\section{Where should sparsity be distributed?}
When searching for efficient network architectures for \gls{drl} it is natural to ask where sparsity is best allocated. To that end, we consider both how to distribute parameters between network types and as well as within them.
\vspace{-0.5em}
\paragraph{Actor or Critic?} Although in value-based agents such as DQN there is a single network, in actor-critic methods such as PPO and SAC there are at least two: an actor and a critic network. It is believed that the underlying functions these networks approximate (e.g. a value function vs. a policy) may have significantly different levels of complexity, and this complexity likely varies across environments.
Actor and critic typically have near-identical network architectures.
However, for a given parameter budget it is not clear that this is the best strategy, as the complexity of the functions being approximated may vary significantly. We thus seek to understand how performance changes as the parameter ratio between the actor and critic is varied for a given parameter budget. In \autoref{fig:betweenNetDist} we assess three parameter budgets: 100\%, 20\% and 10\% of the standard dense parameter count, and two training regimes, dense and sparse. Given the observed similarity in performance between \gls{rigl} and \gls{set} in \autoref{fig:highLevelComparisonERK}, we selected one method, \gls{rigl}, for this analysis.

We observe that assigning a low proportion of parameters to the critic (10 - 20\%) incurs a high performance cost across all regimes. When parameters are more scarce, in 20\% and 10\% of standard dense settings, performance degradation is highest. This effect is not symmetric. Reducing the actor parameters to just 10\% rarely affects performance compared to the default actor-critic split of 34:66 (vertical line).

Interestingly the default split appears well tuned, achieving the best performance in most settings. However in the more challenging Humanoid environment we see that for smaller dense networks, reducing the actor parameters to just 10\% yields the best performance. Sparse networks follow a similar trend, but we notice that they appear to be more sensitive to the parameter ratio, especially at higher sparsities.

Overall this suggests that the value function is the more complex function to approximate in these settings, benefiting from the lion's share of parameters. It also suggests that tuning the parameter ratio may improve performance. Furthermore, FLOPs at evaluation time is determined only by the actor network. Since the actor appears to be easier to compress, this suggests large potential FLOPs savings for real-time usage of these agents. Finally, this approach could be used to better understand the relative complexity of policies and values functions across different environments.
\vspace{-0.5em}
\paragraph{Within network sparsity} In \autoref{fig:hyper-ablations} (left) we turn our attention to the question of distributing parameters within networks and compare two strategies; uniform and \gls{erk} \citep{evci2019rigl}. Given a target sparsity of, say 90\%, uniform achieves this by making each layer 90\% sparse; \gls{erk} distributes them proportional to the sum of its dimension, which has the effect of making large layers relatively more sparse than the smaller ones. Due to weight sharing in convolutional layers, ERK sparsity distribution doubles the FLOPs required at a given sparsity \citep{evci2019rigl}, which we also found to be the case with the convolutional networks used by DQN in the Atari environments (see \autoref{app:flops} for further discussion). On the other hand, ERK has no effect on the FLOPs count of fully connected networks used in MuJoCo environments.
Our results show that \gls{erk} significantly improves performance over uniform sparsity and thus we use \gls{erk} distribution in all of our experiments and share results with uniform distribution in \autoref{app:sec:additional_results_uniform}.

We hypothesize the advantage of \gls{erk} is because it leaves input and output layers relatively more dense, since they typically have few incoming our outgoing connections, and this enables the network to make better use of (a) the observation and (b) the learned representations at the highest layers in the network. It is interesting to observe that maintaining a dense output layer is one of the key design decisions made by \citet{Sokar2021DynamicST} for their proposed algorithm.

\begin{figure*}[!t]
    \centering
    \includegraphics[width=0.32\textwidth]{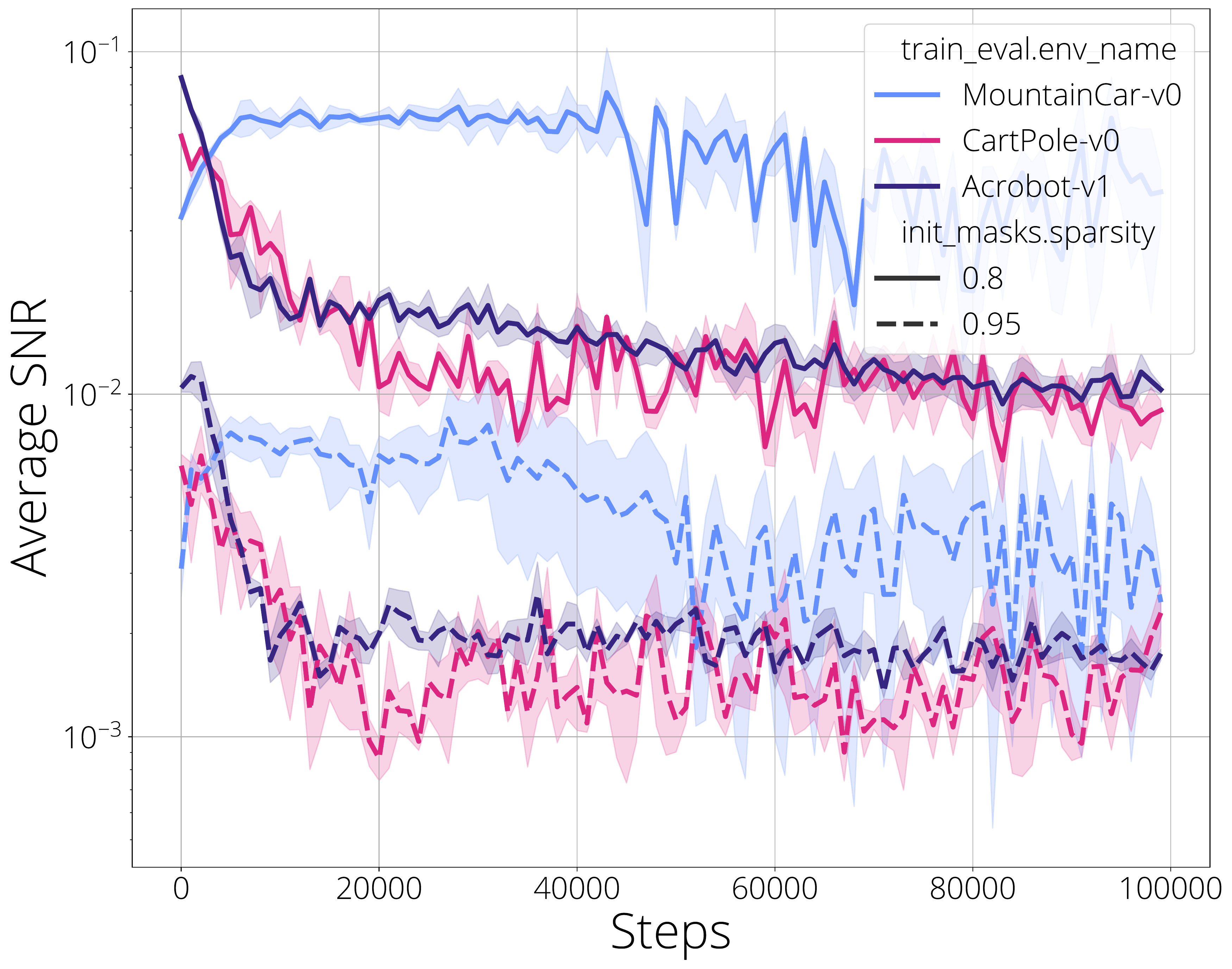}
    \includegraphics[width=0.32\textwidth]{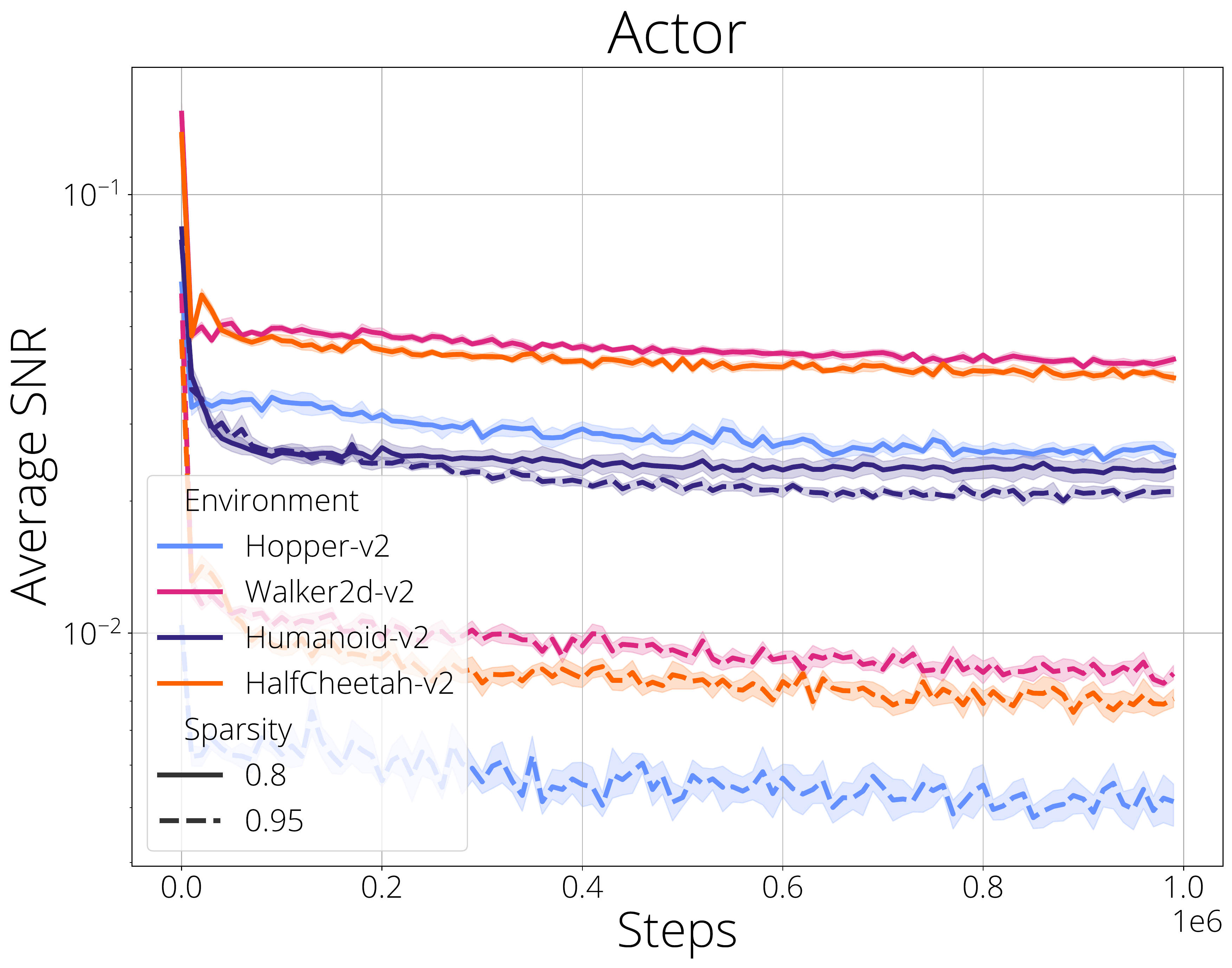}
    \includegraphics[width=0.32\textwidth]{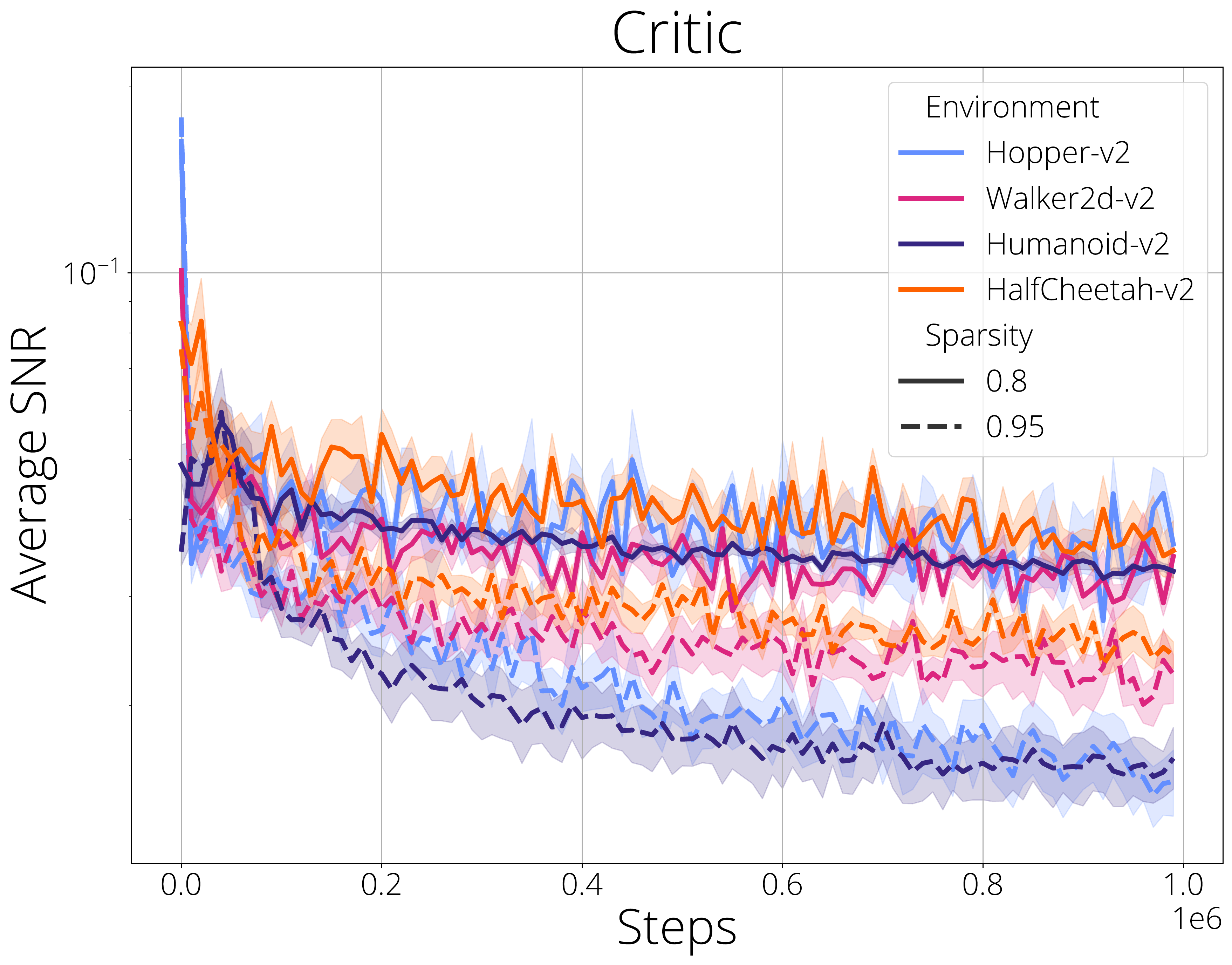}
    
    \caption{Signal-to-noise (SNR) ratio comparison of different gradient based DRL algorithms. We calculate SNR for every parameter in corresponding network (including the inactive/pruned weights) and report the mean SNR value. (left) DQN networks when training on classic control environments. SAC actor (center) and critic (right) networks during the training on MuJoCo environments.}
    \label{fig:snr}
    \vspace{-0.5em}
\end{figure*}
\vspace{-0.5em}
\section{Sensitivity analysis}
In this section we assess the sensitivity of some key hyper-parameters for sparse training. Plots and commentary for the remaining hyper-parameters (drop fraction, topology update interval, and batch size) are shared in \autoref{app:additional_hypers}.
\vspace{-0.5em}
\paragraph{Weight decay}
In \autoref{fig:hyper-ablations} (center) we evaluate the effect of weight decay and find that a small amount of weight decay is beneficial for pruning, \gls{rigl}, and \gls{set}. This is to be expected since network topology choices are made based on weight magnitude, although we do note that the improvements are quite minor. Surprisingly weight decay seems to help dense even though it is not often used in \gls{drl}.

\textit{Findings:} We recommend using small weight decay.
\vspace{-0.5em}
\paragraph{Sparsity-aware initialization}
In \autoref{fig:hyper-ablations} (right) we evaluate the effect of adjusting layer weight initialization based on a layer's sparsity on static, \gls{rigl} and \gls{set}. A common approach to initialization is to scale a weight's initialization inversely by the square root of the number of incoming connections. Consequently, when we drop incoming connections, the initialization distribution should be scaled proportionately to the number of incoming connections \citep{Evci2020GradientFI}. \autoref{fig:hyper-ablations} (right) shows that this sparsity-aware initialization consistently improves performance when using uniform distribution over layer sparsities. However the difference disappears when using \gls{erk} for \gls{rigl} and \gls{set} and may even harm performance for static.

\textit{Findings:} Performance is not sensitive to sparsity-aware initialization when using \gls{erk} and helps when using uniform layer sparsity. For \gls{rigl} and \gls{set} we recommend always using sparsity-aware weight initialization (since it never appears to harm performance) but for static this may depend on layer sparsity.
\vspace{-0.5em}
\section{Signal-to-noise ratio in DRL environments}
Variance reduction is key to training deep models and often achieved through using momentum based optimizers \cite{schmidt2011,Kingma2015AdamAM}. However when new connections are grown such averages are not available, therefore noise in the gradients can provide misleading signals. In \autoref{fig:snr} we share the \gls{snr} for the Classic control and MuJoCo environments over the course of training. \gls{snr} is calculated as $\frac{|\mu|}{\sigma}$ where $\mu$ is the mean and $\sigma$ is the standard deviation of gradients over a mini-batch. A low \gls{snr} means the signal is dominated by the variance and thus the mean (the signal) is uninformative. We calculate \gls{snr} for all parameters separately and report the mean. Mini-batch gradients can have average \gls{snr} values as low as 0.01 starting early in training. Higher sparsities seem to cause lower \gls{snr} values. Similarly, actor networks have lower \gls{snr}.  

\textit{Findings:} We find the average \gls{snr} for gradients to decrease with sparsity, potentially explaining the difficulty of using gradient based growing criteria in sparse training.

\begin{figure*}[!h]
    \centering
    \includegraphics[width=0.3\textwidth]{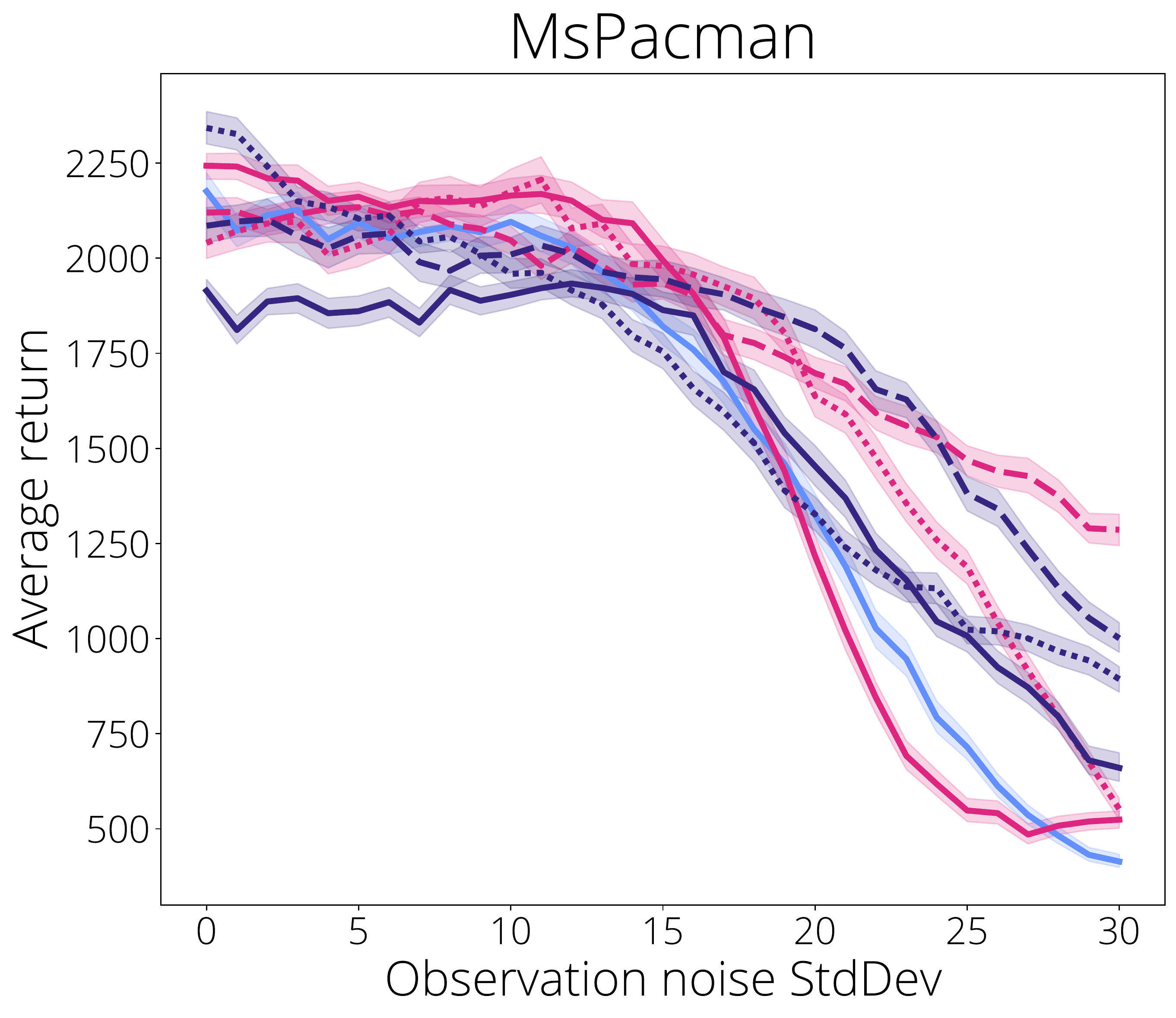}
    \includegraphics[width=0.3\textwidth]{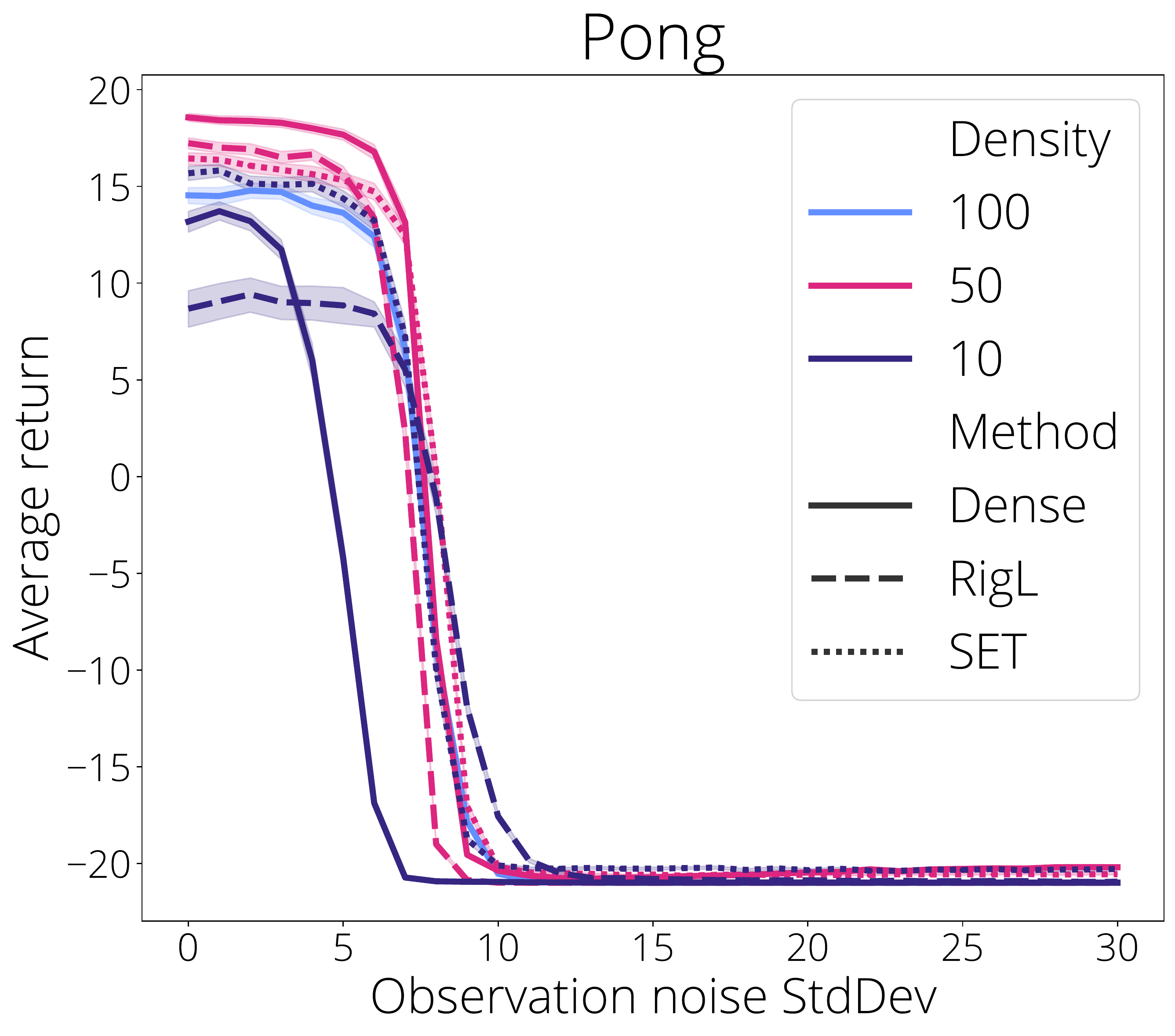}
    \includegraphics[width=0.3\textwidth]{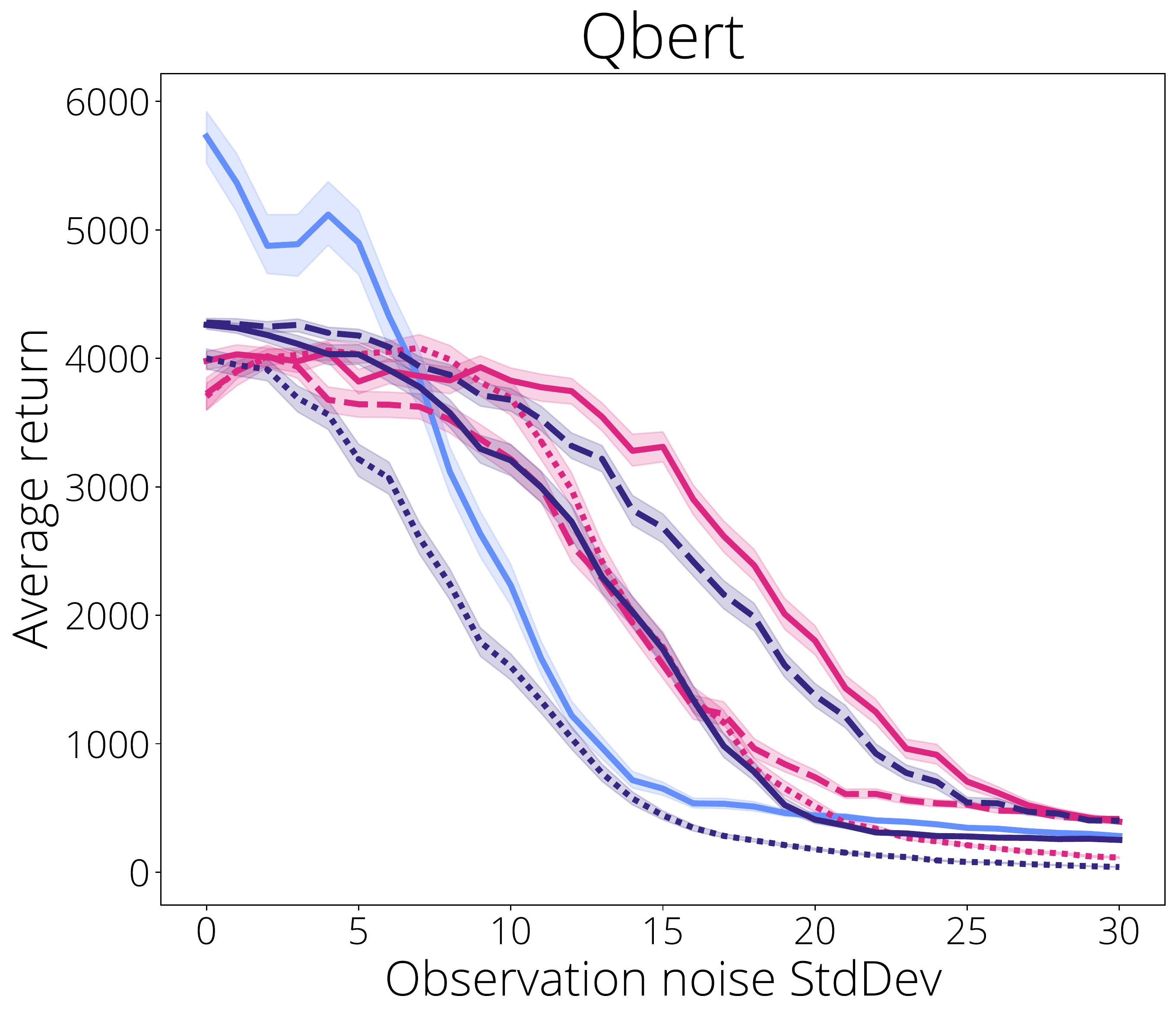}
    \caption{Robustness to observation noise. We test the robustness of networks trained using sparse and dense methods (denoted by line style) by adding Gaussian noise to the observations. We examine three parameter regimes, 100\% (blue), 50\% (pink) and 10\% (purple) of the standard dense model parameter count. All policies were trained using DQN.}
    \label{fig:generalization}
\end{figure*}
\vspace{-0.5em}
\section{Are sparse networks robust to noise?}
\label{sec:generalization}
Sparse neural networks can improve results on primary metrics such accuracy and rewards, yet they might have some unexpected behaviours in other aspects \citep{Hooker2020WhatDC}. In \autoref{fig:generalization} we assess the effect of adding increasing amounts of noise to the observations and measuring their effect on a trained policy. Noise was sampled $\sim \mathcal{N}(0, \sigma), \sigma \in [0, 1, ..., 30]$, quantized to an integer, and added to each observation's pixel values ($\in [0, 255]$) before normalization. Noise was sampled independently per pixel. We look at three data regimes; 100\%, 50\% and 10\% of the standard dense model parameter count and compare dense and sparse training (\gls{rigl} and \gls{set}). 
We made an effort to select policies with comparable performance for all the methods, chosen from the set of all policies trained during this work.

We observe that (1) smaller models are generally more robust to high noise than larger models, (2) sparse models are more robust to high noise than dense models on average, and (3) in most cases there are minimal differences when the noise is low.

We can see that in the very low data regime (10\% full parameter count) policies trained using \gls{rigl} are more robust to high noise compared with their dense counterparts, a fact observed across every environment. In the moderate data regime (50\% full parameter count) the ordering is more mixed. In Qbert the dense model is most robust but the picture is reversed for Pong and McPacman. Finally, \gls{set} appears less robust to high noise than \gls{rigl}, although we note this is not the case for Pong at 50\% density.

Although a preliminary analysis, it does suggest that sparse training can produce networks that are more robust to observational noise, even when experienced post-training.

\vspace{-0.5em}
\section{Related work}
\paragraph{Sparse training} Though research on pruning has a relatively long story by deep learning standards \citep{mozer1989skeletonization,sietsma1988neural,han2015learning,variational-dropout,louizos2017bayesian}, training sparse networks from scratch has only recently gained popularity. \citet{Goyal2017}, \citet{Frankle2018lottery}, and \citet{Evci2019difficulty} showed that training sparse networks from a random initialization is difficult compared to dense neural networks. Despite this, various approaches have been recently proposed to improve sparse training, most notably lottery tickets \citep{Frankle2019stabilizing,Zhou2019deconstructing}, dynamic sparse training \citep{dettmers2019,Mostafa2019,Mocanu2018,Bellec2017,evci2019rigl,Liu2021DoWA} and one-shot pruning \citep{snip,grasp,tanaka2020pruning,Liu2020FindingTS}. Solutions that focus on initialization alone have been shown to be ineffective for contemporary models \citep{Evci2020GradientFI,Frankle2020pruningatinit}, possibly due to the catapult mechanism observed early in training \citep{lewkowycz2020large}. For an in-depth survey on the topic, please see \citet{Hoefler2021SparsityID}.

\vspace{-0.5em}
\paragraph{Sparse networks in RL} \citet{Livne2020PoPSPP} used pruning as an intermediary step to guide the width of dense neural networks for DQN and A2C agents. Most of the work investigating the use of sparse sparse training in RL are in the context of the lottery ticket hypothesis;  
\citet{morcos2019one} studied the existence of lucky sparse initializations using pruning and late-rewinding; \citet{hasani20naturallotttery} proposed an interesting approach by repurposing the sparse circuitry of the C. elegans soil-worm for RL tasks; \citet{Vischer2021OnLT} observed that the success of such initializations dependens heavily on selecting correct features for the input data, and not on any general qualities of the different initializations. \citet{Lee2021GSTGT} proposed the use of block-circulant masks during early steps of training to improve the efficiency of pruning on TD3 agents, while \citet{Arnob2021SingleShotPF} applied one-shot pruning algorithms in an offline-RL setting. Perhaps the work closest to ours is the algorithm proposed by \citet{Sokar2021DynamicST}, where authors applied the \gls{set} algorithm for end-to-end training of sparse networks in two actor-critic algorithms (TD3 and SAC). By a carefully chosen topology update schedule and dynamic architecture design, the proposed algorithm was able to match the dense network with a sparsity of around 50\%.

\vspace{-0.5em}
\paragraph{Novel architectures in \gls{drl}} A number of works have focused on evolving network architectures for RL policies. \citet{Nadizar2021} applied pruning together with evolution algorithms. \citet{whiteson06a} combined NEAT  \citep{neat} with Q-learning \citep{Watkins:1989} to evolve better learners, \citet{wann2019} evolved strong architectural priors, resulting in networks that could solve tasks with a single randomly initialized shared weight, whilst \citet{tang-self-interp-20} evolve compact self-attention architectures as a form of indirect network encoding. \citet{zambaldi2018deep} similarly explored self-attention enabling agents to perform relational reasoning and achieve state-of-the-art performance on the majority of StarCraft II mini-games. Another line of research seeks to improve the stability \citep{parisotto20} and efficiency \citep{parisotto2021efficient} of transformers applied to \gls{drl}, whilst \citet{rrl-shah21a} explore the utility of using features extracted from a pre-trained Resnet in the standard \gls{drl} pipeline. Consistent with our observations in this work, \citet{world-mod-2018} showed it is possible to train very compact controllers (i.e. actors) albeit in a the context of model-based instead of the model-free RL setting considered here.

\vspace{-0.5em}
\section{Discussion and Conclusion}
In this work we sought to understand the state of sparse training for \gls{drl} by applying pruning, static, \gls{set} and \gls{rigl} to DQN, PPO, and SAC agents trained on a variety of environments. We found sparse training methods to be a drop-in alternative for their dense counterparts providing better results for the same parameter count. From a practical standpoint we made recommendations regarding hyper-parameter settings and showed that non-uniform sparse initialization combined with tuning actor:critic parameter ratios improves performance. 

We hope this work establishes a useful foundation for future research into sparse \gls{drl} algorithms and highlights a number of interesting research questions. In contrast to the computer vision domain, we observe that \gls{rigl} fails to match pruning results. Low SNR in high sparsity regimes offers a clue but more work is needed to understand this phenomena. Our results in \autoref{sec:generalization} also suggest that sparse networks may aid in generalization and robustness to observational noise; this is an active area of interest and research in the \gls{drl} community, so a more thorough understanding could result in important algorithmic advances.

\vspace{-0.5em}
\section*{Acknowledgements}
We thank Fabian Pedregosa, Rishabh Agarwal, and Adrien Ali Taïga for their helpful feedback on the manuscript, Oscar Ramirez for his help with on the TF-Agents codebase, and Trevor Gale and Sara Hooker for inspiring the title of this work. We also thank Brain Montreal RL team for their useful feedback on an early version of this work. Finally, we thank Bram Grooten for pointing out \citet{plasmaRL2022} and their contribution to the motivation for this work.

\nocite{langley00}

\bibliography{main.bib}
\bibliographystyle{icml2022}

\newpage
\appendix
\onecolumn
\section{Experimental Details}
\subsection{Training schedule}
\label{app:exp-details}
During training all agents are allowed $M$ environment transitions, with policies being evaluated for $K$ episodes / steps every $N$ environment frames, where the values vary per suite and shared below. Atari experiments use a frame skip of 4, following \cite{mnih15human} thus 1 environment step = 4 environment frames. 
\begin{center}
\begin{tabular}{c c c c} 
 \toprule
 Environment & M & N & K \\
 \midrule
 \multicolumn{4}{l}{\textit{K = episodes}} \\
 {\bf Classic control} & $100,000$ & $2,000$ & $20$ \\ 
 {\bf MujoCo (SAC)} & $1,000,000$ & $10,000$ & $30$ \\ 
 {\bf MujoCo (PPO)} & $1,000,000$ & $2,000$ & $20$ \\
 \midrule
 \multicolumn{4}{l}{\textit{K = environment steps}} \\
 {\bf Atari Suite} & $40,000,000$ & $1,000,000$ & $125,000$ \\ 
 \bottomrule
\end{tabular}
\end{center}

\subsection{FLOPs behaviour of Sparse ERK networks in \gls{drl}}
\label{app:flops}
As reported in \citet{evci2019rigl}, using ERK sparsity distribution often doubles the FLOPs needed for sparse models compared using the uniform sparsity distribution. This is due to the parameter sharing in convolutional layers. The spatial dimensions, kernel size and the stride of a convolutional layer affects how many times each weight is used during the convolution which in turn determines the contribution of each weight towards the total FLOPs count. In modern CNNs, the spatial dimensions of the feature maps often decreases monotonically towards the output of the network, making the contribution of the connections in later layers to the total FLOPs count smaller. Furthermore, the size of the layers typically increases towards the output and thus ERK removes a larger proportion of the connections from these later layers compared to uniform. Consequently a network sparsified using the ERK distribution will have a larger FLOPs count compared to one sparsified using a uniform distribution.

\begin{figure*}[!h]
    \centering
    \includegraphics[width=0.48\textwidth]{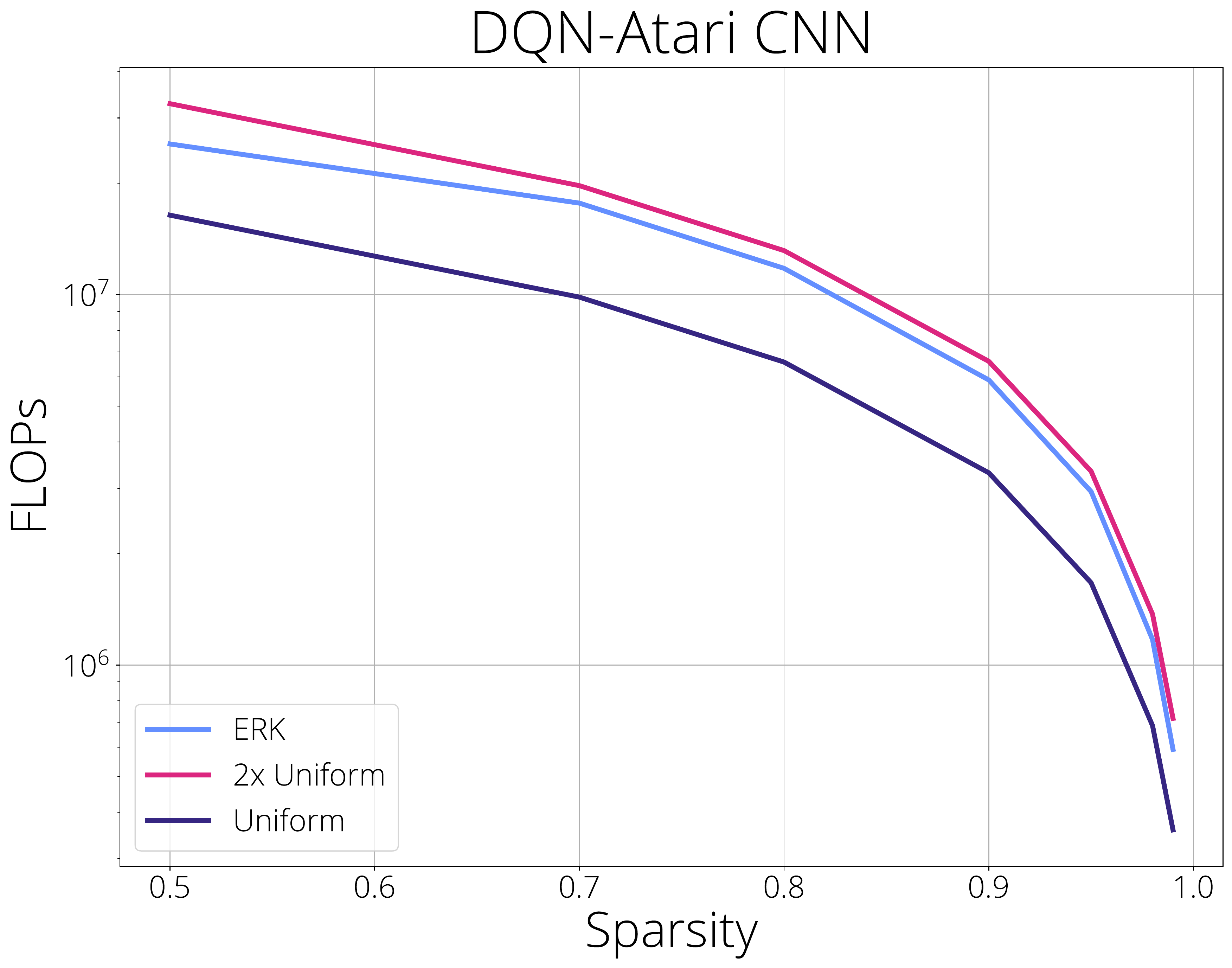}
    \includegraphics[width=0.48\textwidth]{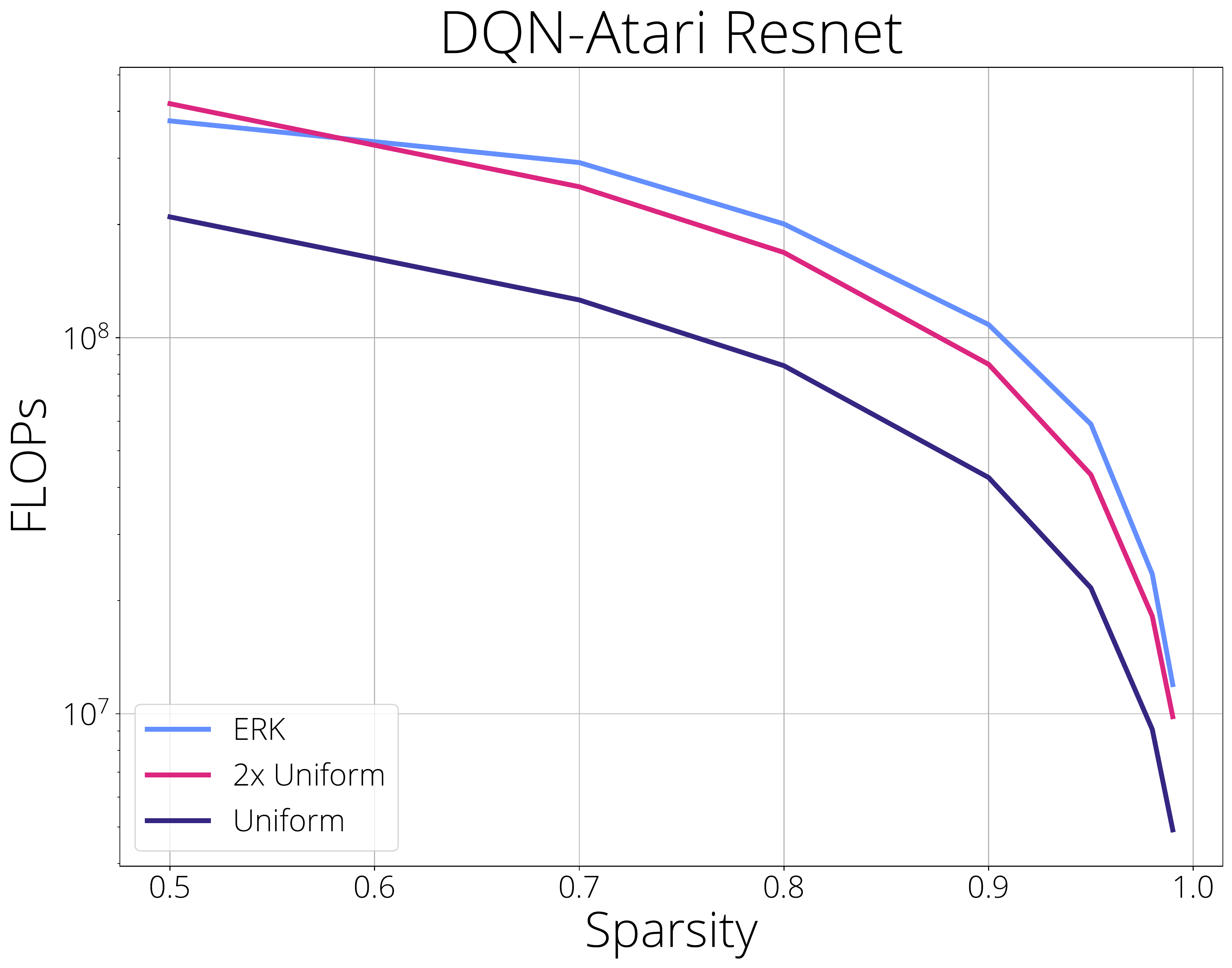}
    \caption{FLOPs scaling of different sparsity distributions on (left) Nature CNN and (right) Impala ResNet architectures used in MsPacman environment.}
    \label{fig:flops}
\end{figure*}

Due to the lack of parameter sharing fully connected layers used in MuJoCo and classic control experiments, sparse networks with ERK have same amount of FLOPs as the uniform. Networks used in Atari experiments, however, uses convolutional networks and thus ERK doubles the FLOPs required compared to uniform. FLOPs scaling of sparse networks with ERK distributions used for Pong game can be found in \autoref{fig:flops}. Atari games have differently sized action spaces (Pong has 6 actions for example), which affects the number of neurons in the last layer. However since the last layer is very small and fully connected, it should have a very little effect on the results provided here.

\subsection{Hyper-parameter sweep}
\label{app:hyper-details}
We perform a grid search over different hyper parameters used in \denseshp{} \textit{Dense}, \pruneshp{} \textit{Prune}, \staticshp{} \textit{Static}, \setshp{} \textit{\gls{set}} and \riglshp{} \textit{\gls{rigl}} algorithms. Unless otherwise noted, we use hyper-parameters used in regular dense training. When pruning, we start pruning around 20\% of training steps and stop when 80\% of training is completed following the findings of \citet{gale2019state}. We use same default hyper-parameters for \gls{set} and \gls{rigl}. Fort both algorithms we start updating the mask at initialization and decay the drop fraction over the course of the training using a cosine schedule, similar to pruning stopping the updates when 80\% of training is completed. 

We search over the following parameters:
\begin{enumerate}
    \item \textbf{Weight decay (\denseshp{} \pruneshp{} \staticshp{} \setshp{} \riglshp{}):} Searched over the grid [0, 1e-6, 1e-4, 1e-3].
    \item \textbf{Update Interval (\pruneshp{} \setshp{} \riglshp{}):} refers to how often models are pruned or sparse topology is updated. Searched over the grid [100, 250, 500, 1000, 5000]. 
    \item \textbf{Drop Fraction (\pruneshp{} \setshp{} \riglshp{}):} refers to the maximum percentage of parameters that are dropped and added when network topology is updated. This maximum value is decayed during training according to a cosine decay schedule. Searched over the grid [0.0,0.1,0.2,0.3,0.5]. 
    \item \textbf{Sparsity-aware initialization (\staticshp{} \setshp{} \riglshp{}):} refers to whether sparse models are initialized with scaled initialization or not.
\end{enumerate}

We repeat the hyper-parameter search for each \gls{drl} algorithm using the Acrobot (for DQN) and Walker2D (for PPO and SAC) environments. Best hyper-parameters found in these environments are then used when training in other similar environments (i.e. classic control for DQN and MuJoCO for PPO and SAC). See \autoref{tab:DQN-hypers-sweep}, \autoref{tab:SAC-hypers-sweep}, and \autoref{tab:PPO-hypers-sweep} for the best hyper parameters found in each setting.

\begin{table}[!h]
\renewcommand{\arraystretch}{1.1}
\centering
\caption{DQN best hyper-parameters for Classic Control from sweep}
\label{tab:DQN-hypers-sweep}
\input{tables/dqn_hyper_sweep}
\end{table}

\begin{table}[!h]
\renewcommand{\arraystretch}{1.1}
\centering
\caption{SAC best hyper-parameters from sweep}
\label{tab:SAC-hypers-sweep}
\input{tables/sac_hyper_sweep_80}
\end{table}

\begin{table}[!h]
\renewcommand{\arraystretch}{1.1}
\centering
\caption{PPO best hyper-parameters from sweep}
\label{tab:PPO-hypers-sweep}
\input{tables/ppo_hyper_sweep}
\end{table}

\hfill \break
\hfill \break
\hfill \break
\hfill \break
\hfill \break
\hfill \break

\paragraph{Atari hyper-parameters} Due to computational constraints we did not search over hyper-parameters for the Atari environments, except for a small grid-search to tune the dense ResNet. The CNN architecture from \citet{mnih15human} has been used in many prior works thus was already well tuned. The ResNet hyper-parameter sweep for the original dense model is detailed below:
\begin{enumerate}
    \item \textbf{Weight decay:} Searched over the grid [0, 1e-6, 1e-5, 1e-4].
    \item \textbf{Learning rate:} Searched over the grid [1e-4, 2.5e-4, 1e-3, 2.5e-3].
\end{enumerate}

The final hyper-parameters we used for the Atari environments are shown in in \autoref{tab:DQN-CNN-Atari-hypers} for the CNN and \autoref{tab:DQN-ResNet-Atari-hypers} for the ResNet.

\begin{table}[!h]
\renewcommand{\arraystretch}{1.1}
\centering
\caption{DQN (CNN) Atari hyper-parameters}
\label{tab:DQN-CNN-Atari-hypers}
\input{tables/dqn_atari_cnn_hypers}
\end{table}

\begin{table}[!h]
\renewcommand{\arraystretch}{1.1}
\centering
\caption{DQN (ResNet) Atari hyper-parameters}
\label{tab:DQN-ResNet-Atari-hypers}
\input{tables/dqn_atari_resnet_hypers}
\end{table}

\paragraph{Remaining hyper-parameters} Next, we include details of the \gls{drl} hyper-parameters used in all training settings for DQN (\autoref{tab:DQN-hypers} and \autoref{tab:DQN-atari-arch}), SAC (\autoref{tab:SAC-hypers}), and PPO (\autoref{tab:SAC-hypers}).

\begin{table}[!h]
\renewcommand{\arraystretch}{1.1}
\centering
\caption{DQN Hyperparameters/ Table format from \citet{haarnoja18sac}.}
\label{tab:DQN-hypers}
\input{tables/dqn_hypers}
\end{table}

\begin{table}[!h]
\renewcommand{\arraystretch}{1.1}
\centering
\caption{DQN Atari: CNN and ResNet Architectures}
\label{tab:DQN-atari-arch}
\input{tables/dqn_network_architectures}
\end{table}

\begin{table}[!h]
\renewcommand{\arraystretch}{1.1}
\centering
\caption{SAC Hyperparameters.}
\label{tab:SAC-hypers}
\input{tables/sac_hypers}
\end{table}

\begin{table}[!h]
\renewcommand{\arraystretch}{1.1}
\centering
\caption{PPO Hyperparameters.}
\label{tab:PPO-hypers}
\input{tables/ppo_hypers}
\end{table}

\hfill \break
\hfill \break
\hfill \break
\hfill \break
\hfill \break
\hfill \break
\hfill \break
\hfill \break
\hfill \break
\hfill \break
\hfill \break
\hfill \break
\hfill \break
\hfill \break
\hfill \break
\hfill \break
\hfill \break
\hfill \break
\hfill \break
\hfill \break
\hfill \break
\hfill \break
\hfill \break
\newpage

\subsection{Atari Game Selection}
\label{sec:atari-game-selection}

Our original three games (MsPacman, Pong, Qbert) were selected to have varying levels of difficulty as measured by DQN's human normalized score in \citet{mnih15human}, Figure 3. To this we added 12 games (Assault, Asterix, BeamRider, Boxing, Breakout, CrazyClimber, DemonAttack, Enduro, FishingDerby, SpaceInvaders, Tutankham, VideoPinball) selected to be roughly evenly distributed amongst the games ranked by DQN's human normalized score in \citet{mnih15human} with a lower cut off of approximately 100\% of human performance.

\section{Sparse Scaling Plots in Other Environments}
\label{sec:additional_results_erk}

\begin{figure*}[!h]
    \centering
    \includegraphics[width=0.3\textwidth]{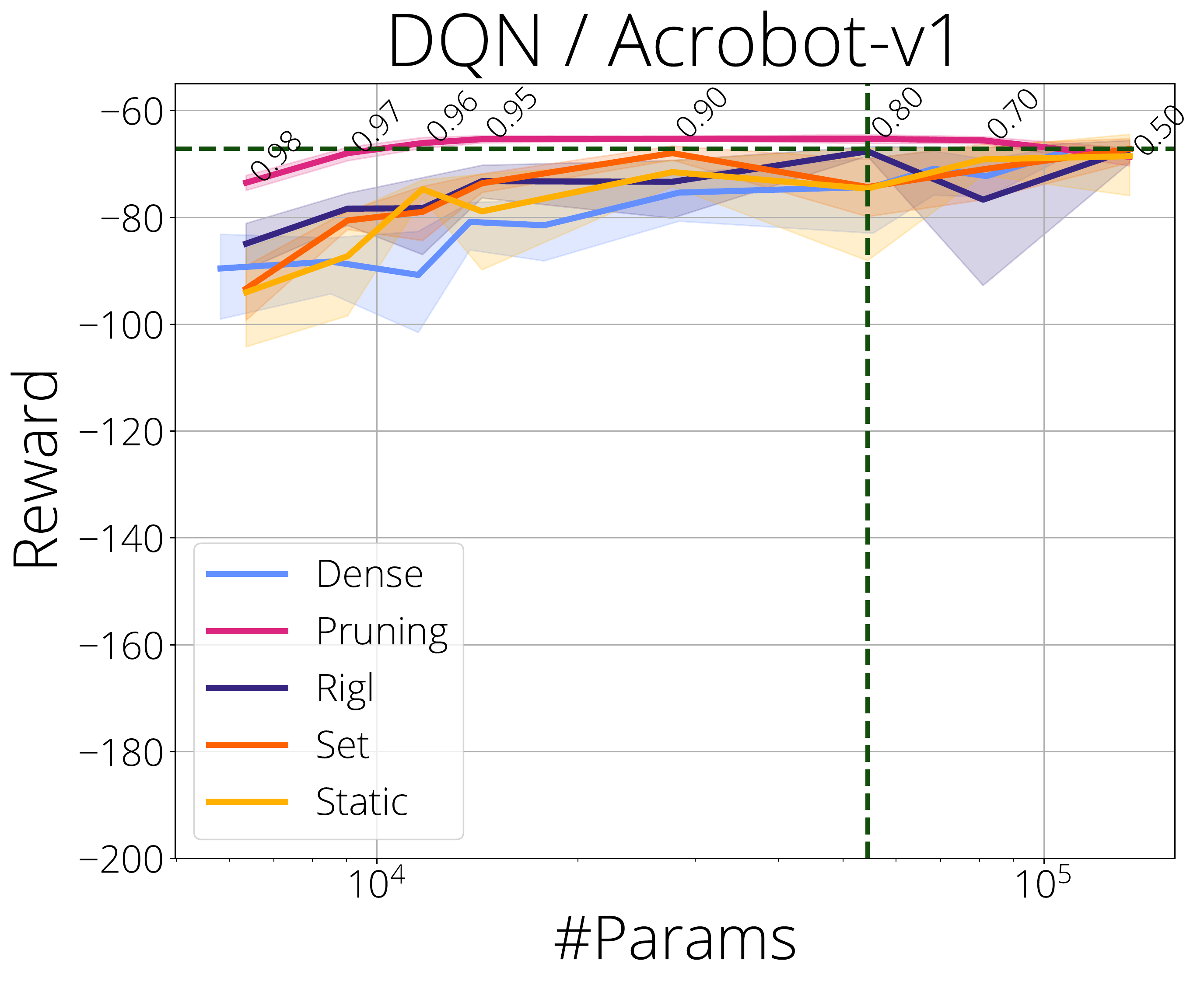}
    \includegraphics[width=0.3\textwidth]{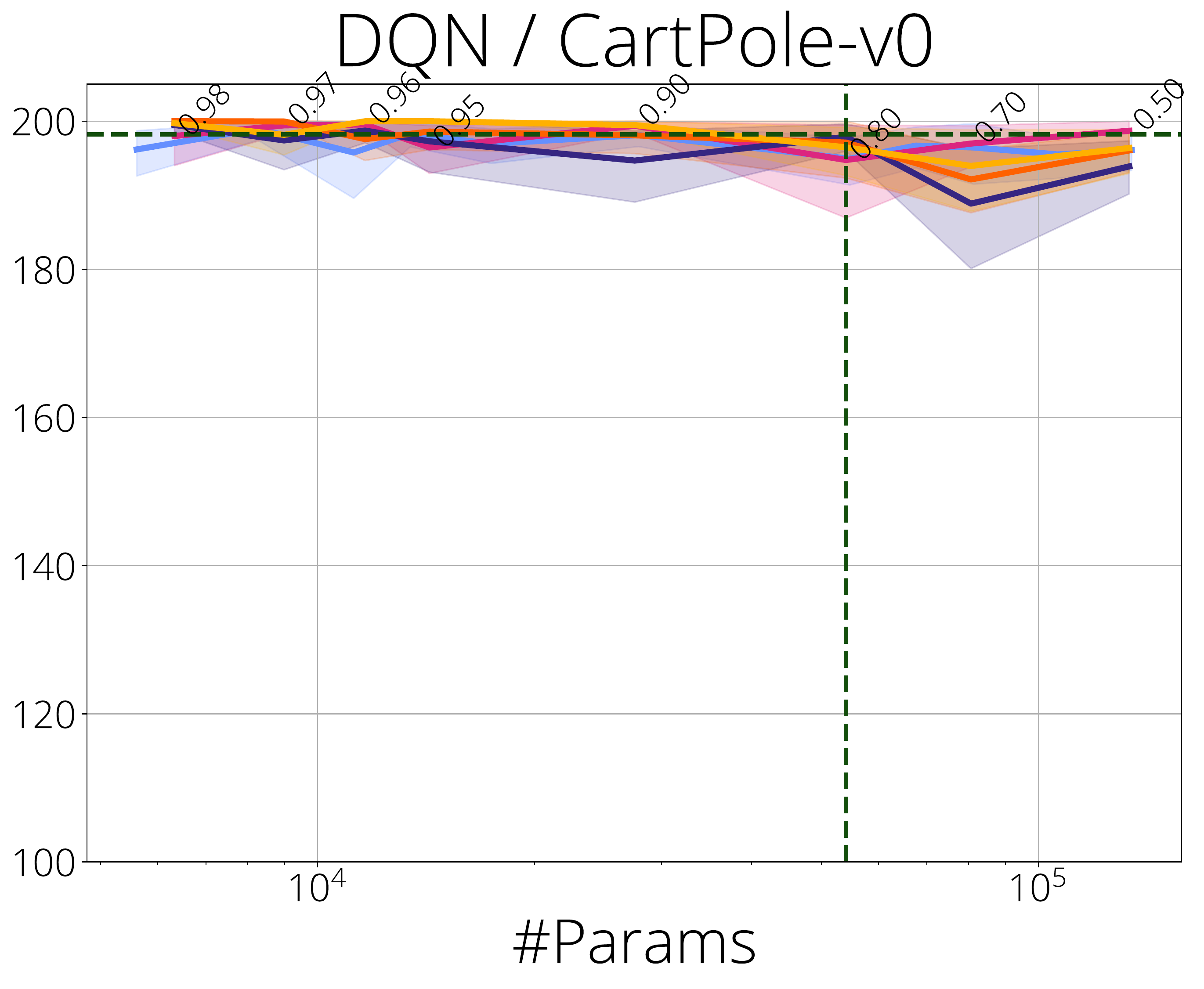}
    \includegraphics[width=0.3\textwidth]{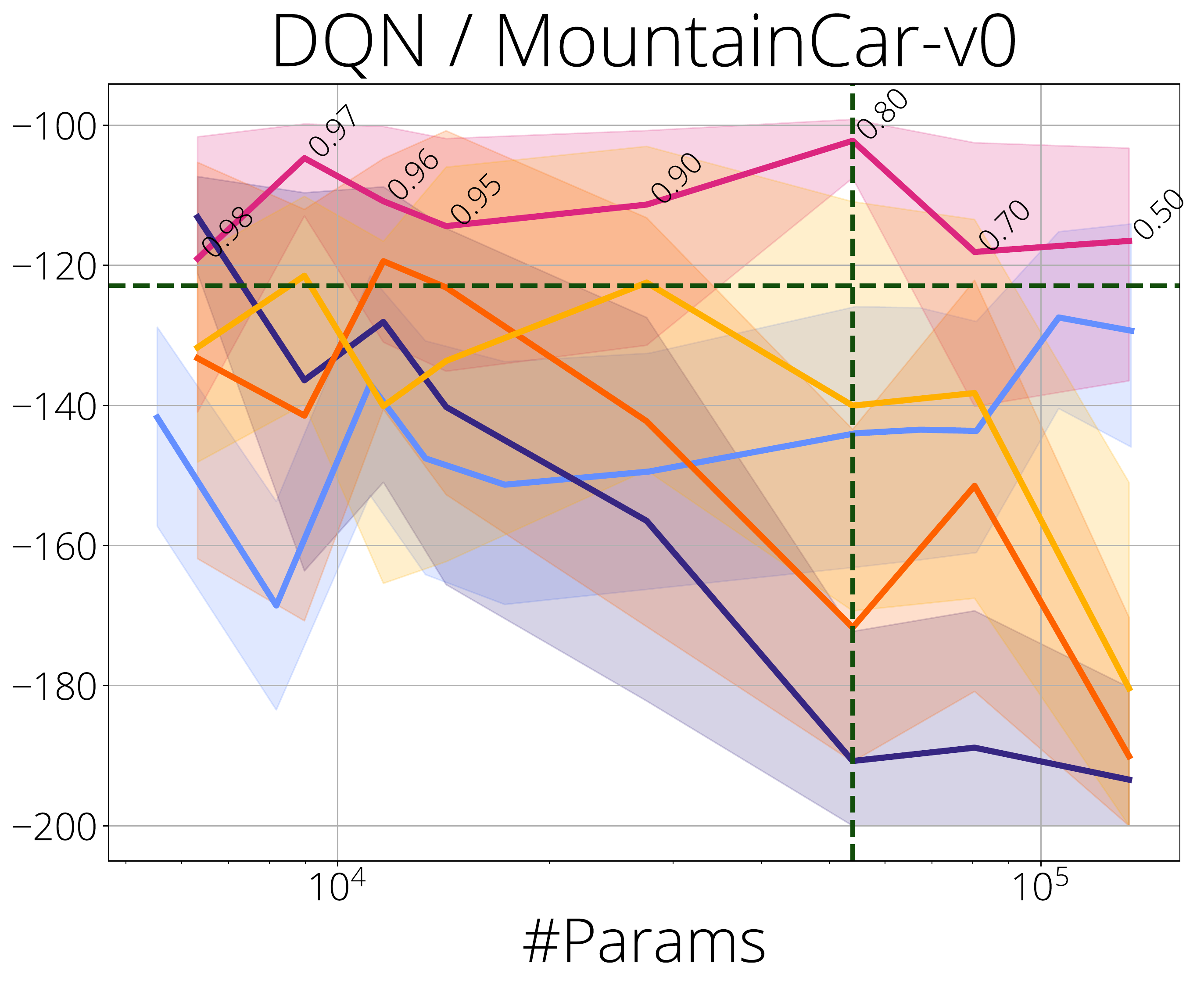}
    \includegraphics[width=0.3\textwidth]{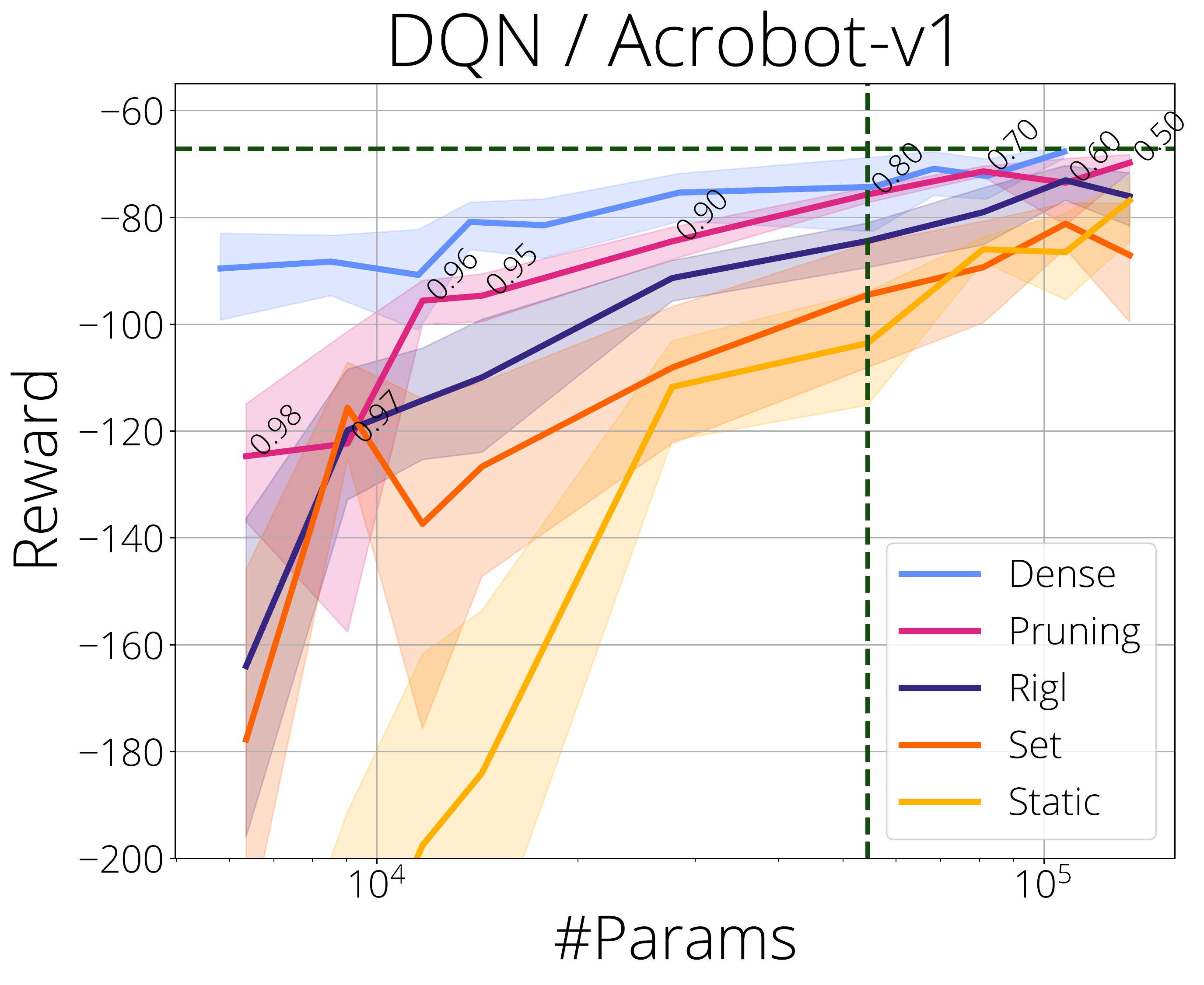}
    \includegraphics[width=0.3\textwidth]{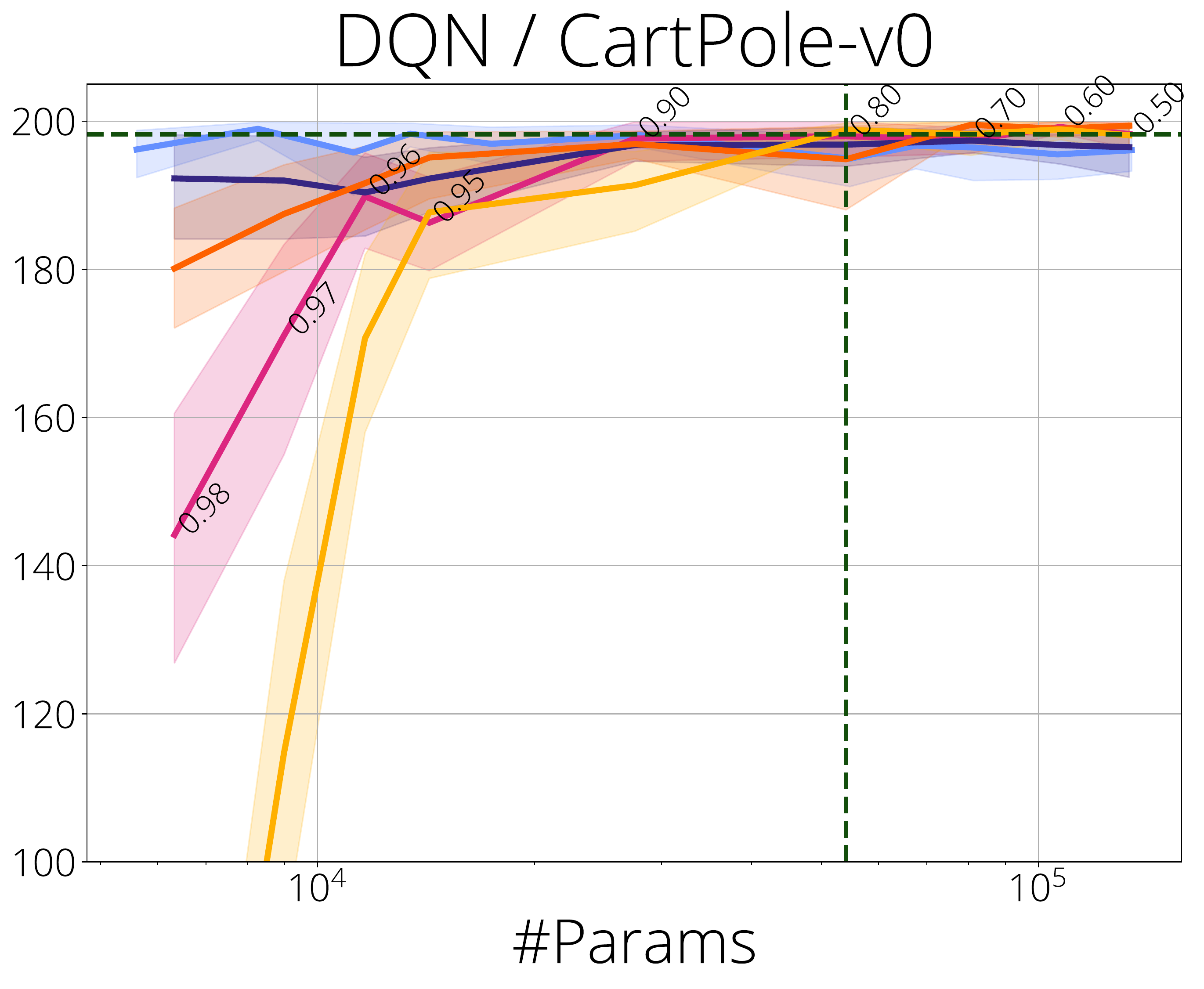}
    \includegraphics[width=0.3\textwidth]{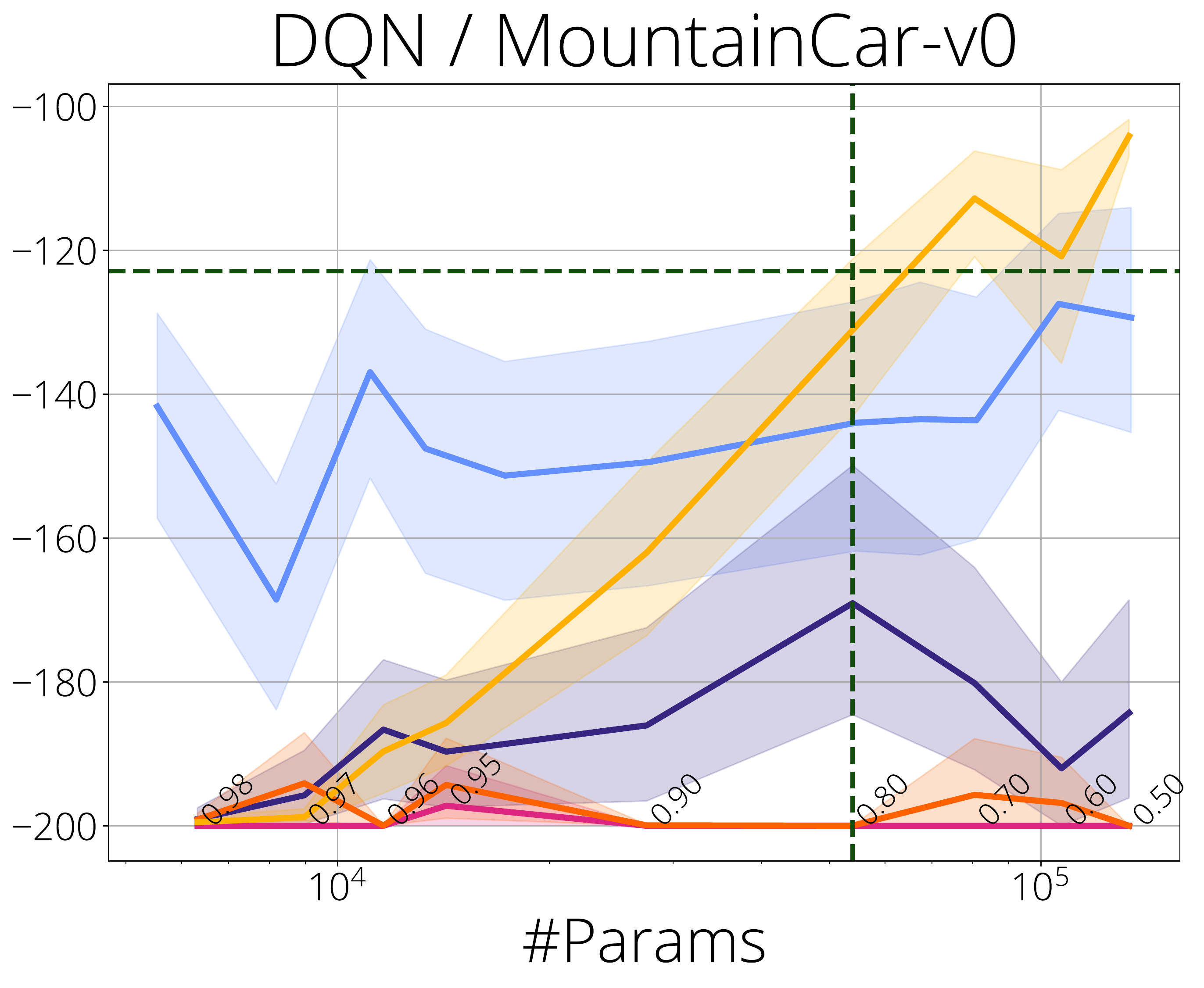}
    \caption{DQN in the Classic Control environments with \gls{erk} network sparsity distribution (top) and uniform network sparsity (bottom).}
    \label{fig:dqn-classic-app}
\end{figure*}

\begin{figure}[H]
    \centering
    \includegraphics[width=0.3\textwidth]{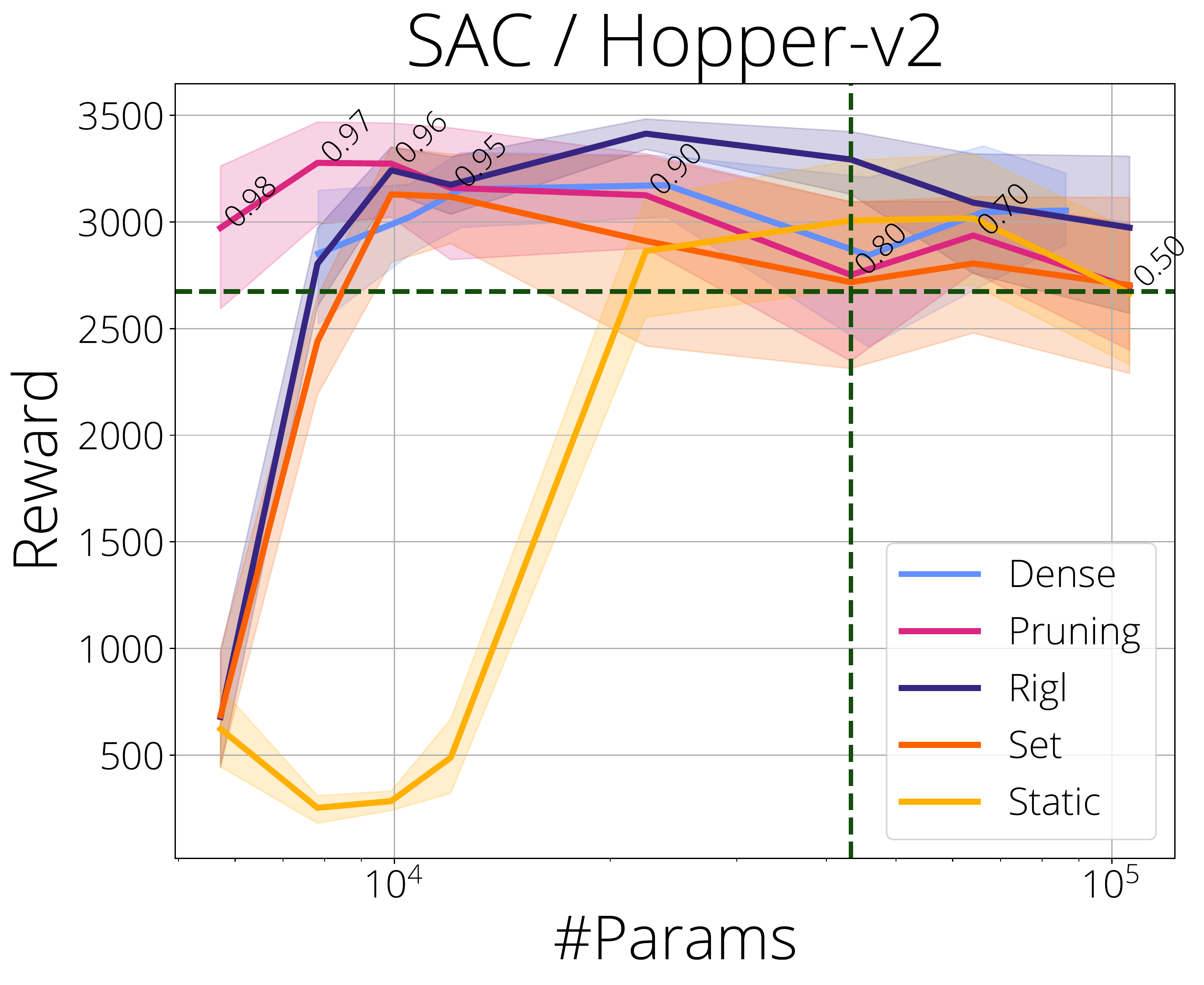}
    \includegraphics[width=0.3\textwidth]{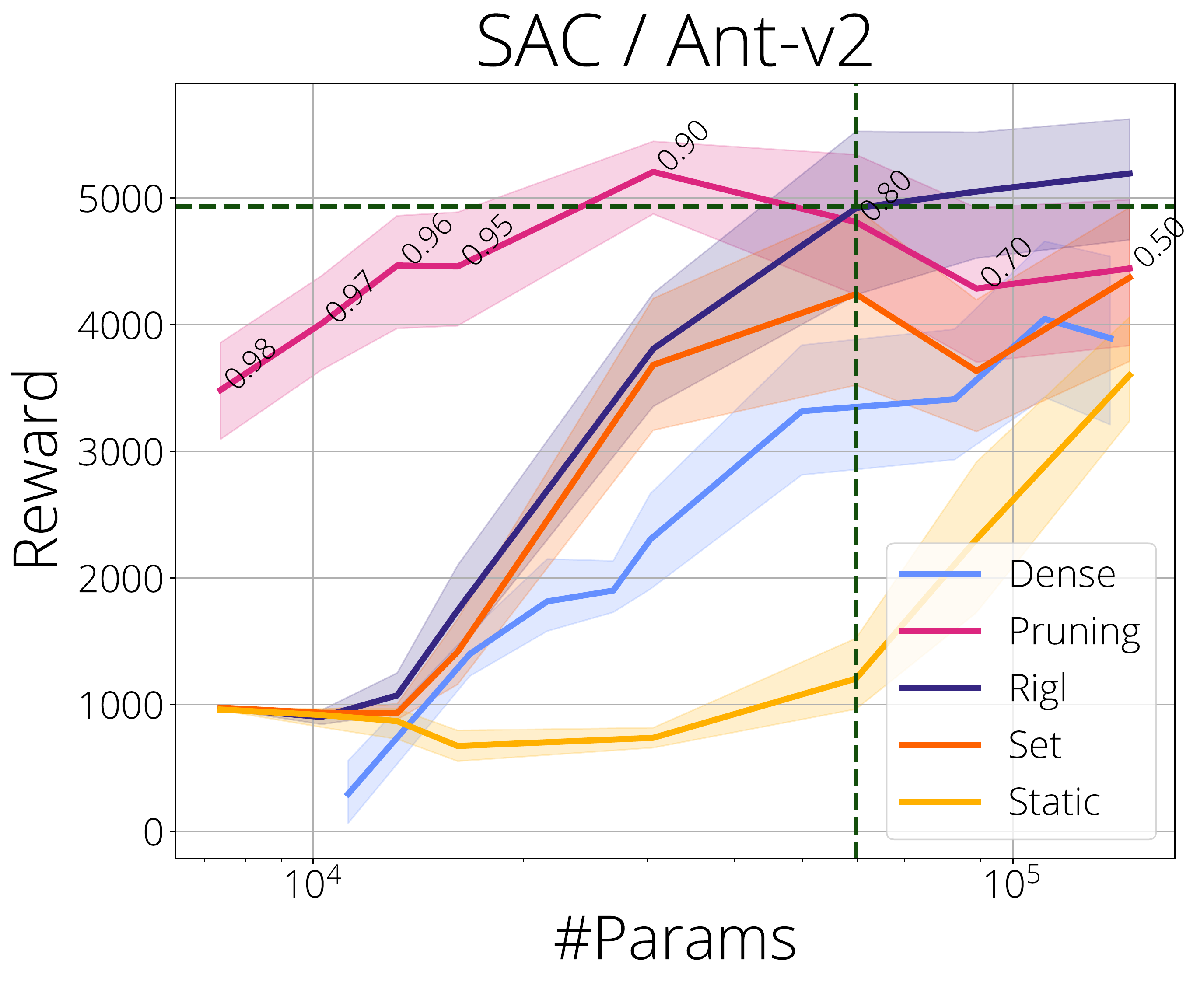} \\
    \includegraphics[width=0.3\textwidth]{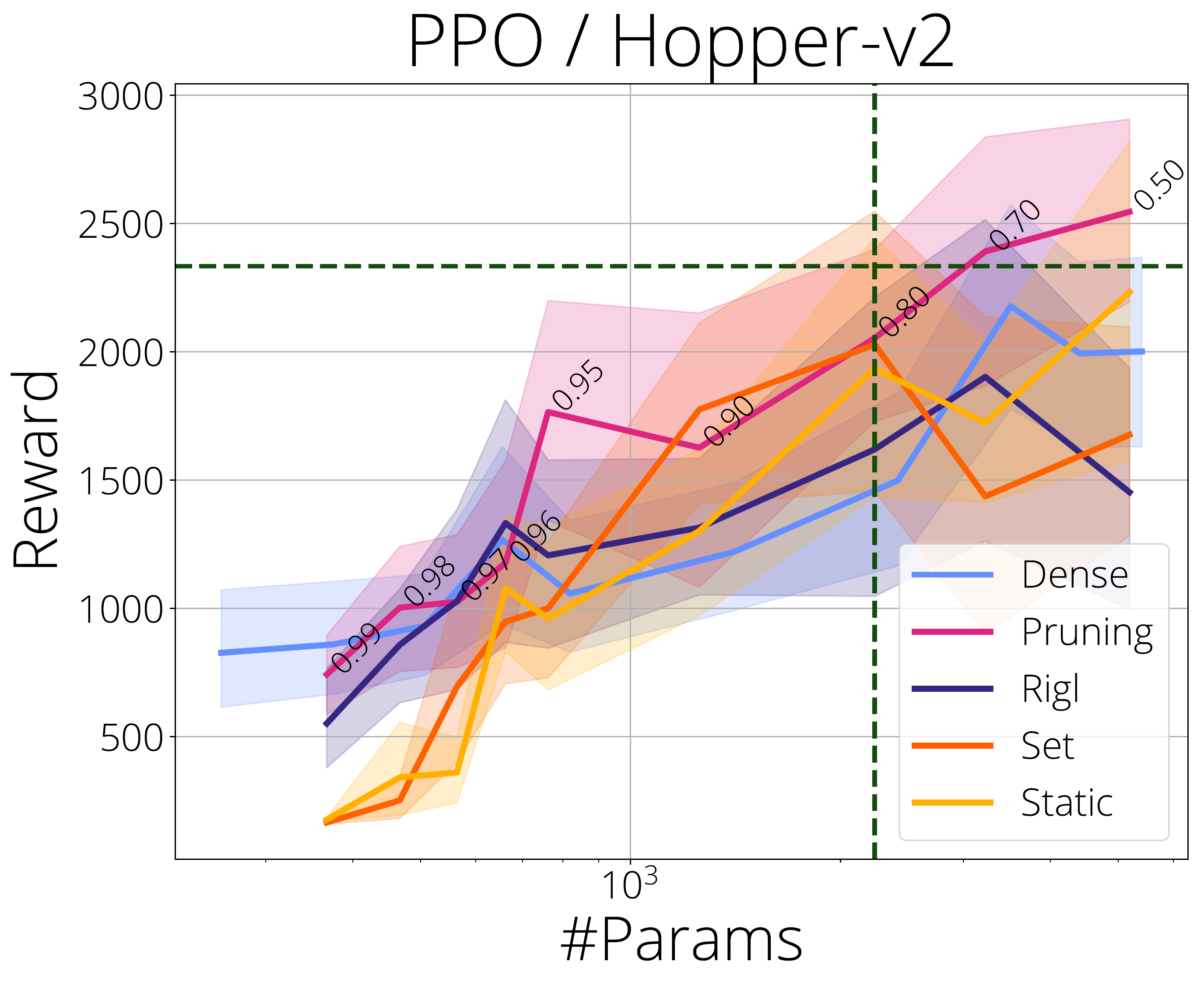}
    \includegraphics[width=0.3\textwidth]{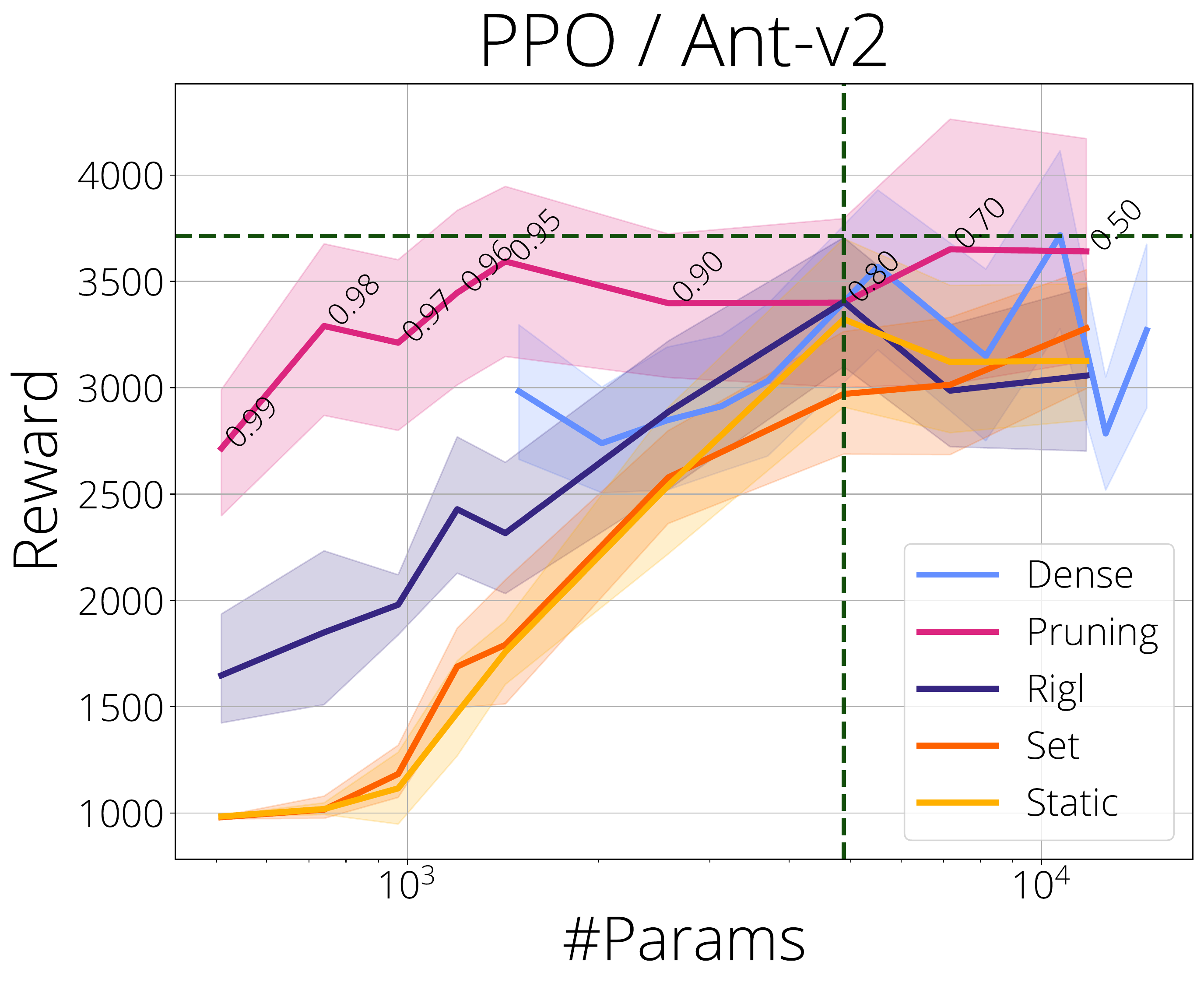} \\
    \caption{Additional MuJoCO environments (Hopper and Ant) for SAC and PPO algorithms. Networks are initialized with \gls{erk} network sparsity distribution.}
    \label{fig:mujoco-extra-app}
\end{figure}

\begin{figure}[H]
    \centering
    \includegraphics[width=0.31\textwidth]{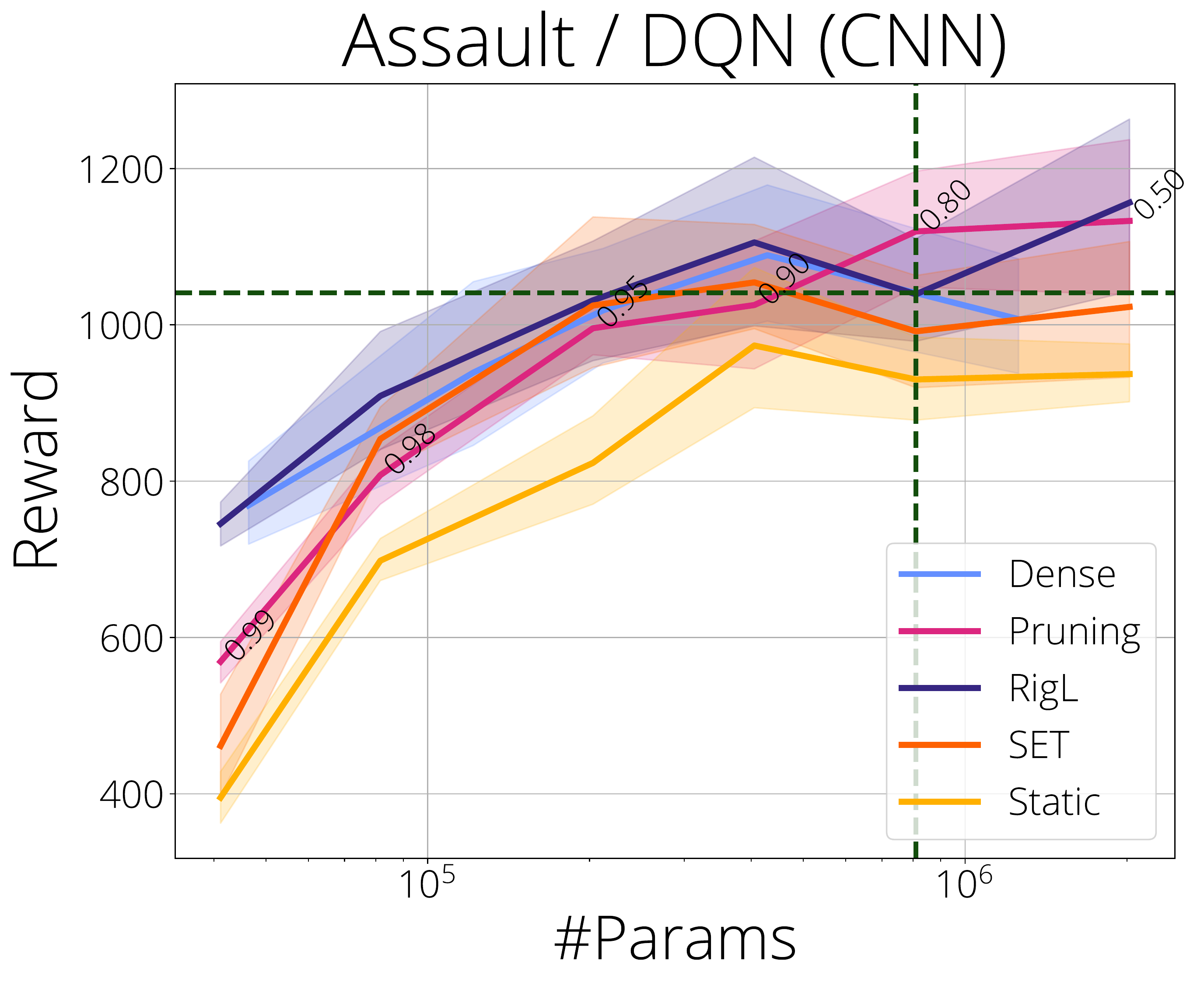}
    \includegraphics[width=0.31\textwidth]{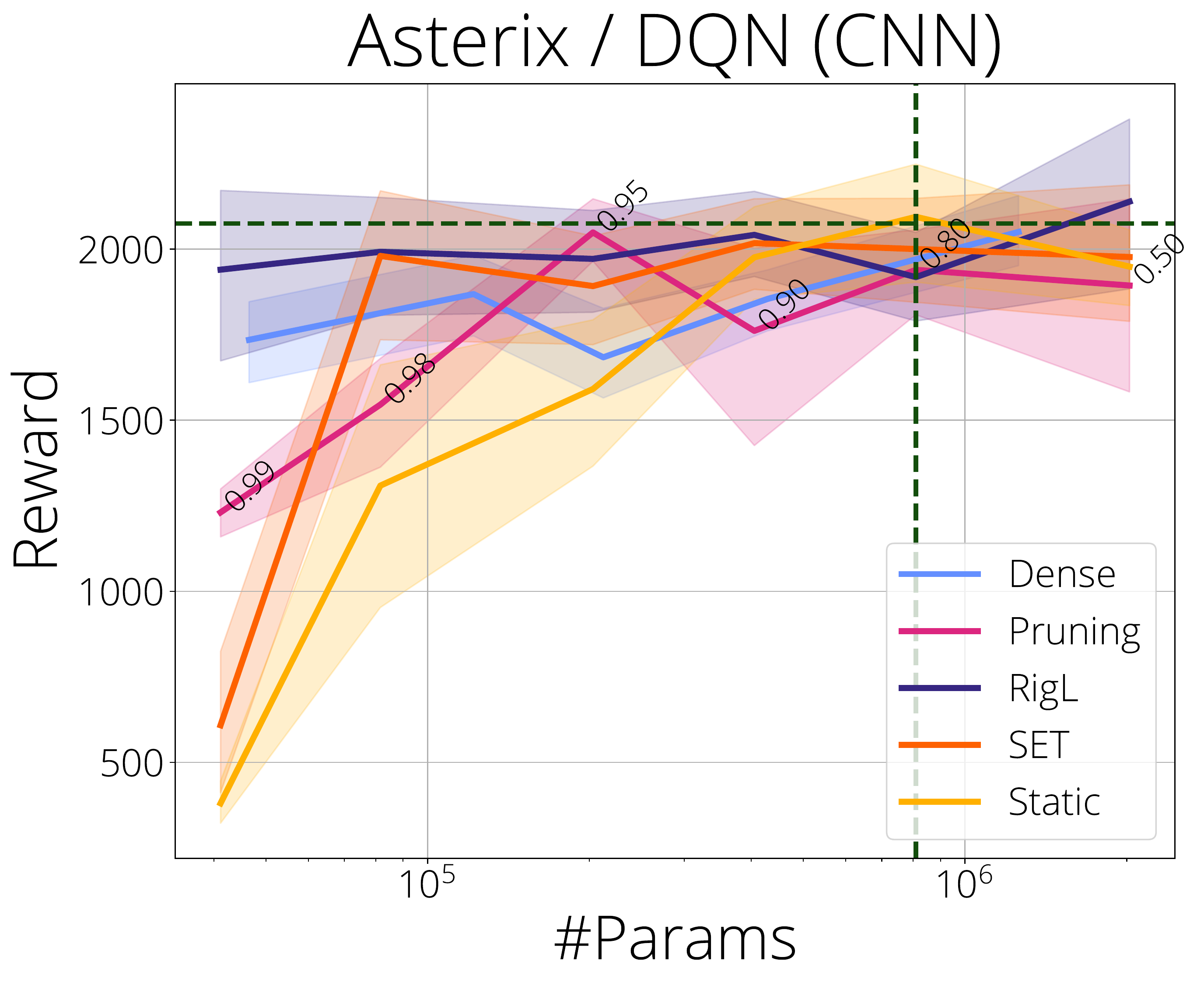}
    \includegraphics[width=0.31\textwidth]{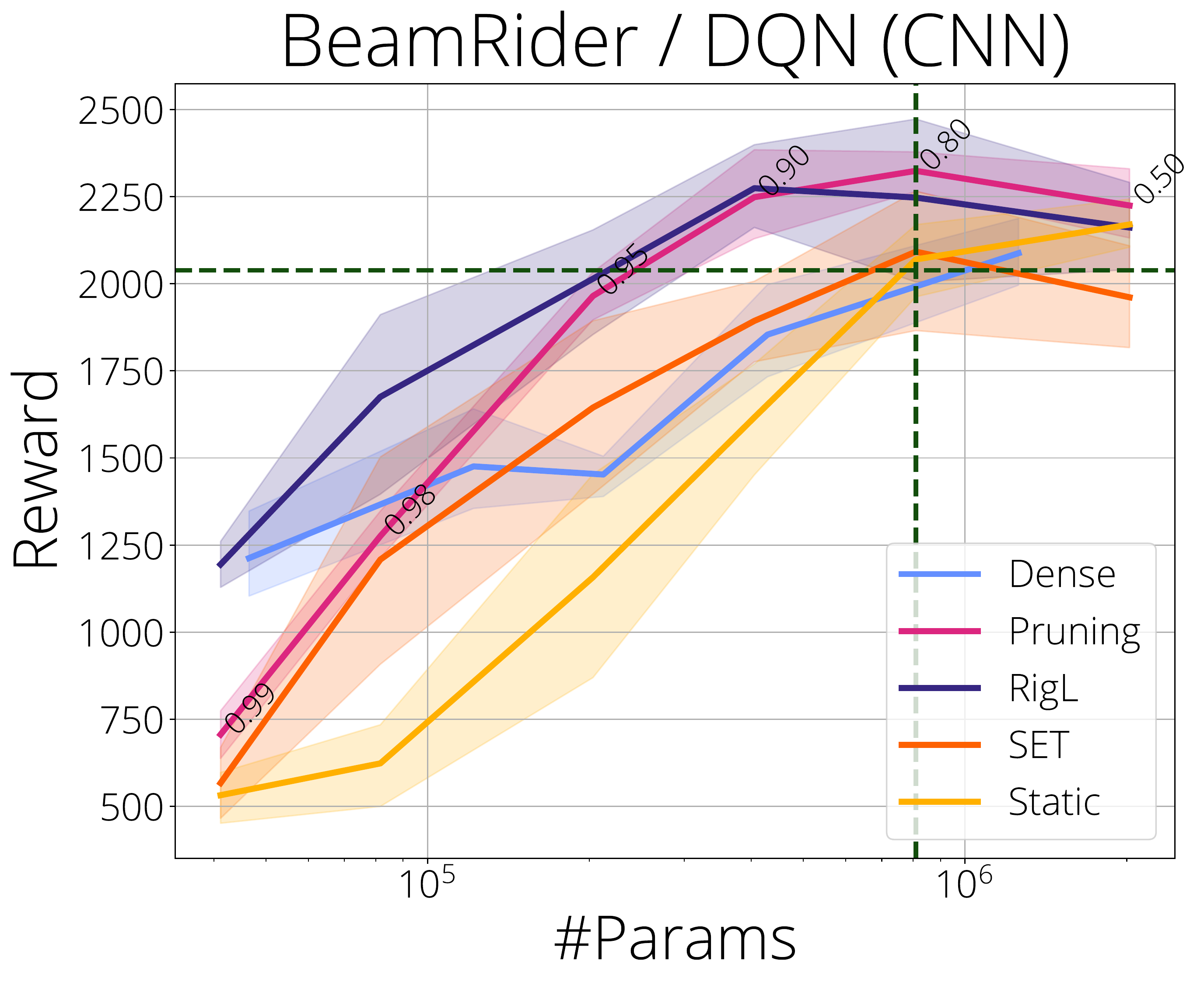}
    \includegraphics[width=0.31\textwidth]{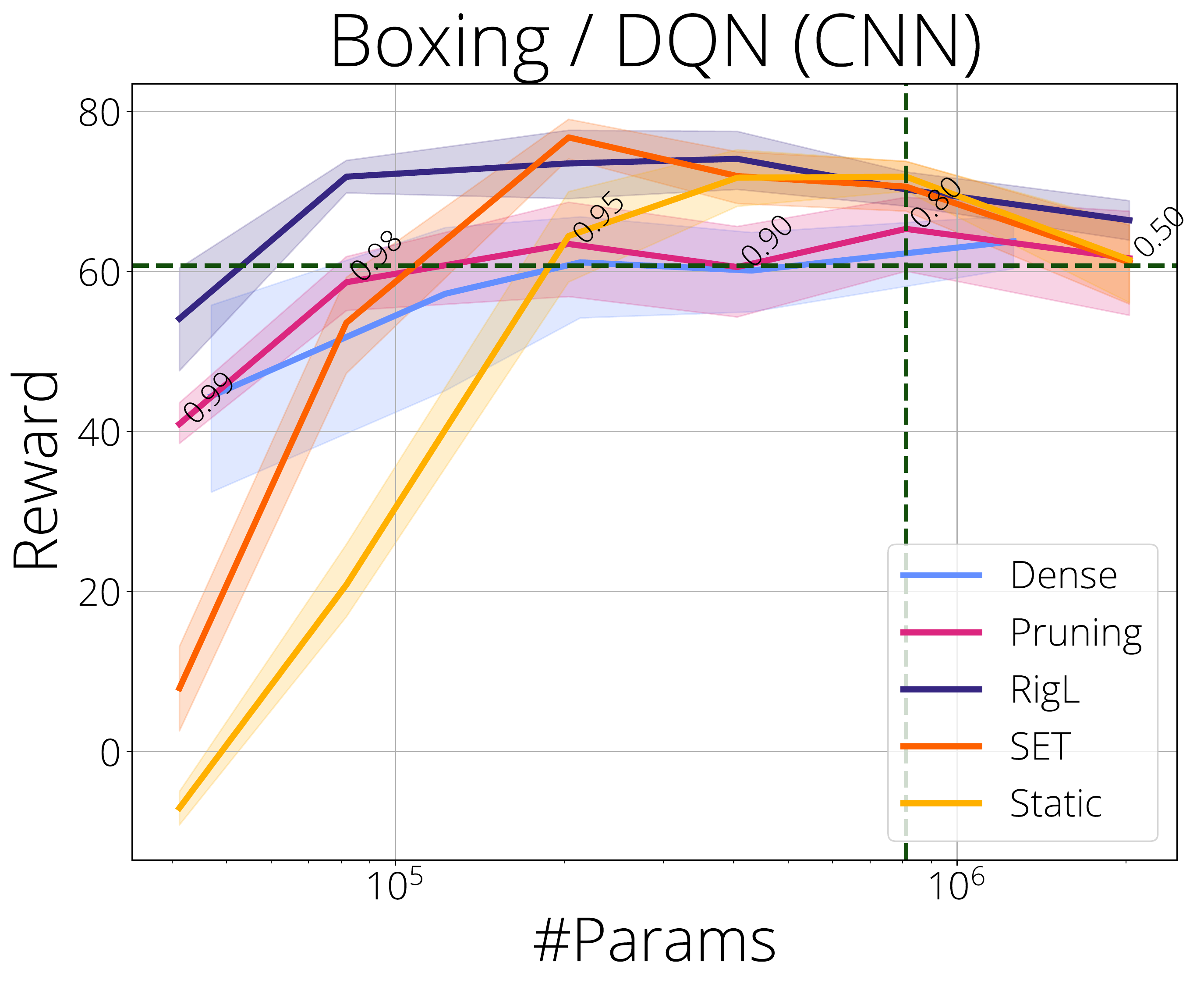}
    \includegraphics[width=0.31\textwidth]{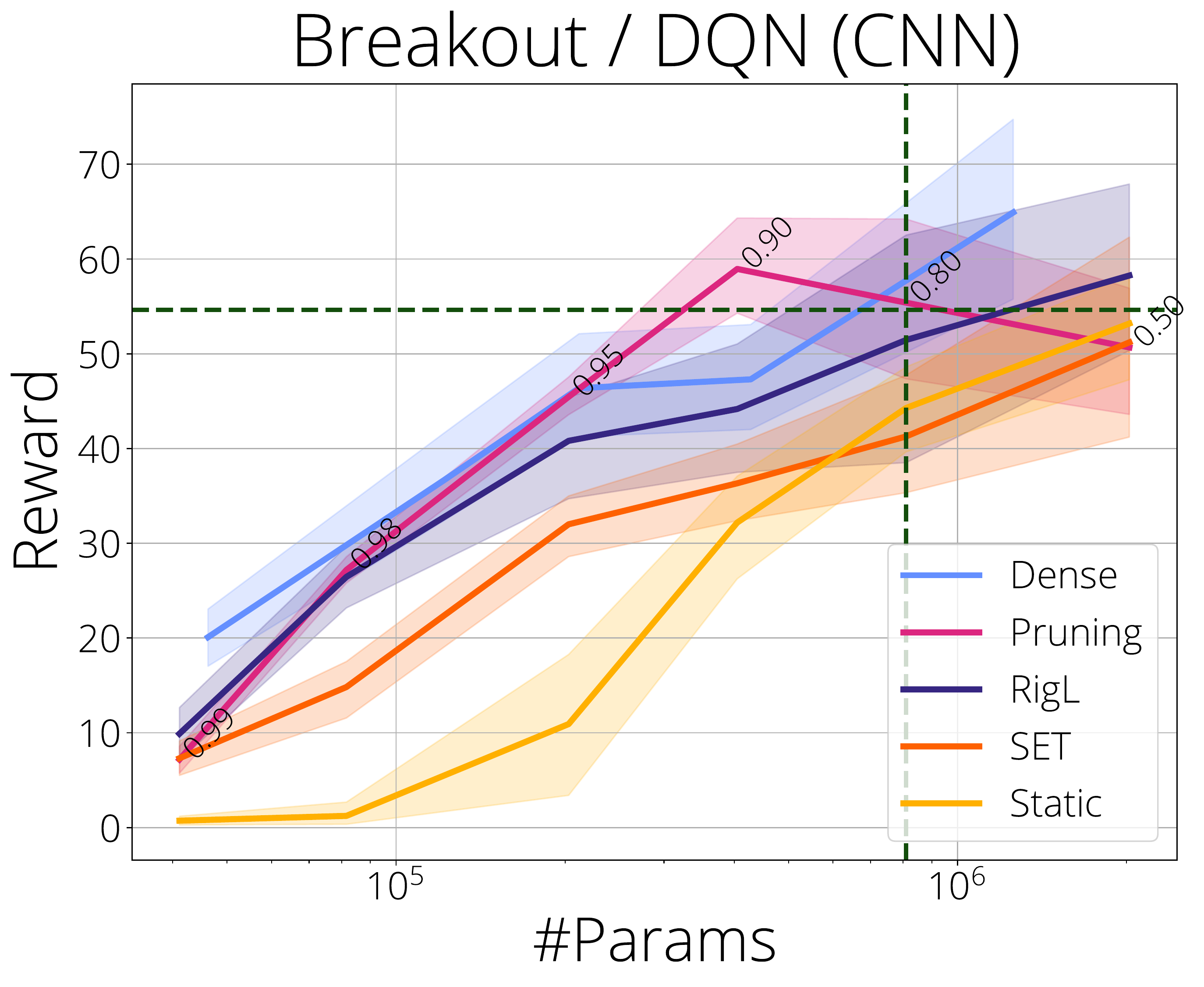}
    \includegraphics[width=0.31\textwidth]{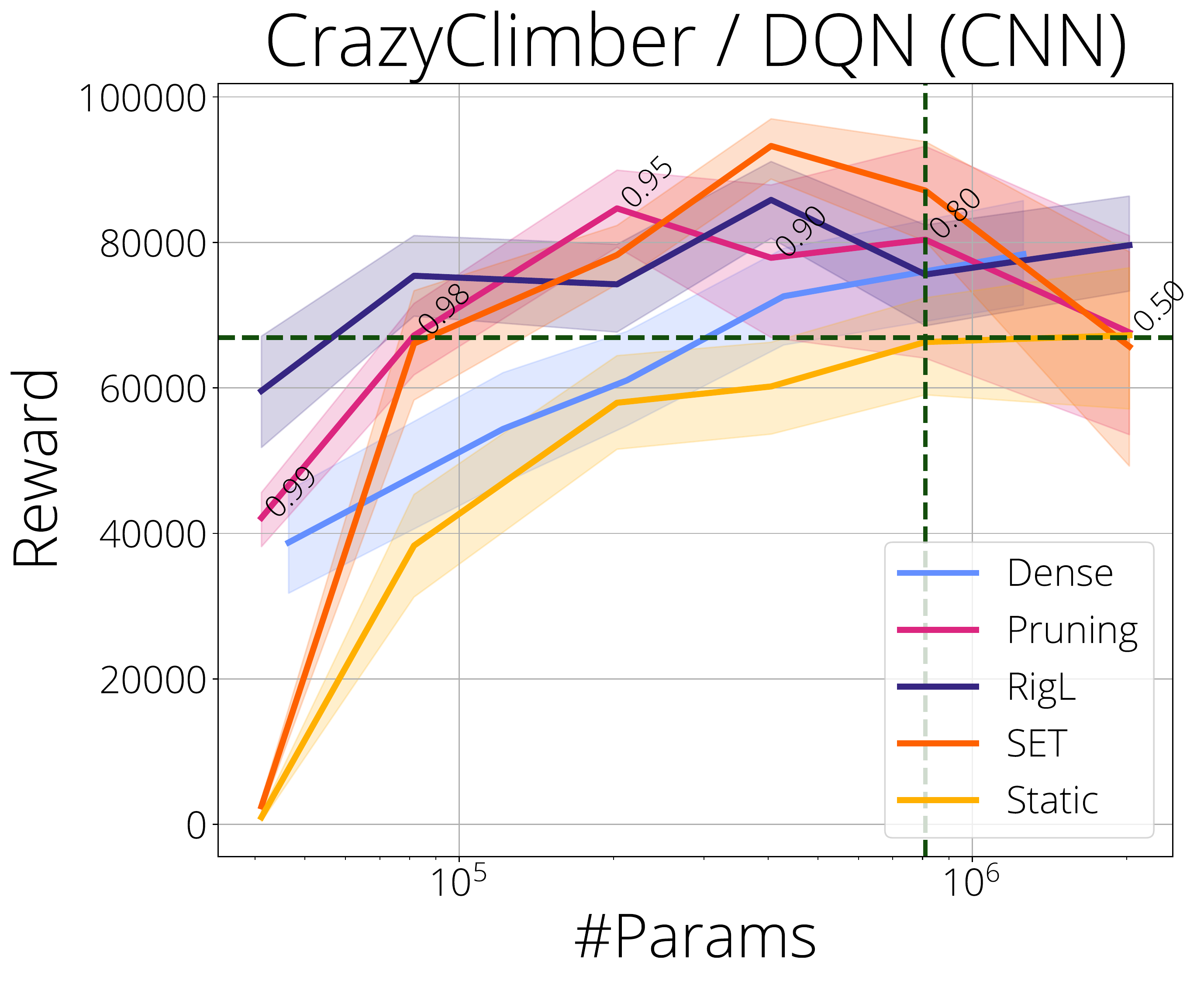}
    \includegraphics[width=0.31\textwidth]{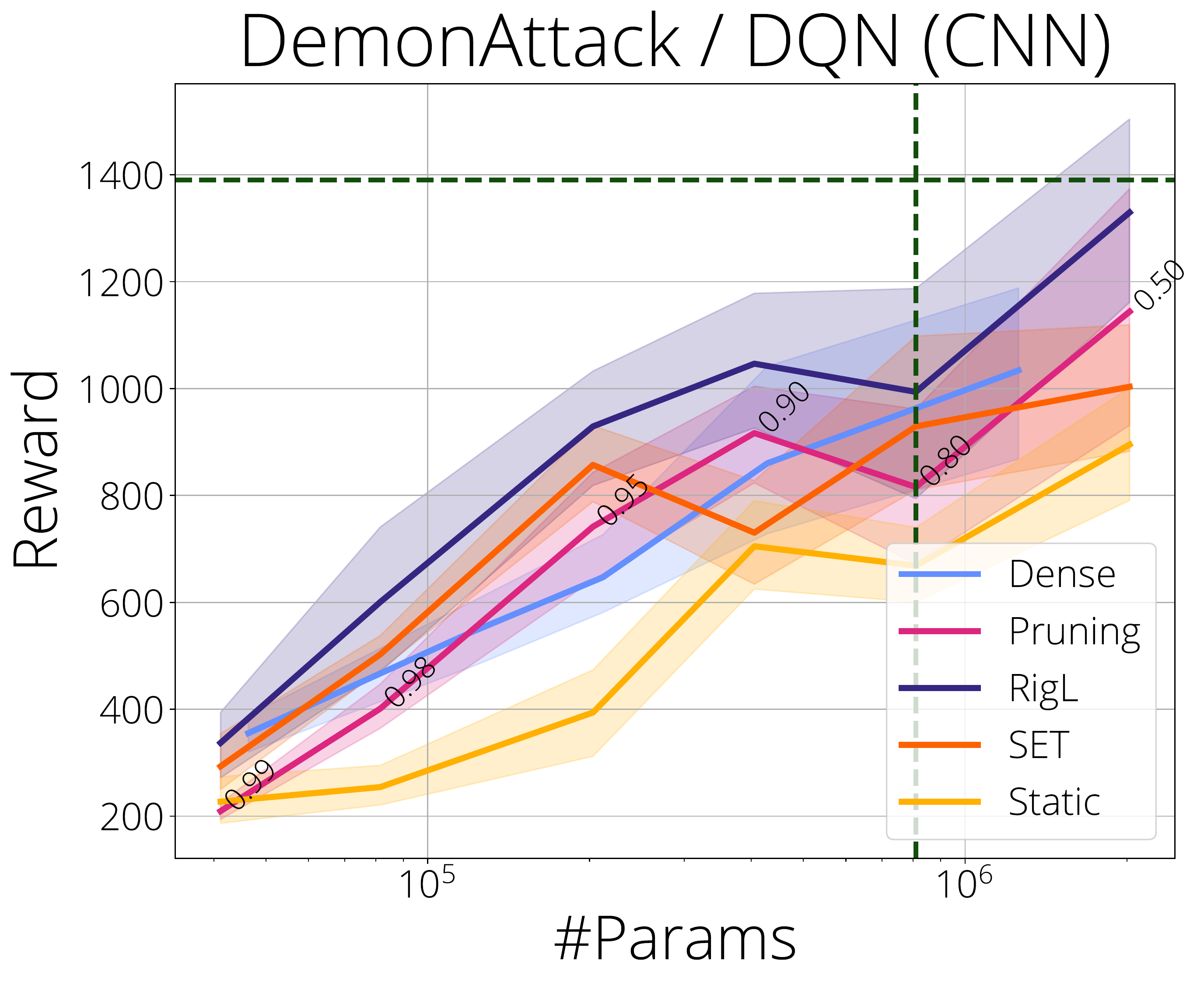}
    \includegraphics[width=0.31\textwidth]{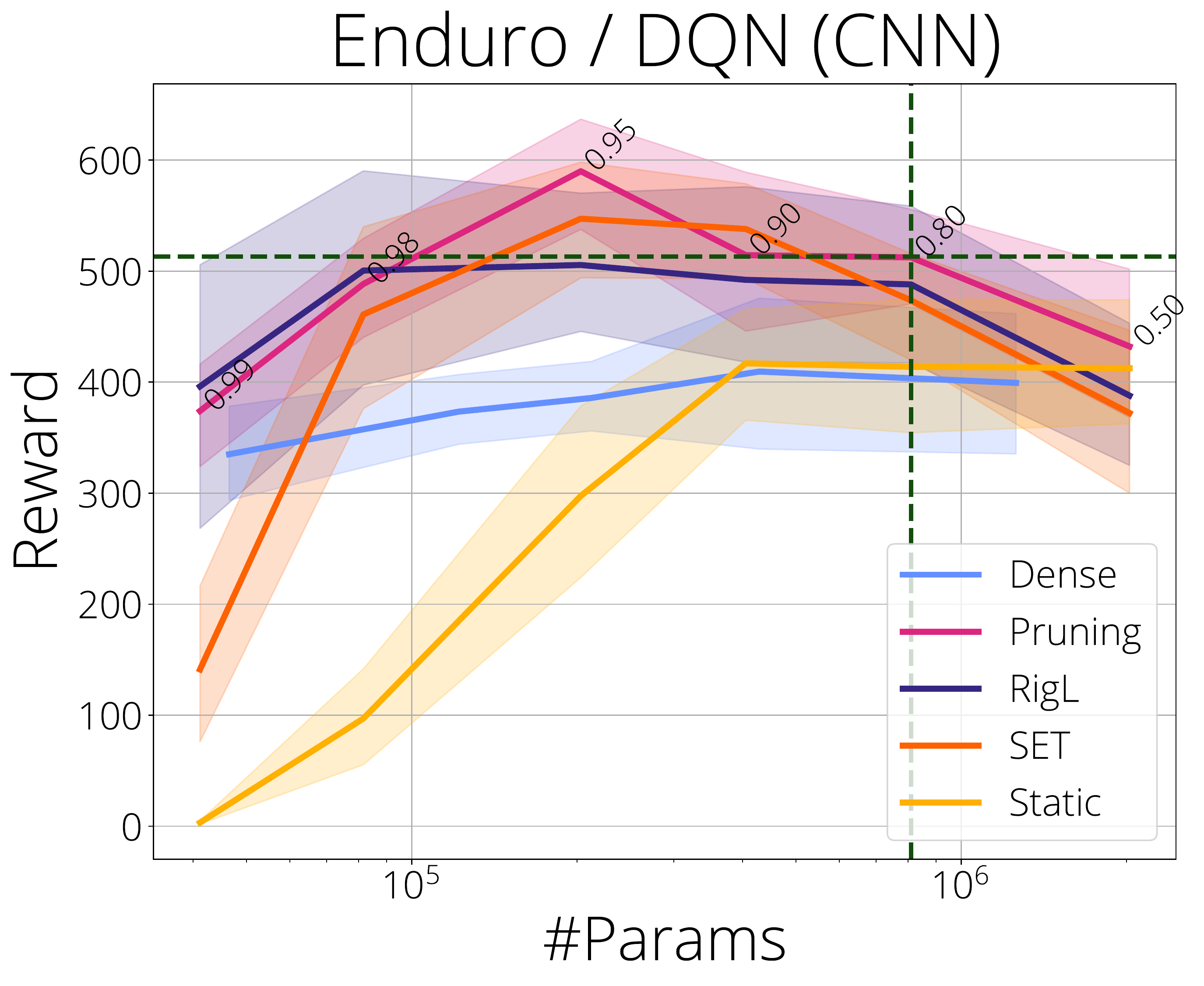}
    \includegraphics[width=0.31\textwidth]{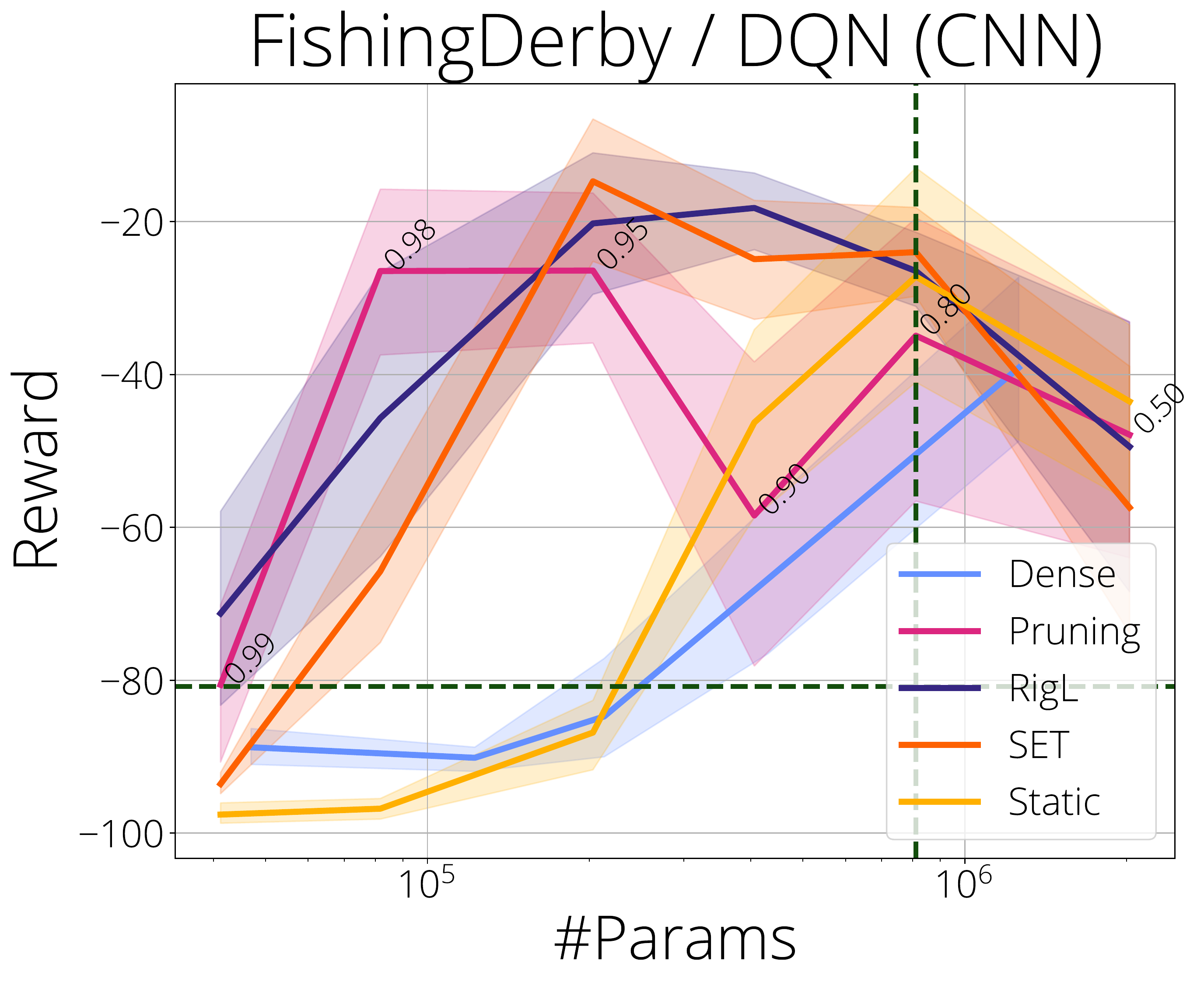}
    \includegraphics[width=0.31\textwidth]{figs/main_dqn_MsPacman_erk.pdf}
    \includegraphics[width=0.31\textwidth]{figs/main_dqn_Pong_erk.pdf}
    \includegraphics[width=0.31\textwidth]{figs/main_dqn_Qbert_erk.pdf}
    \includegraphics[width=0.31\textwidth]{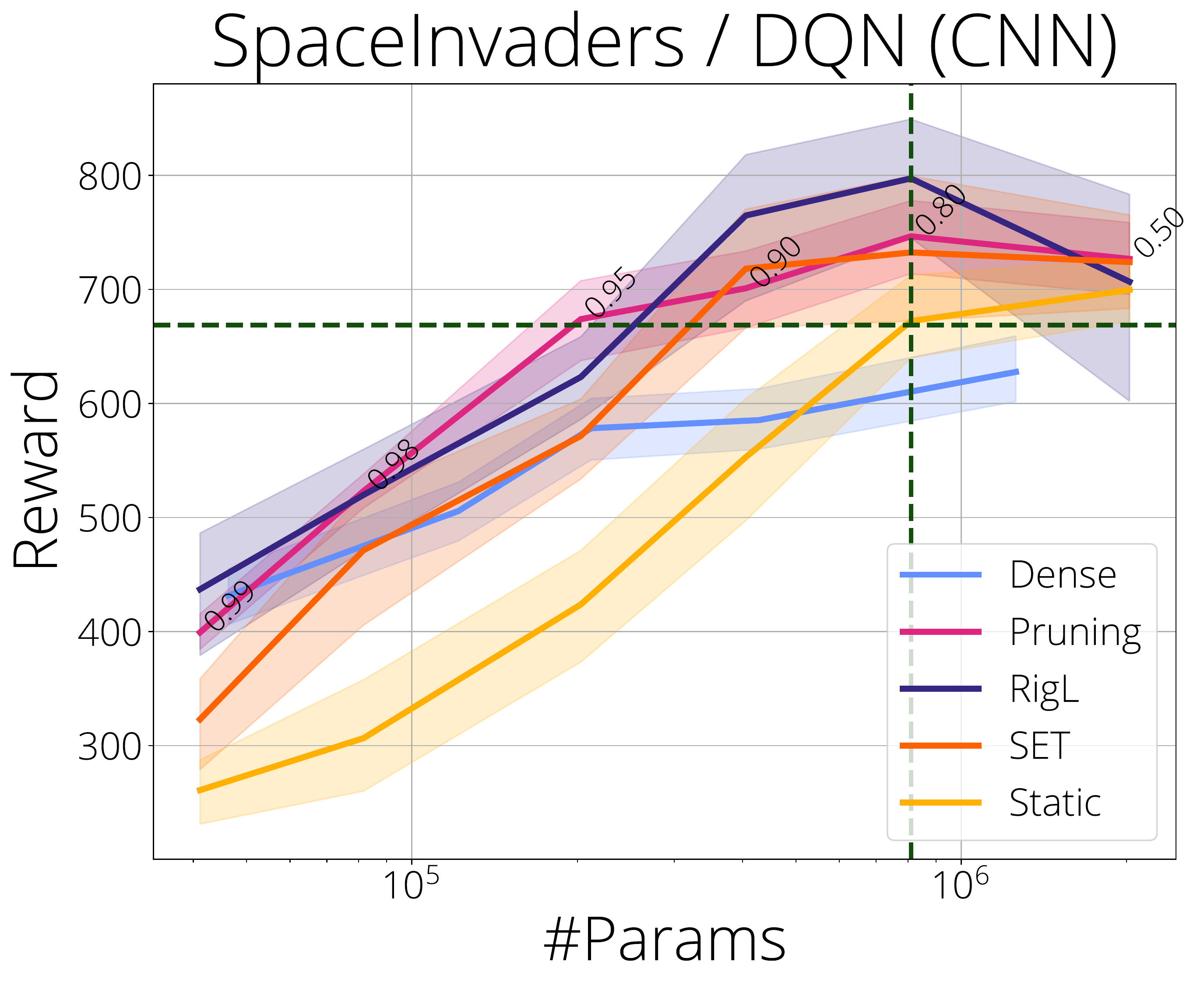}
    \includegraphics[width=0.31\textwidth]{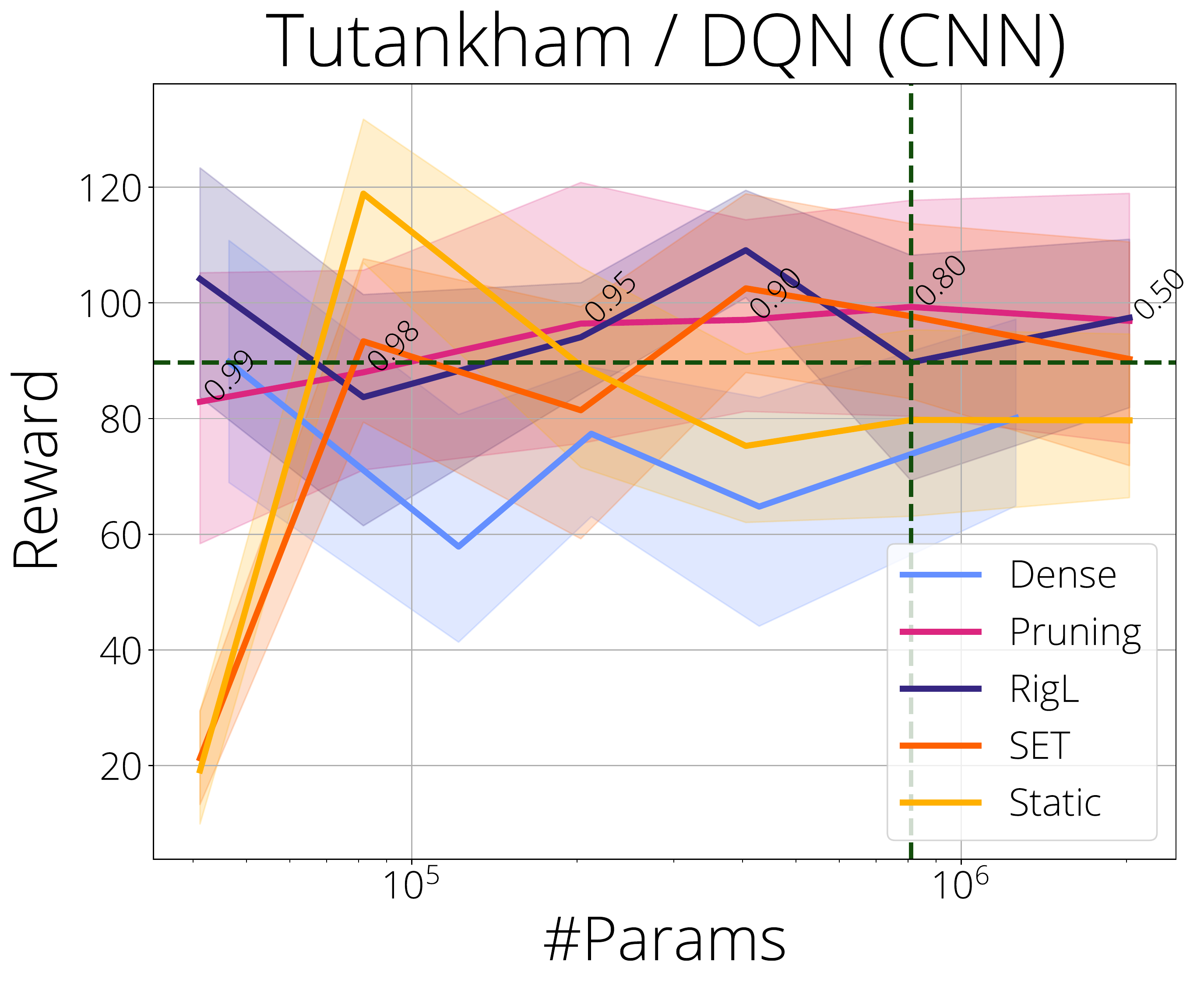}
    \includegraphics[width=0.31\textwidth]{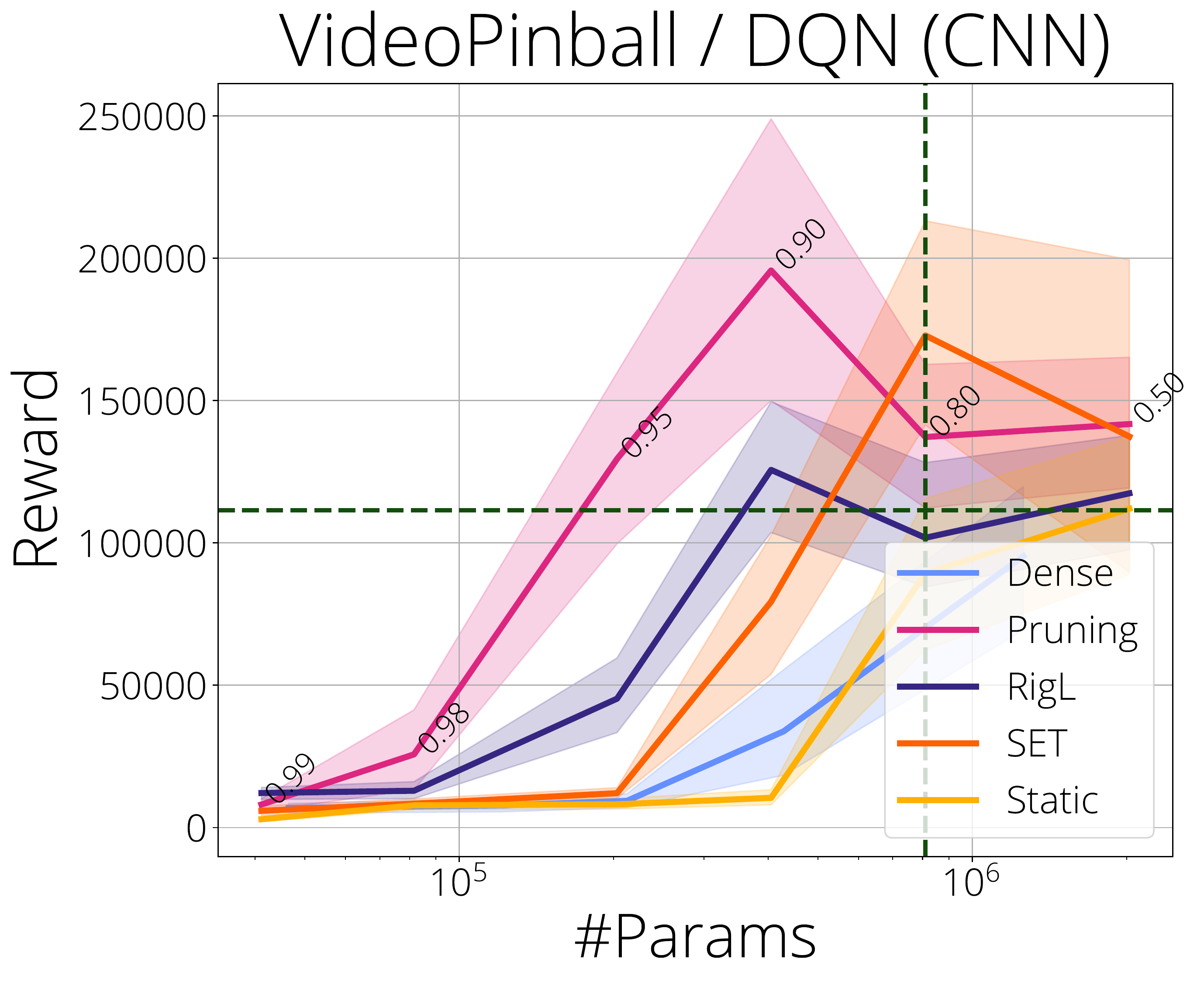}
    \caption{CNN: Sparsity plots per game.}
    \label{fig:atari-cnn}
\end{figure}

\begin{figure}[H]
    \centering
    \includegraphics[width=0.31\textwidth]{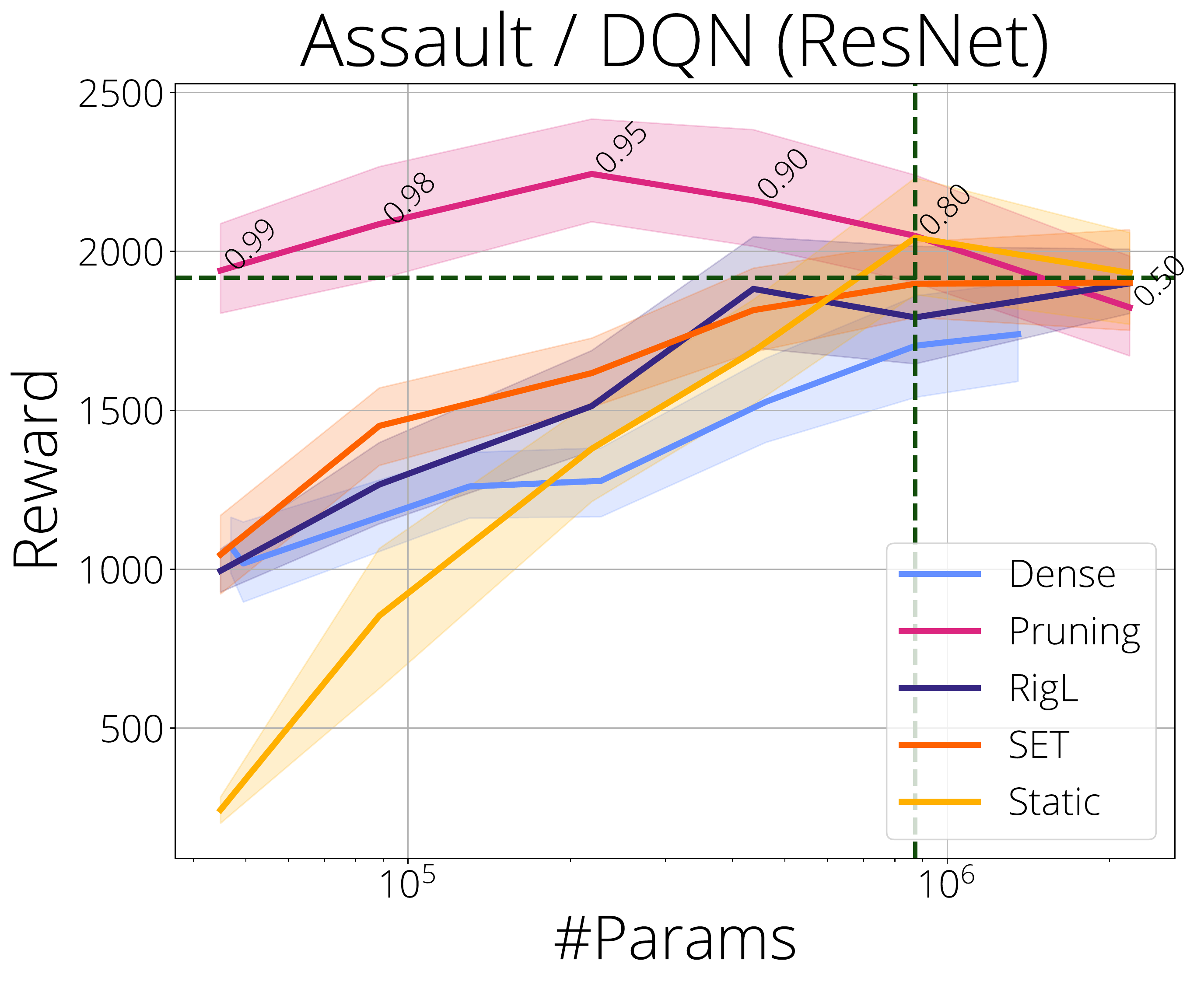}
    \includegraphics[width=0.31\textwidth]{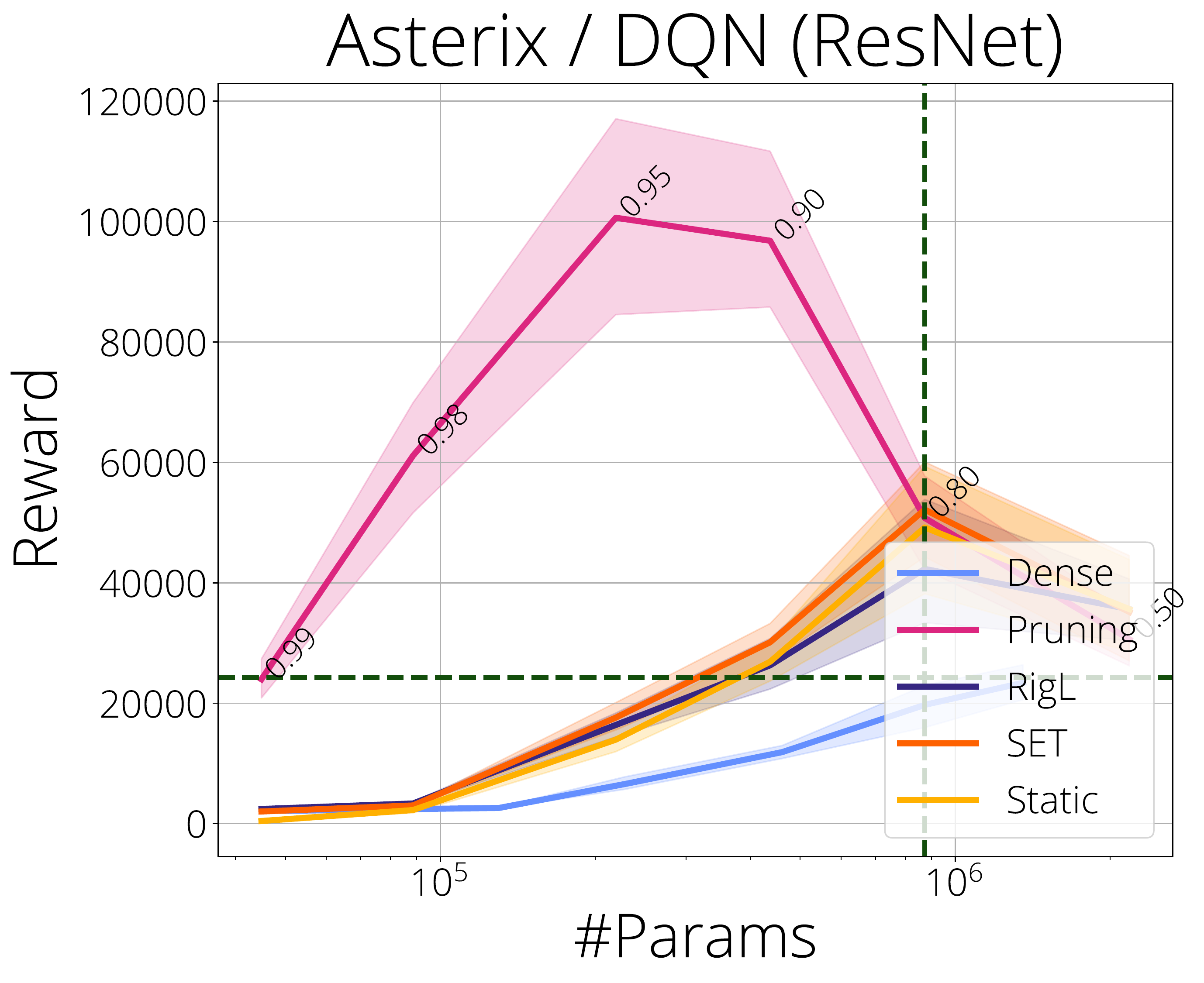}
    \includegraphics[width=0.31\textwidth]{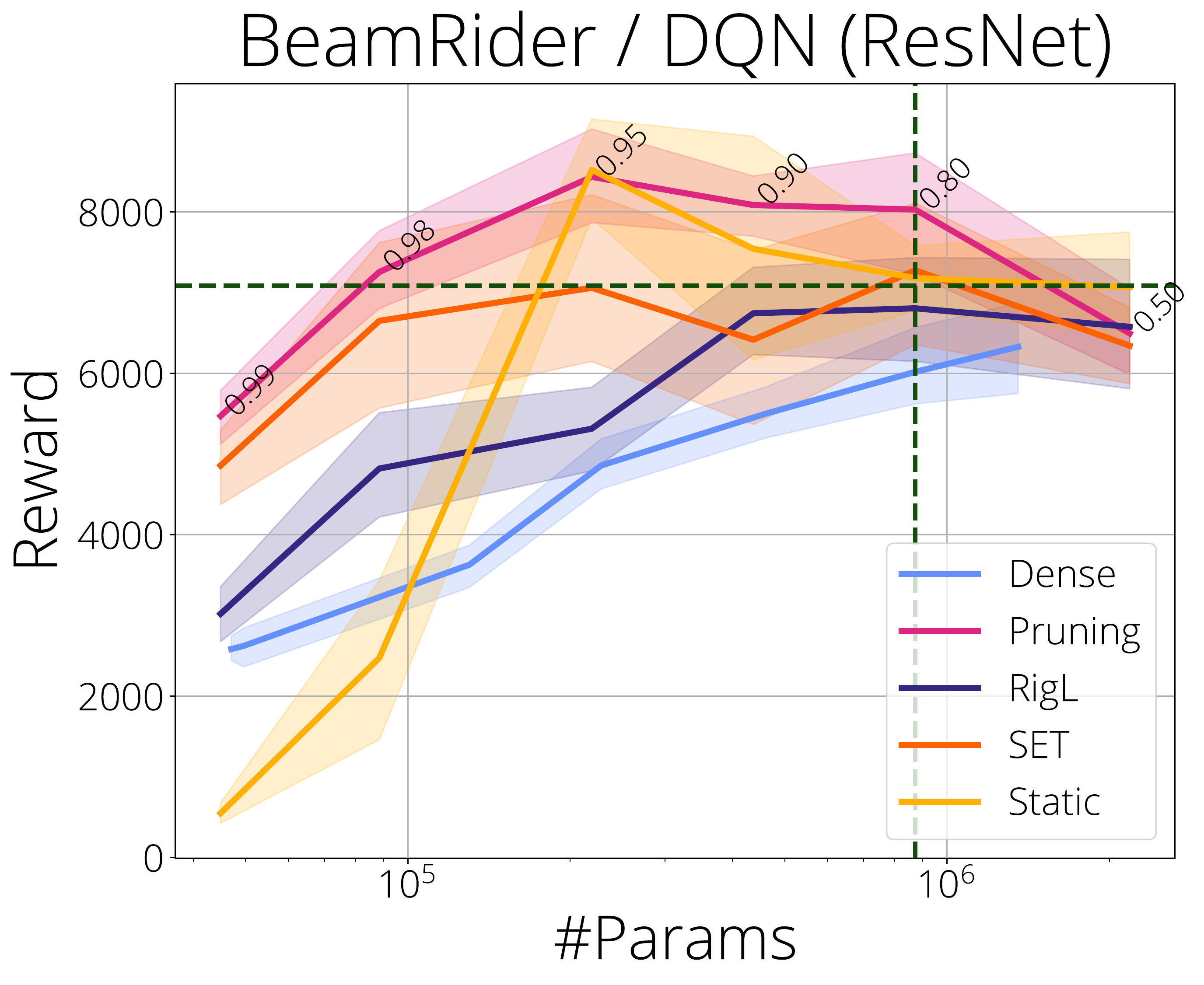}
    \includegraphics[width=0.31\textwidth]{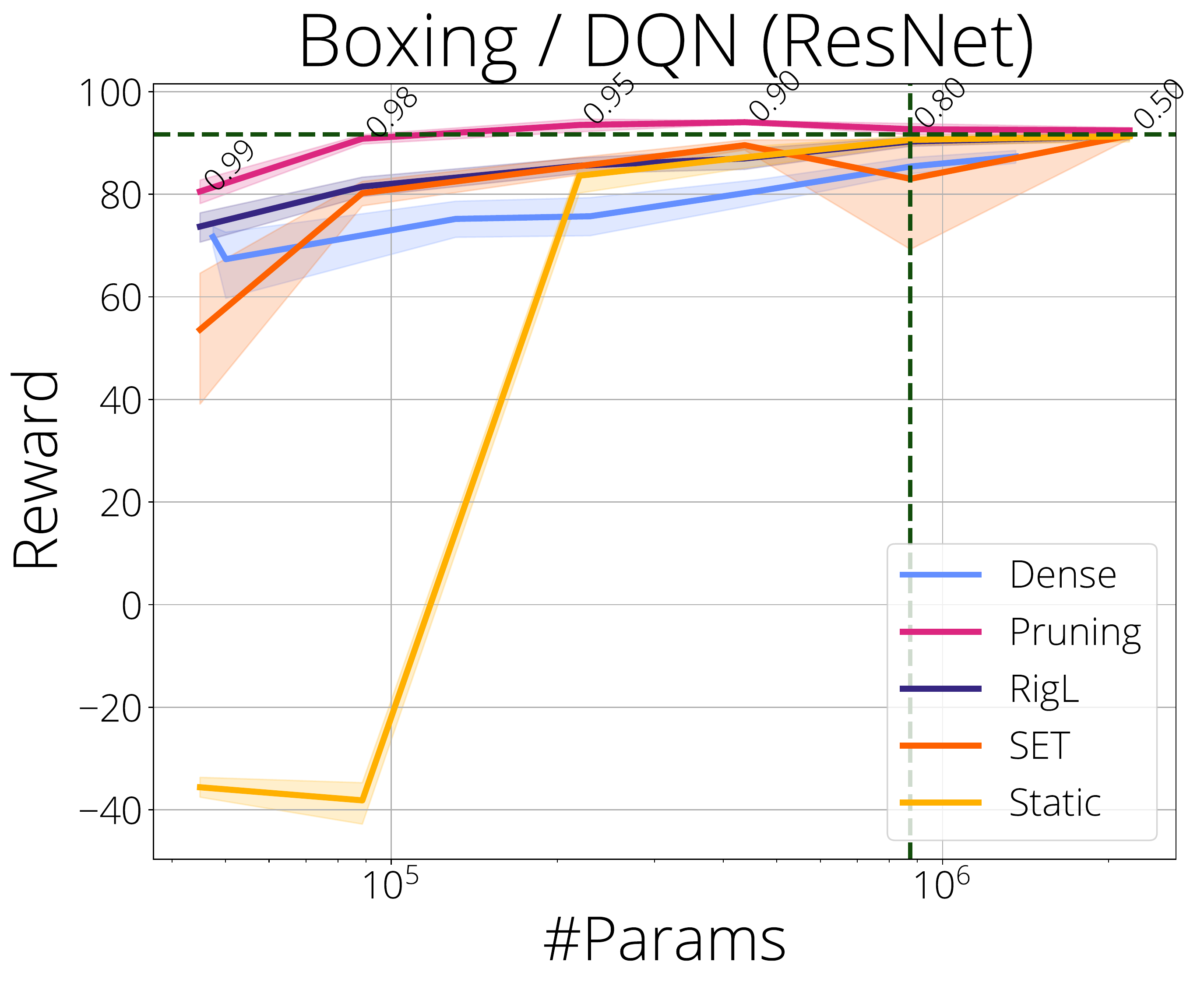}
    \includegraphics[width=0.31\textwidth]{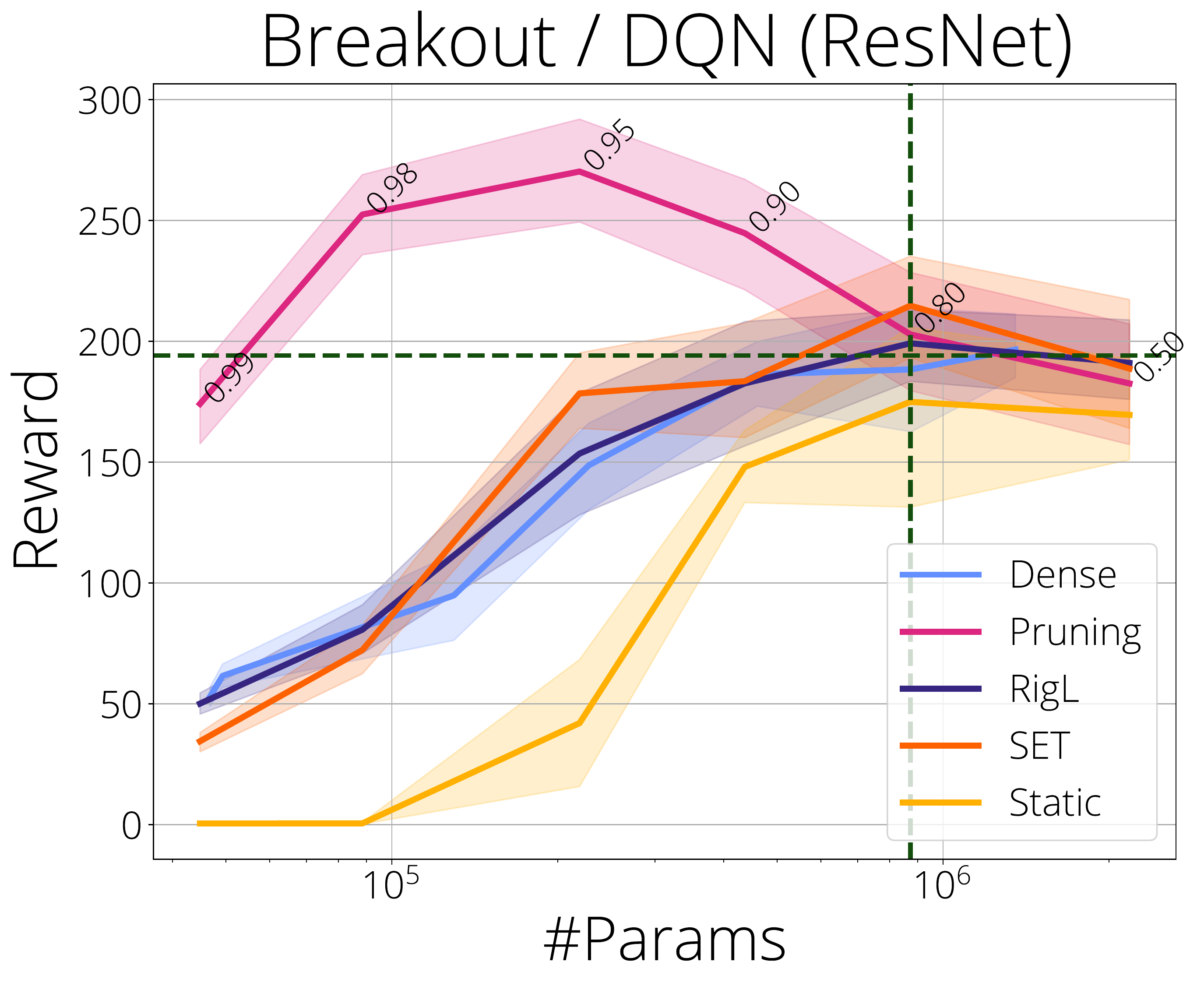}
    \includegraphics[width=0.31\textwidth]{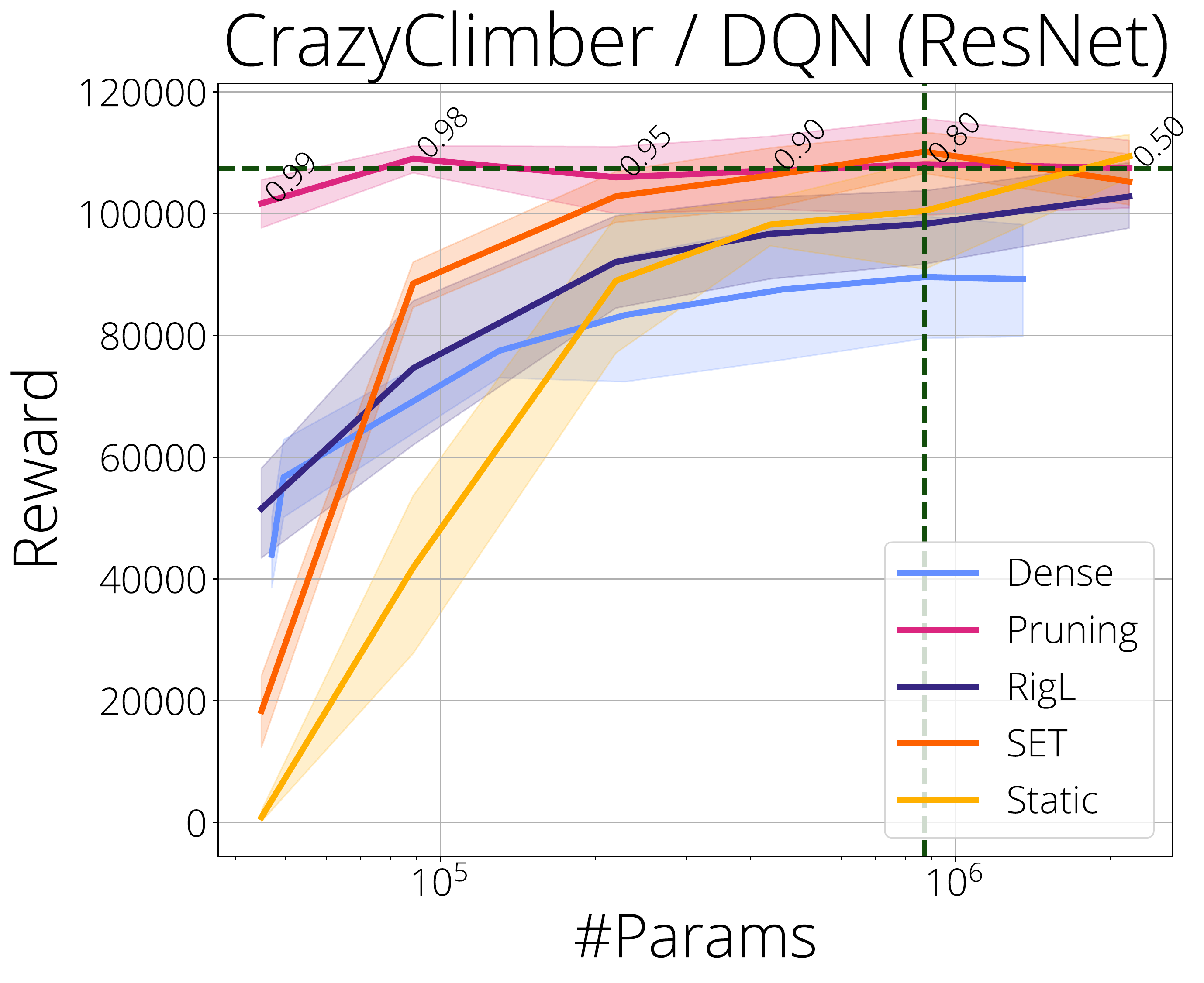}
    \includegraphics[width=0.31\textwidth]{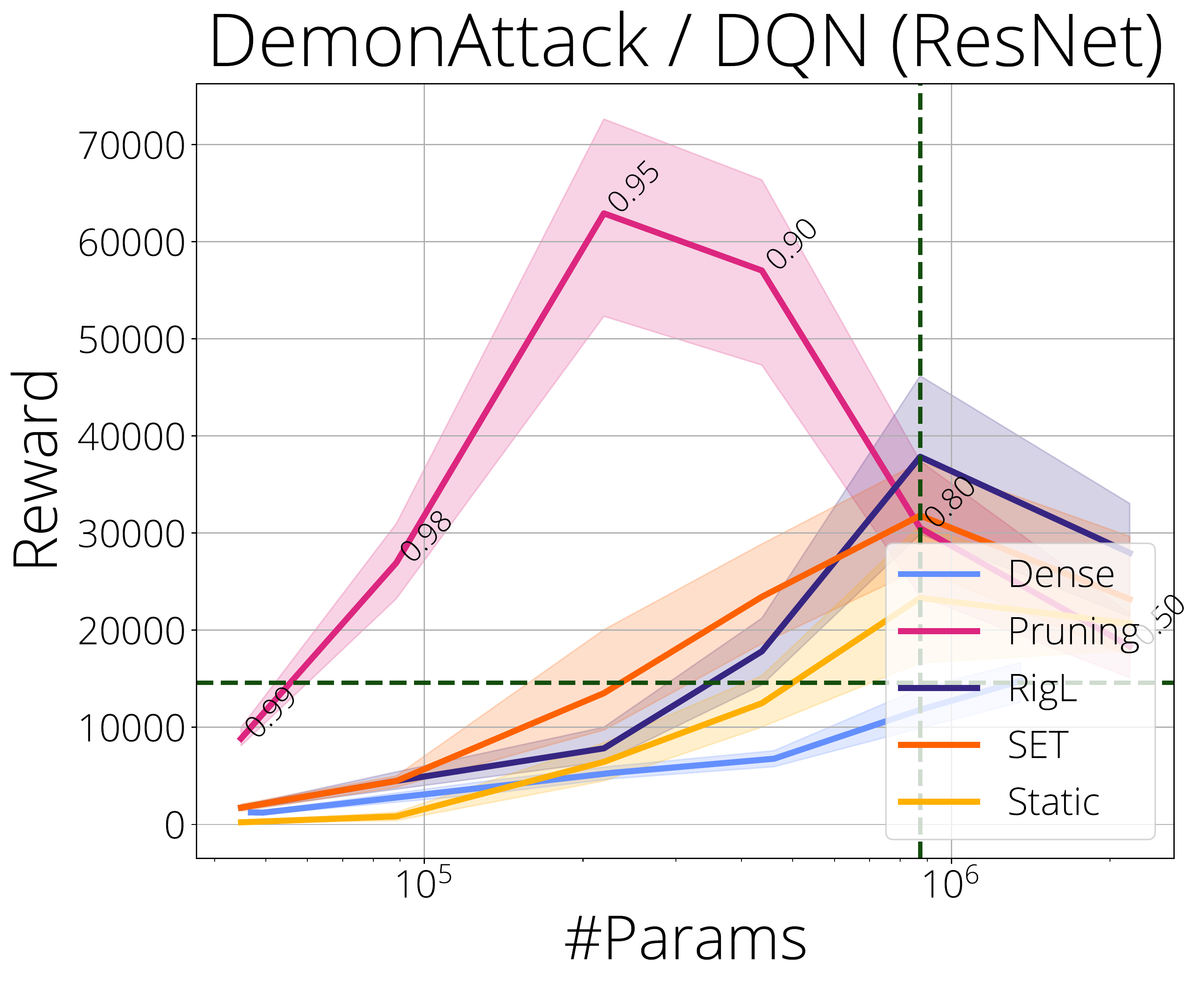}
    \includegraphics[width=0.31\textwidth]{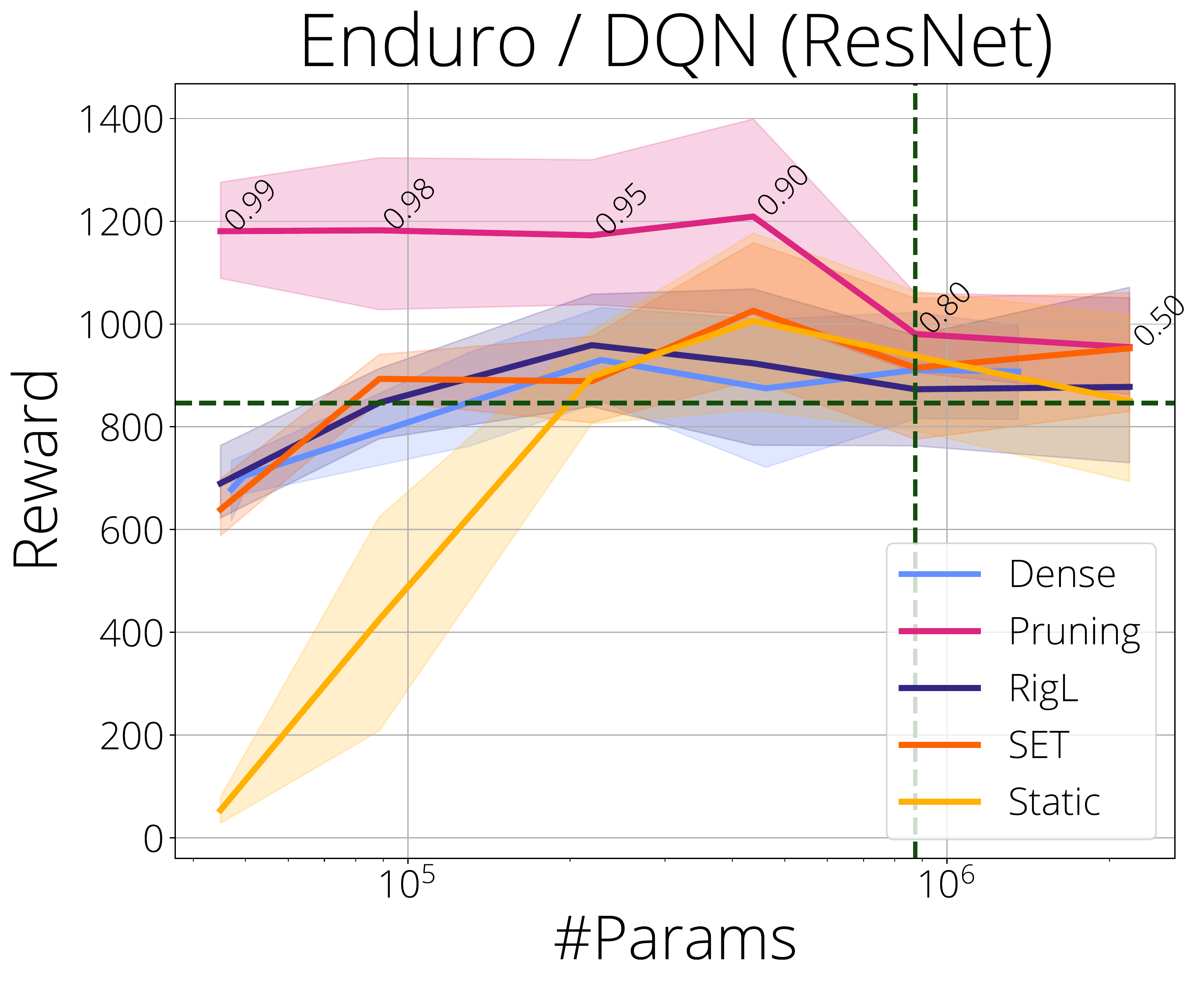}
    \includegraphics[width=0.31\textwidth]{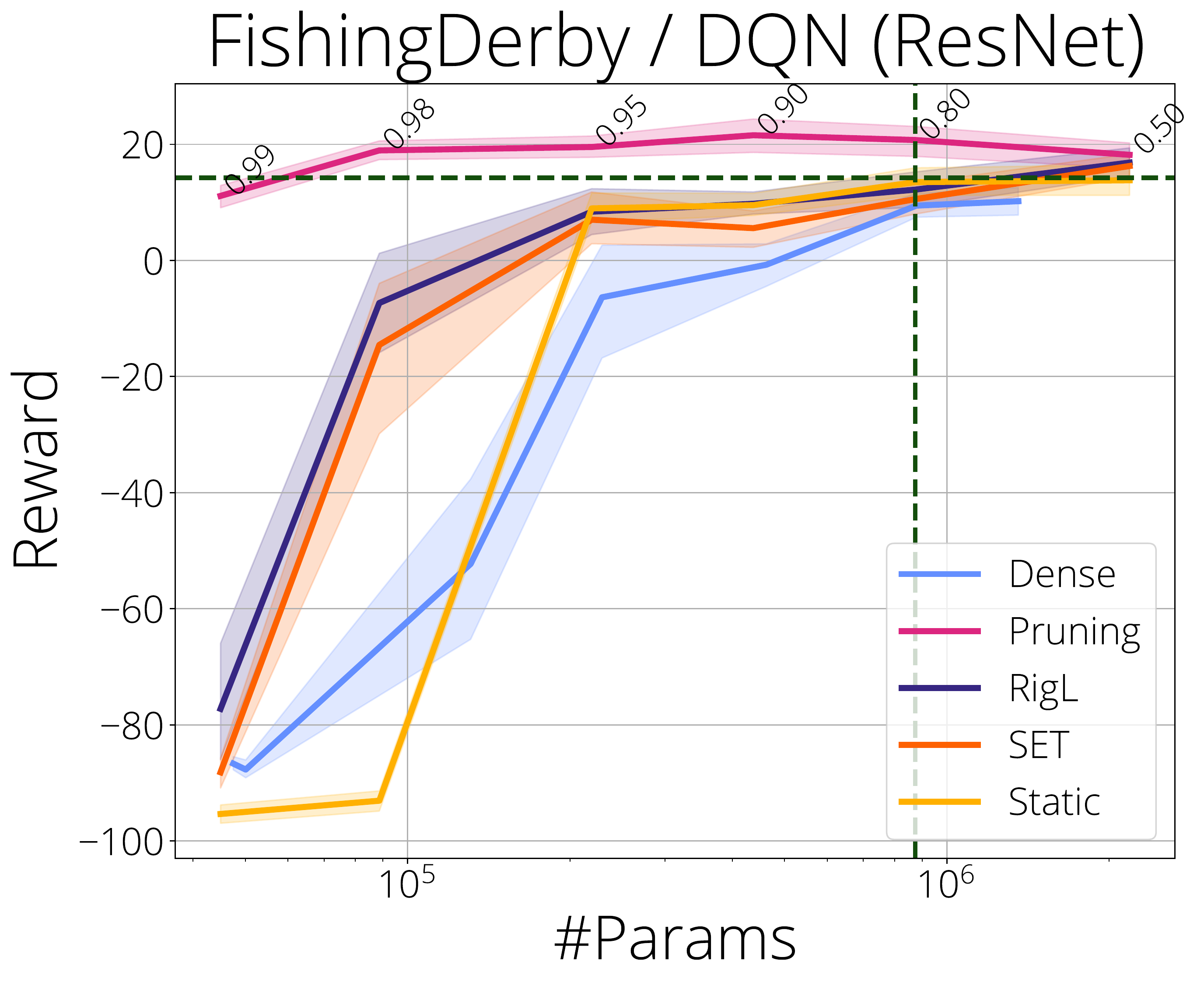}
    \includegraphics[width=0.31\textwidth]{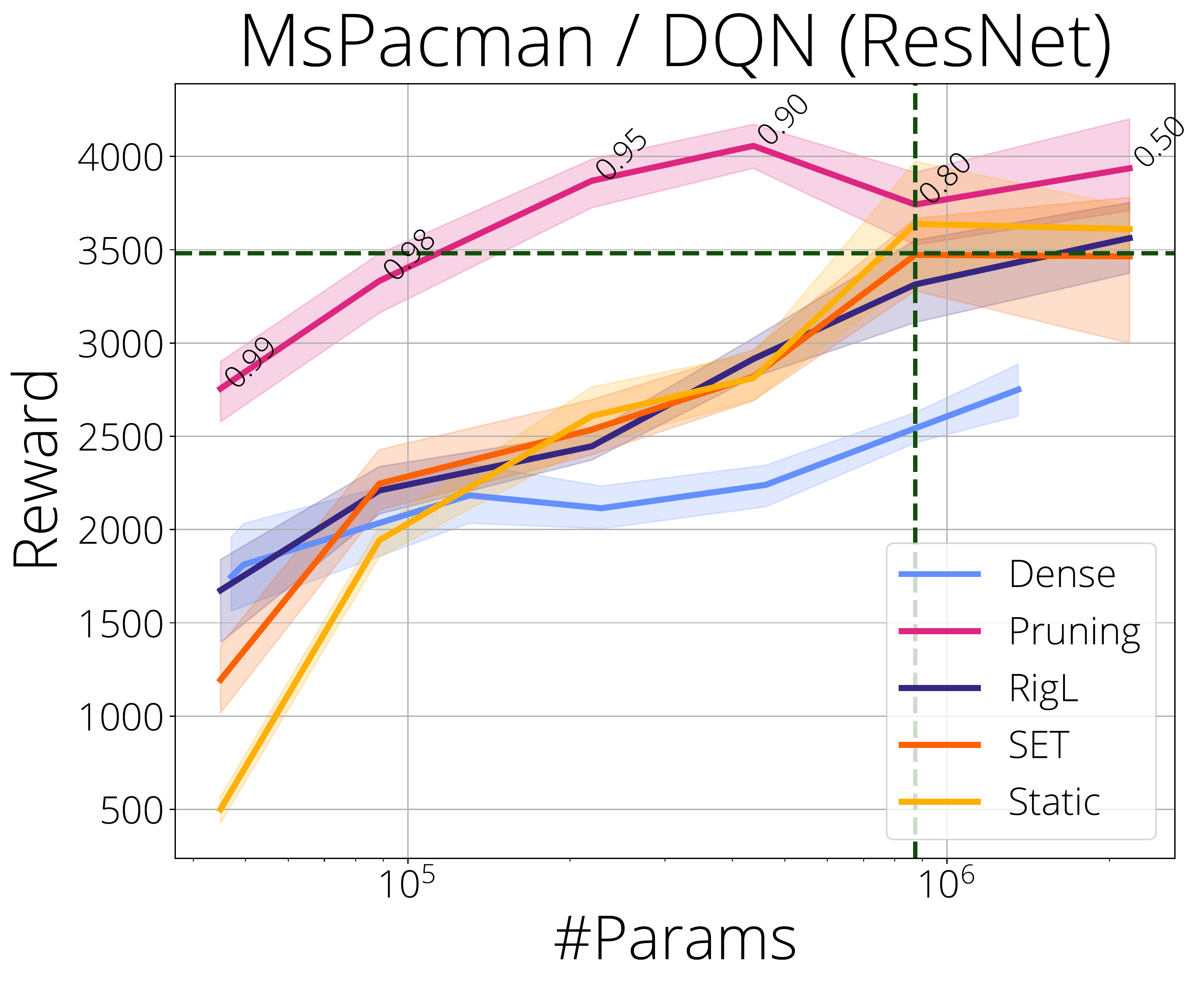}
    \includegraphics[width=0.31\textwidth]{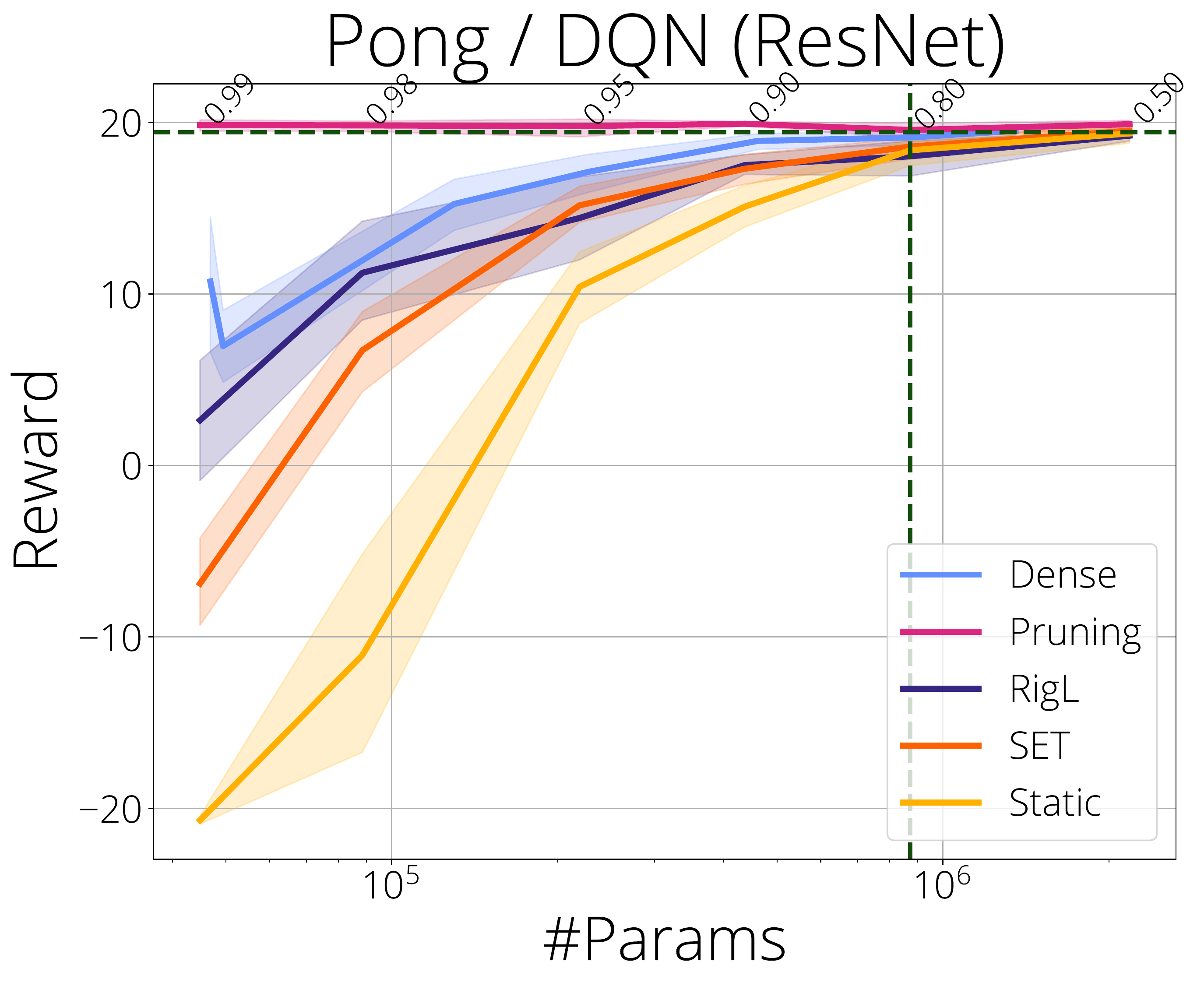}
    \includegraphics[width=0.31\textwidth]{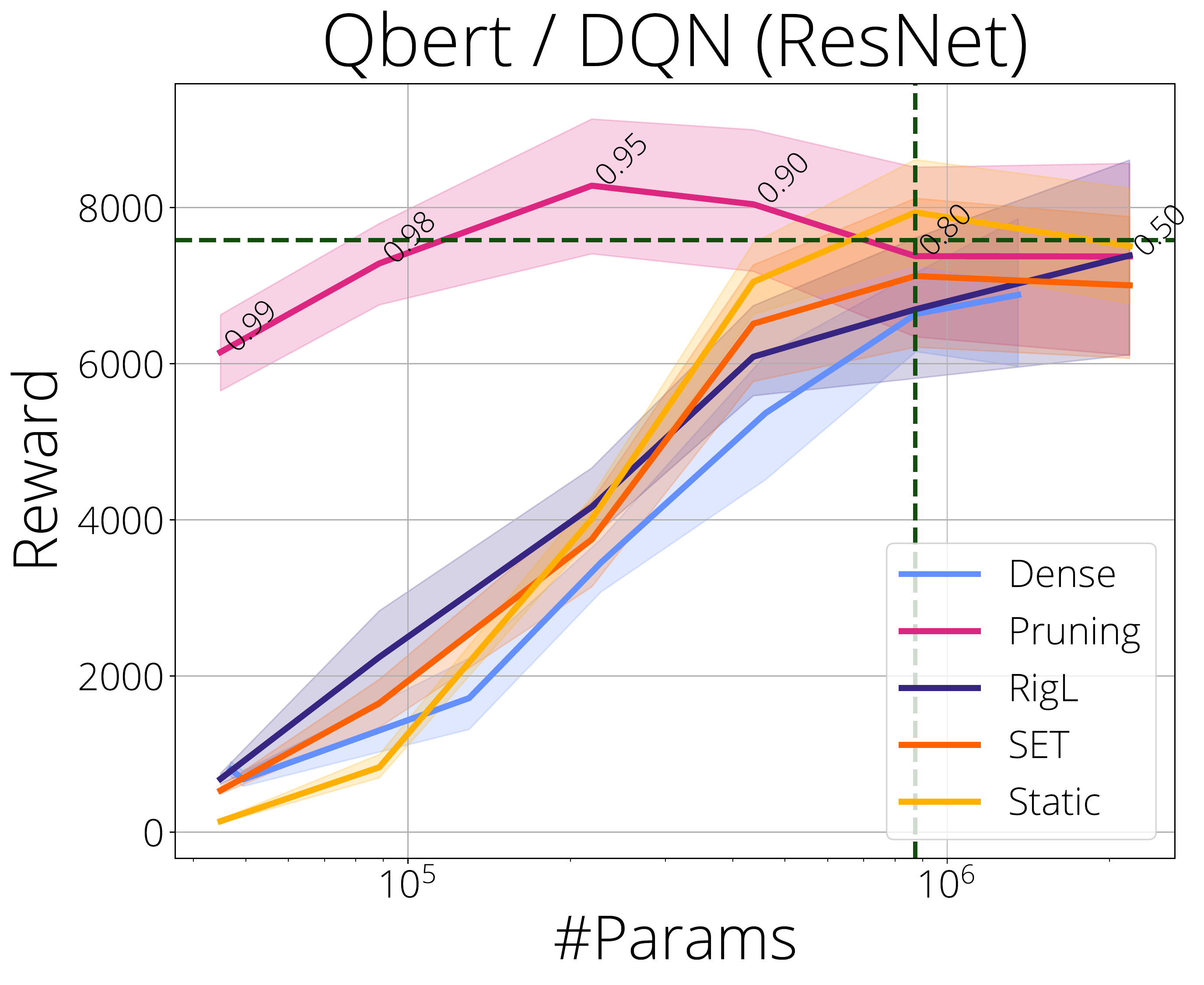}
    \includegraphics[width=0.31\textwidth]{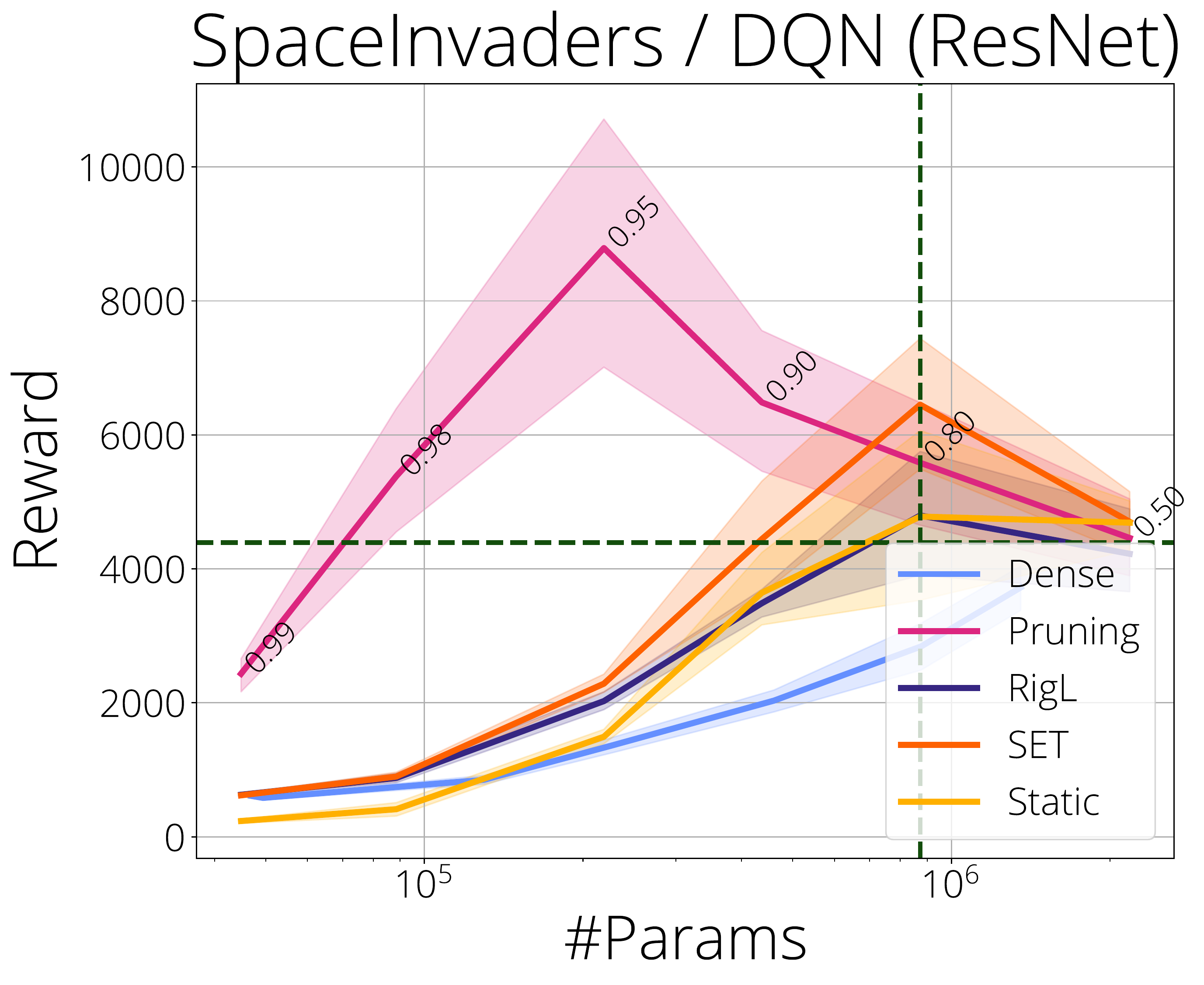}
    \includegraphics[width=0.31\textwidth]{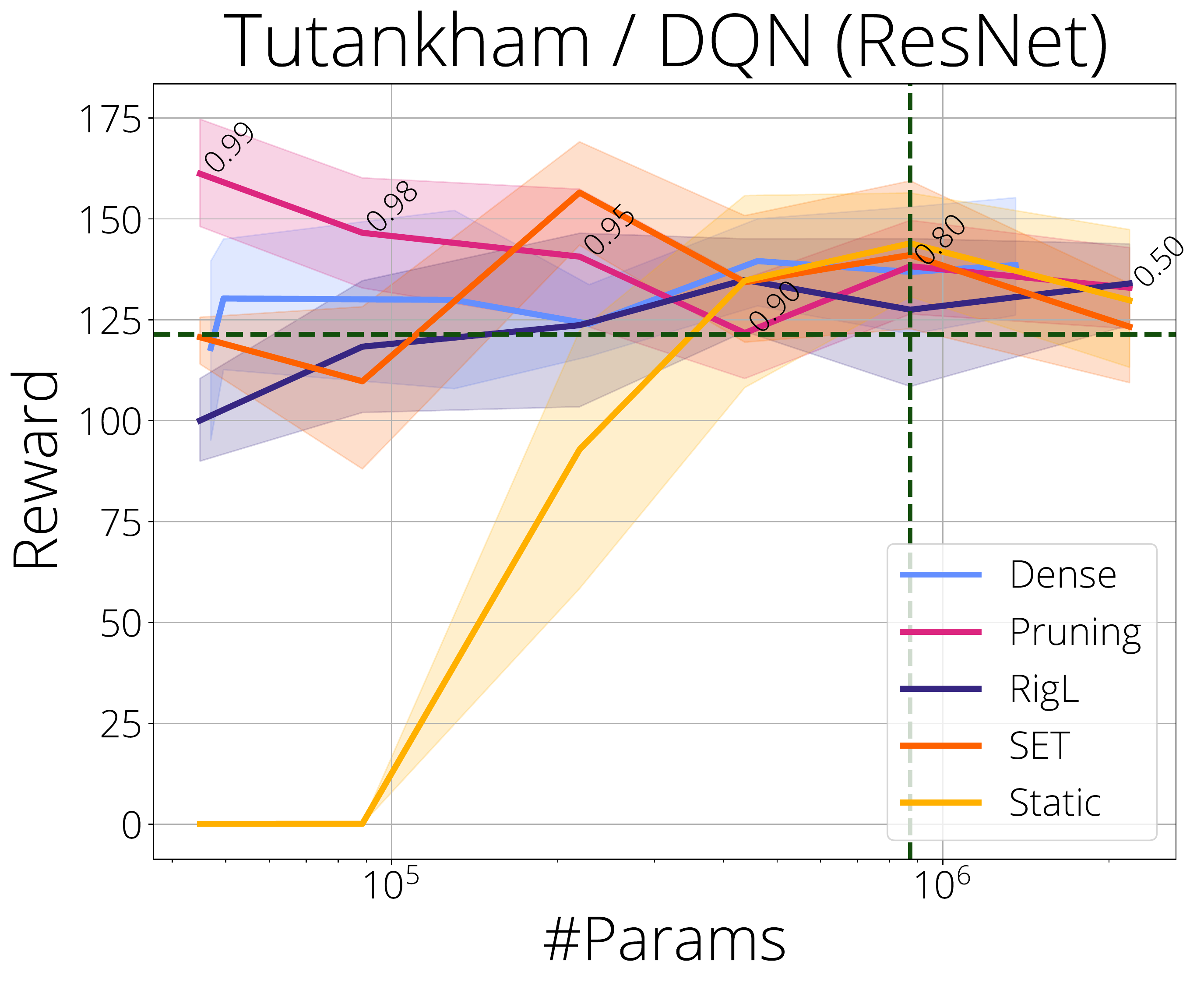}
    \includegraphics[width=0.31\textwidth]{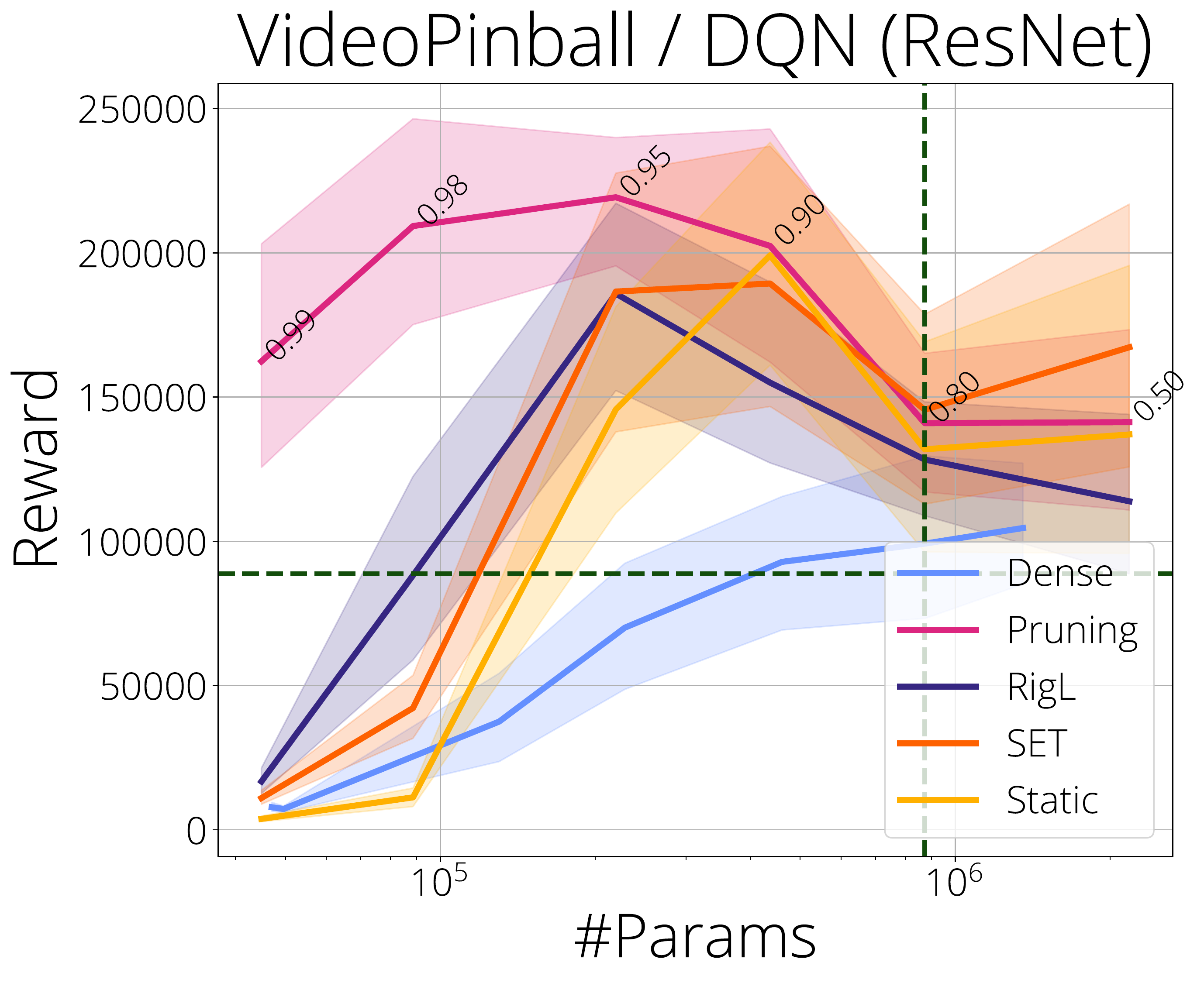}
    \caption{ResNet: Sparsity plots per game.}
    \label{fig:atari-resnet}
\end{figure}

Here we share results on additional environments, Acrobot, CartPole, MountainCar, Hopper, and Ant. \autoref{fig:dqn-classic-app} compares final reward relative to parameter count using DQN. \gls{erk} sparsity distribution was used in the top row whilst uniform was used in the bottom row. \autoref{fig:mujoco-extra-app} presents results on the two remaining MuJoCo environment, Hopper and Ant with SAC (top row) and PPO (bottom row). In Figures \ref{fig:atari-cnn} and \ref{fig:atari-resnet} we show sparsity scaling plots for 15 Atari games using the standard CNN and ResNet respectively.

\newpage

\newpage
\section{Additional Results with Uniform Sparsity Distribution}
\label{app:sec:additional_results_uniform}

\begin{figure*}[!h]
    \centering
    \includegraphics[width=0.3\textwidth]{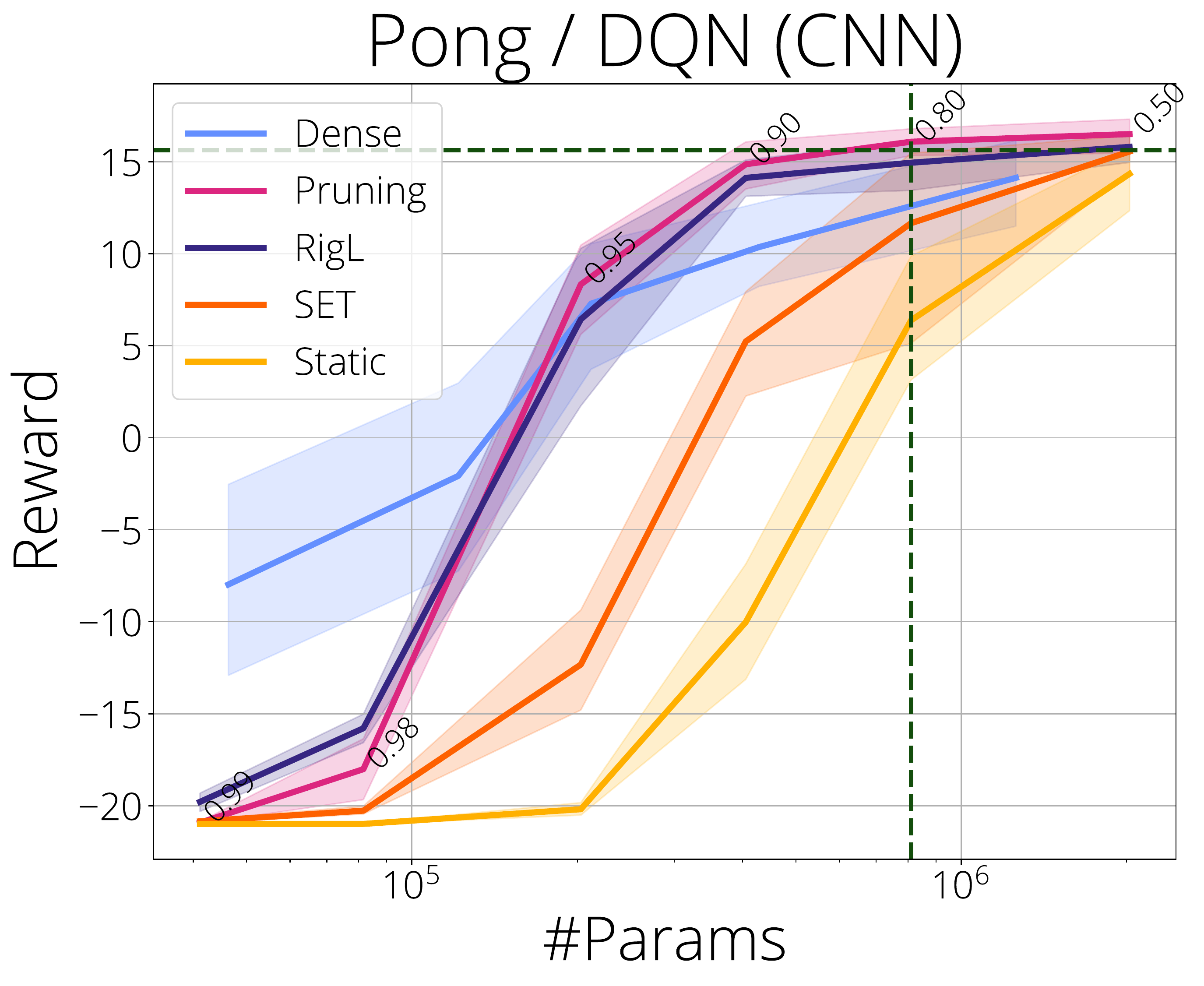}
    \includegraphics[width=0.3\textwidth]{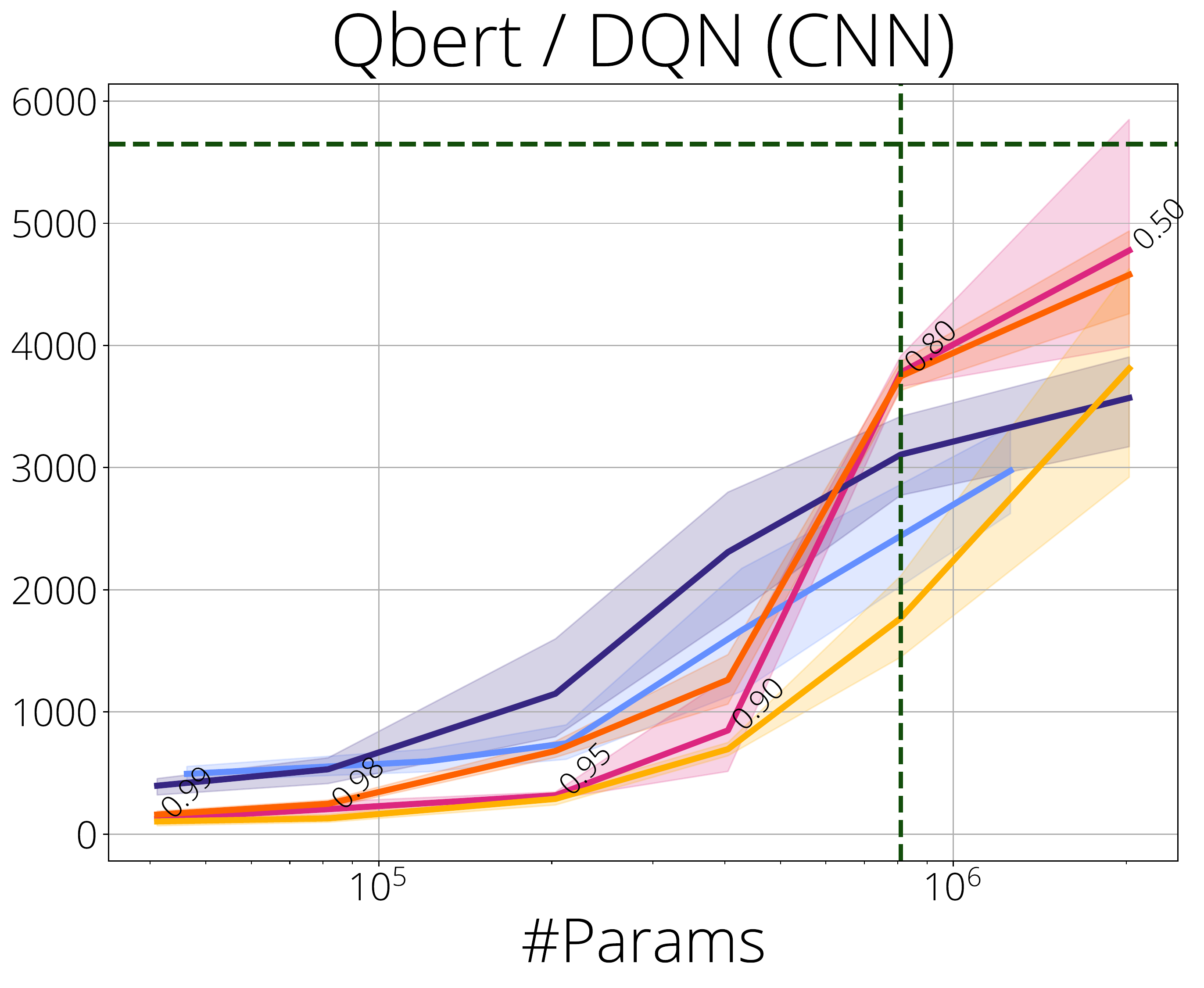}
    \includegraphics[width=0.3\textwidth]{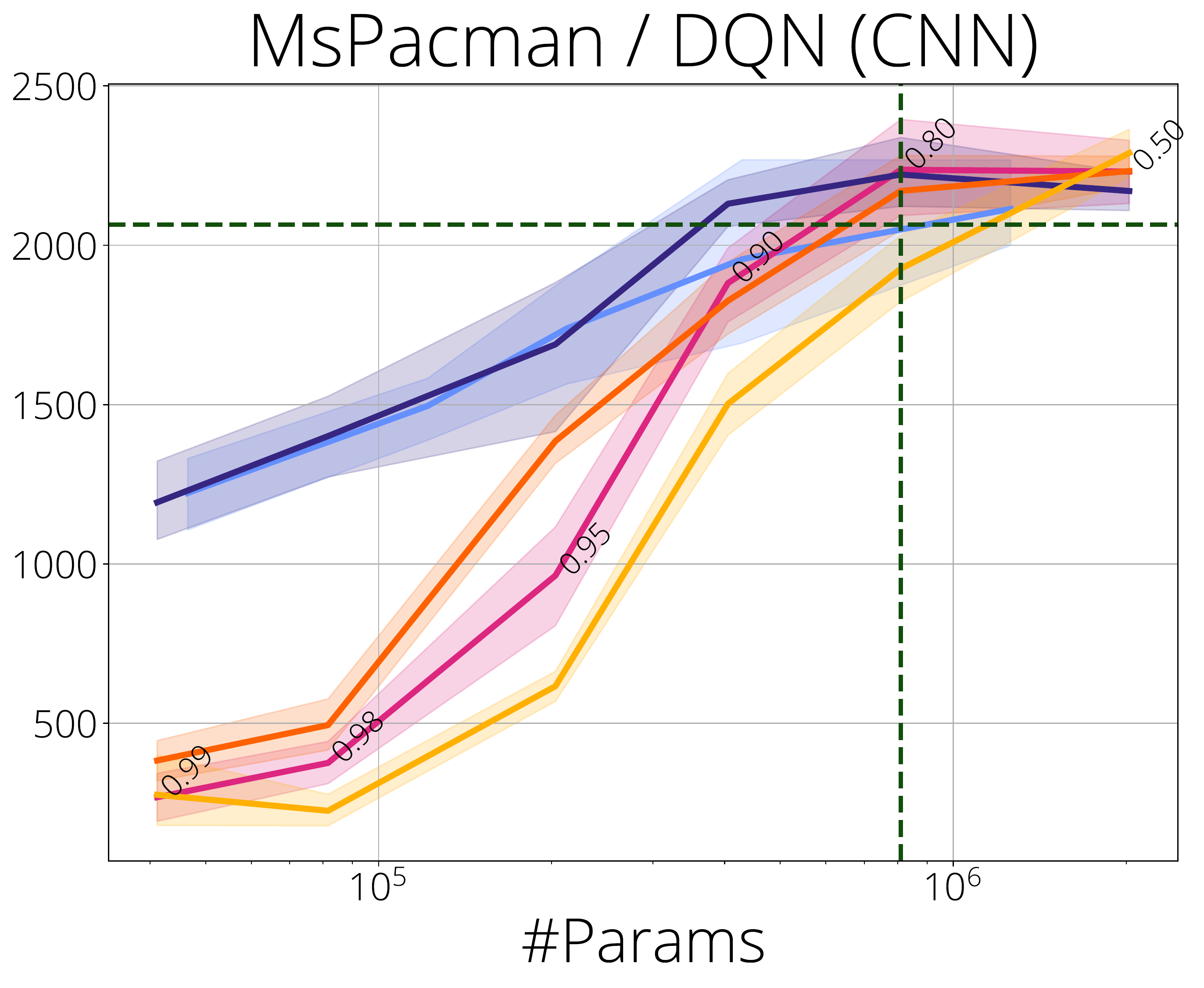}
    \includegraphics[width=0.3\textwidth]{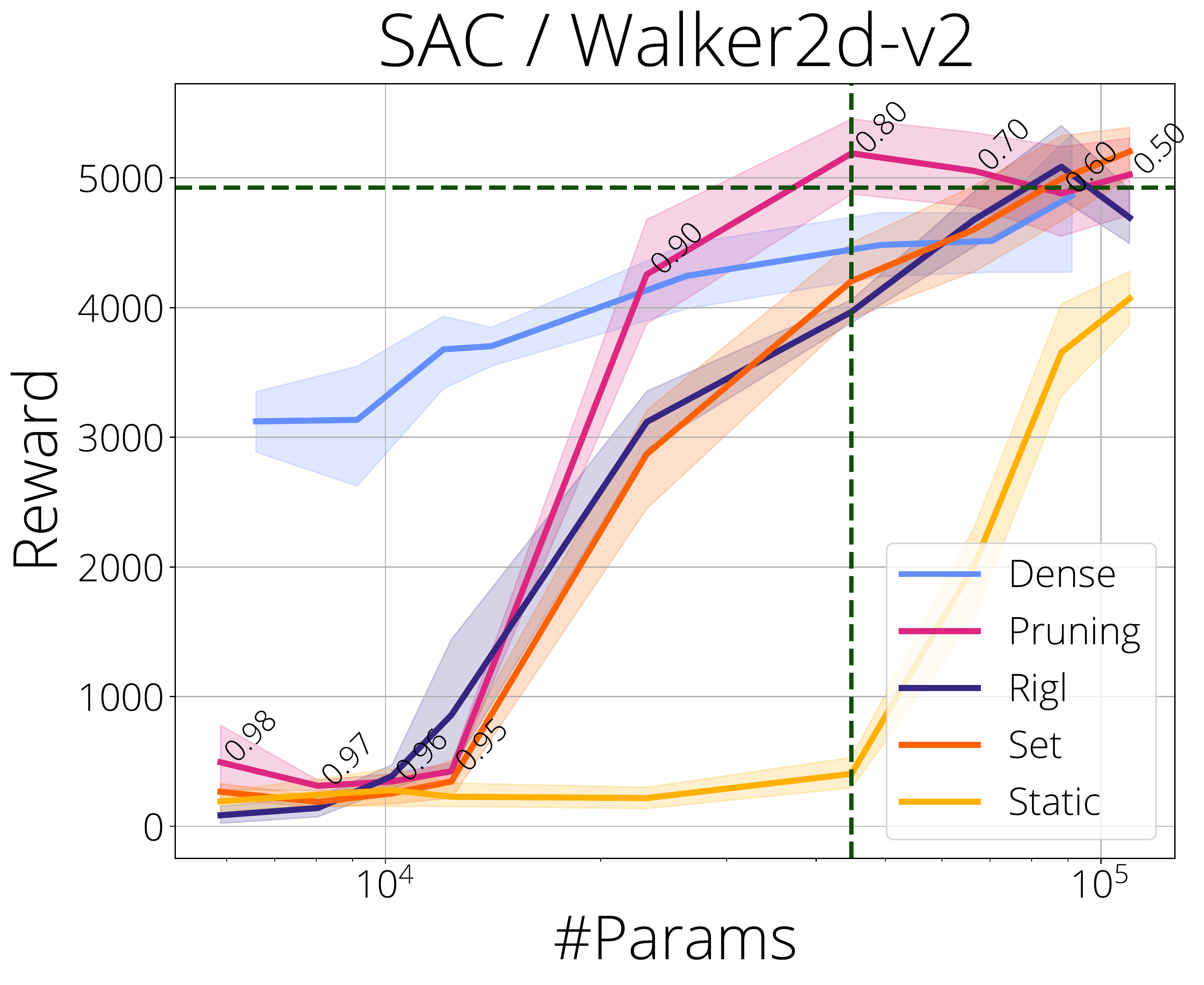}
    \includegraphics[width=0.3\textwidth]{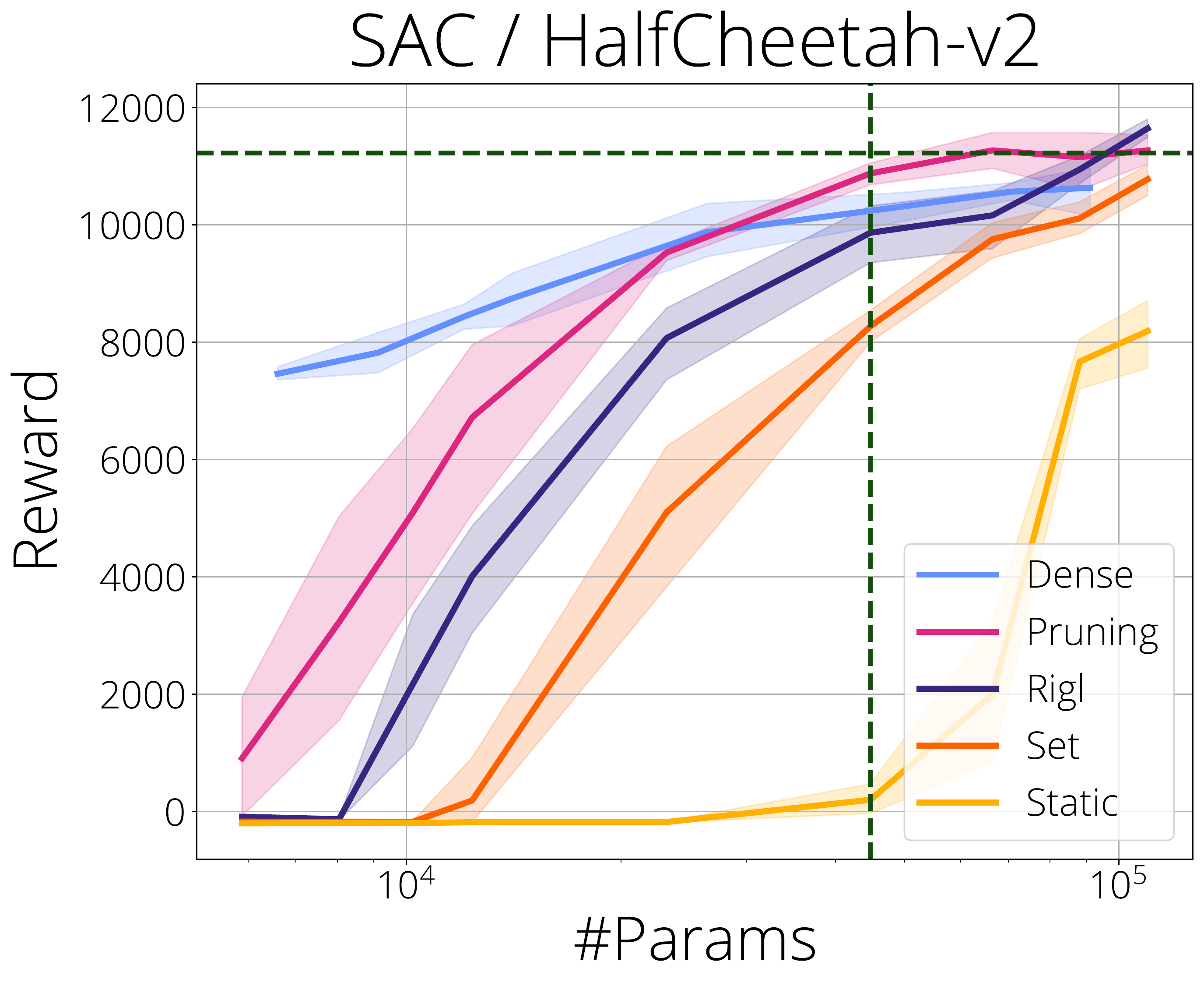}
    \includegraphics[width=0.3\textwidth]{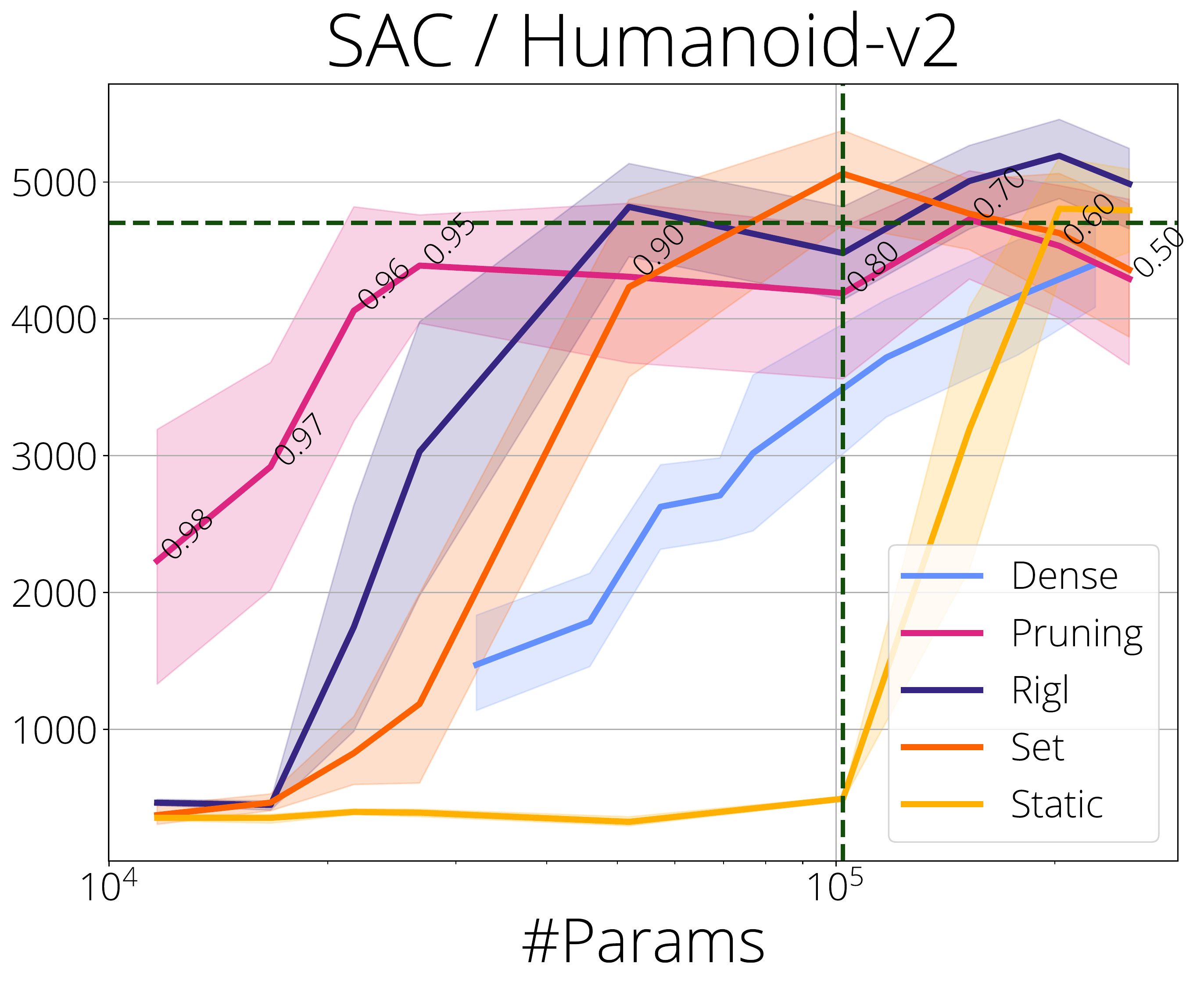}
    \includegraphics[width=0.3\textwidth]{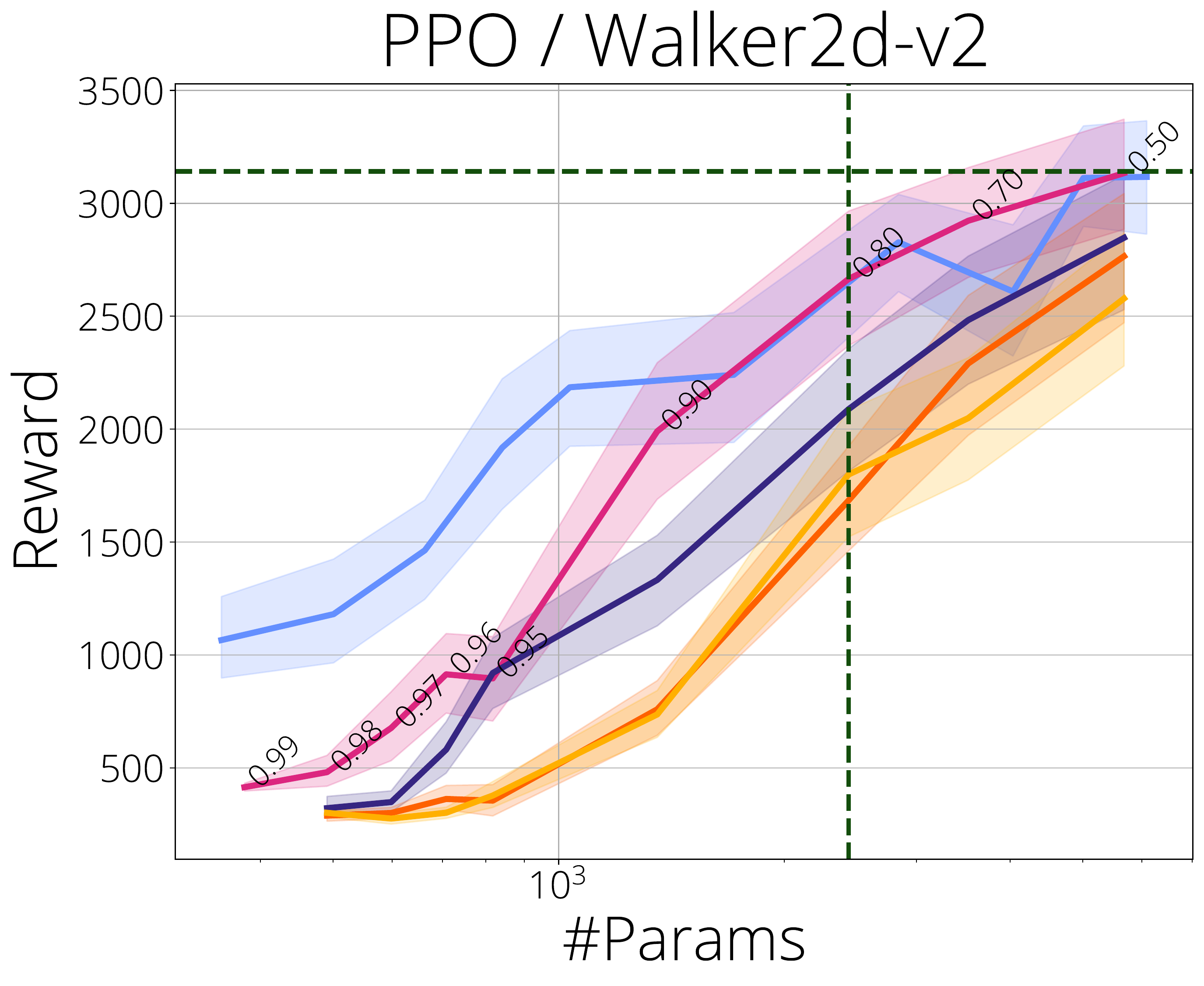}
    \includegraphics[width=0.3\textwidth]{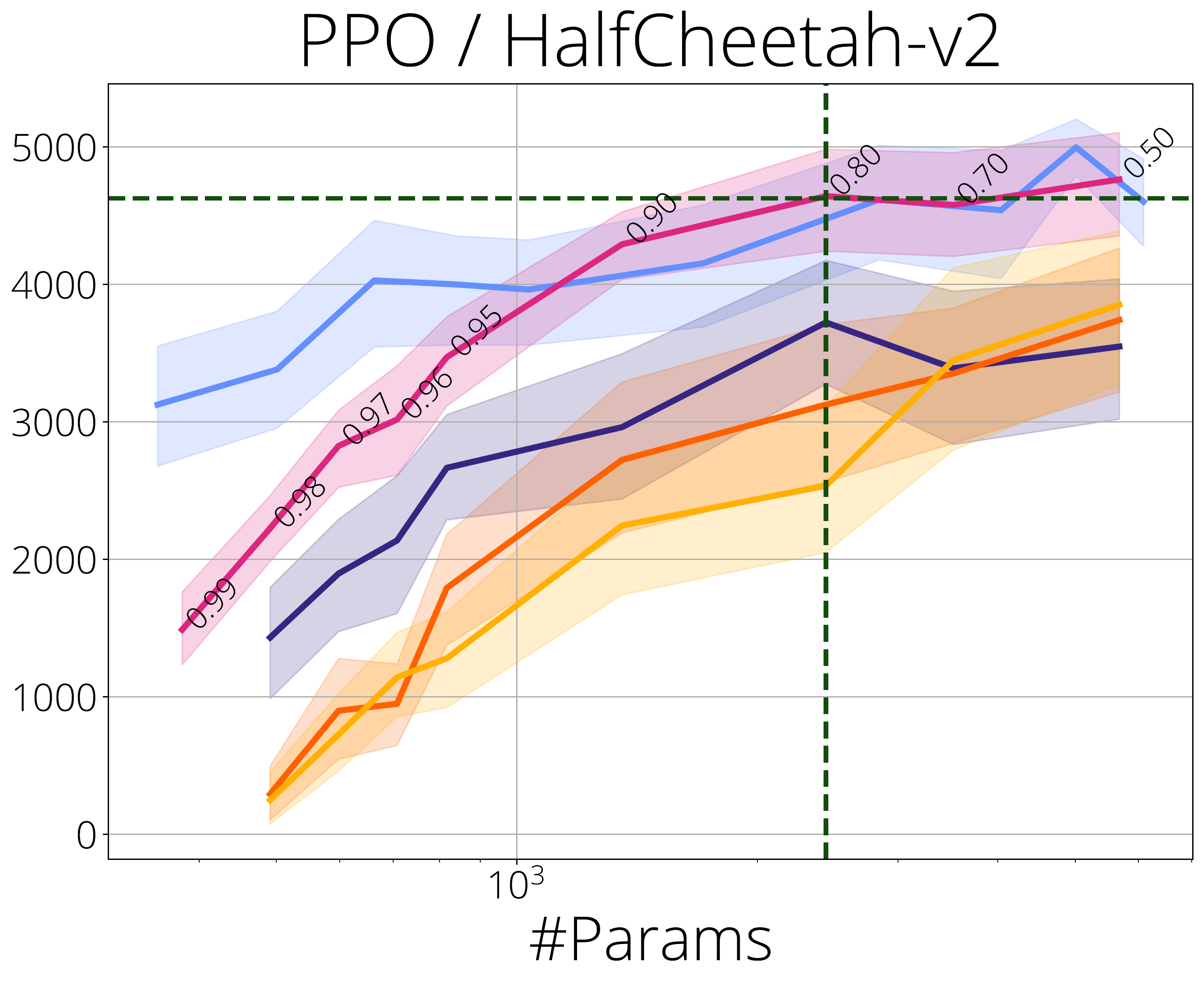}
    \includegraphics[width=0.3\textwidth]{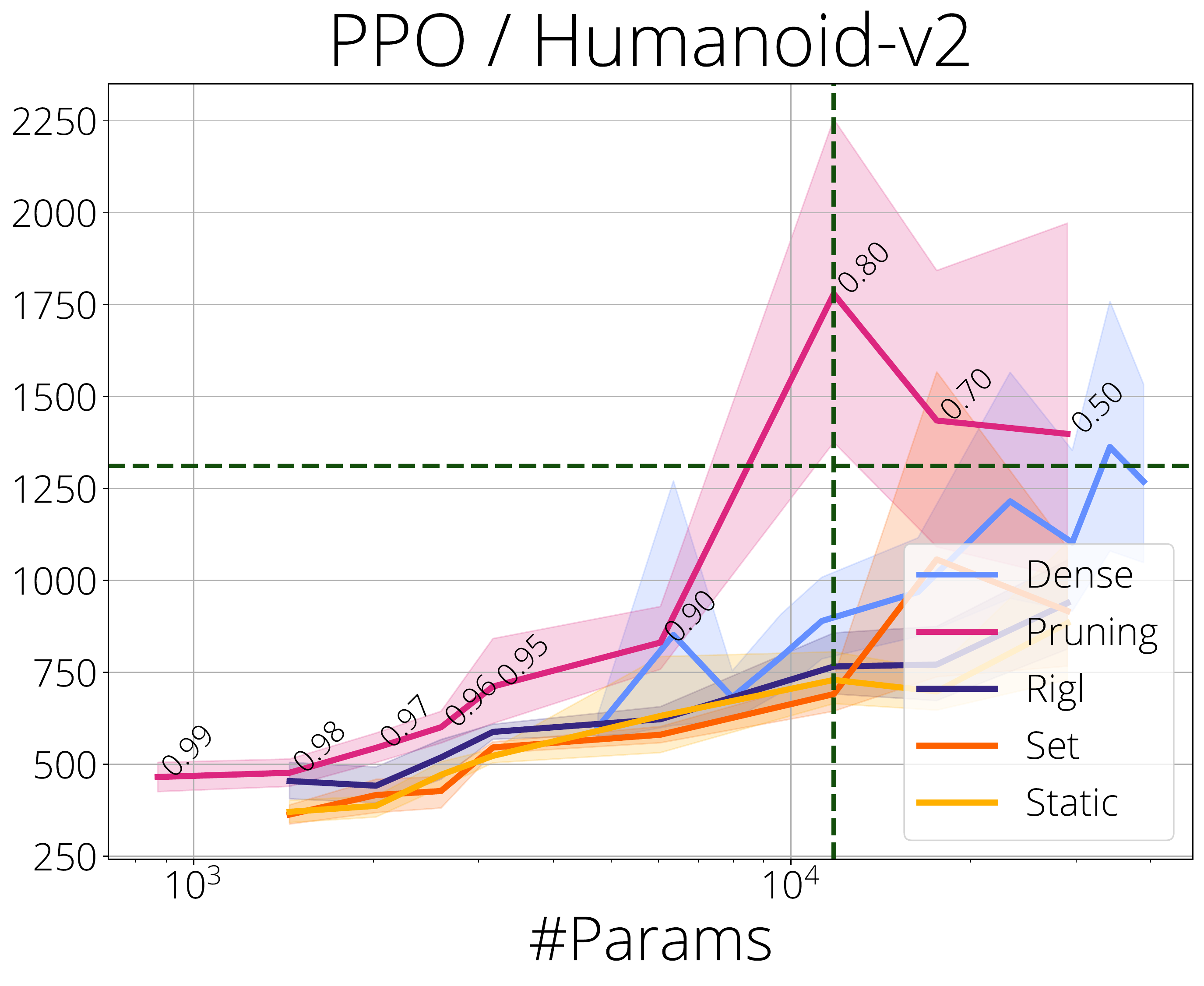}
    \caption{Uniform network sparsity initialization. Comparison of final reward relative to parameter count for the various methods considered: (row-1) DQN on Atari (row-2) SAC on MuJoCo (row-3) PPO on MuJoCo. We consider sparsities from 50\% to 95\% for sparse training methods and pruning. Parameter count for networks with 80\% sparsity and the reward obtained by the dense baseline are highlighted with vertical and horizontal lines. Shaded areas represent 95\% confidence intervals.}
    \label{fig:highLevelComparisonUniform}
\end{figure*}

In \autoref{fig:highLevelComparisonUniform} we repeat the experiments presented in \autoref{fig:highLevelComparisonERK}, however this time using a uniform network sparsity distribution at initialization. These plots provide further evidence as the the benefit of using \gls{erk} over uniform to distribute network sparsity.

\newpage
\section{Additional Hyper-parameter Sensitivity Plots}
\label{app:additional_hypers}
Here we share the remaining plots for our analysis on the sensitivity of sparse training algorithms to various hyper-parameters. Policies area trained using SAC. We note that different architectures, environments and training algorithms might show different curves, which we omit due to high cost of running such analysis.
\begin{figure*}[!h]
    \centering
    \includegraphics[width=0.3\textwidth]{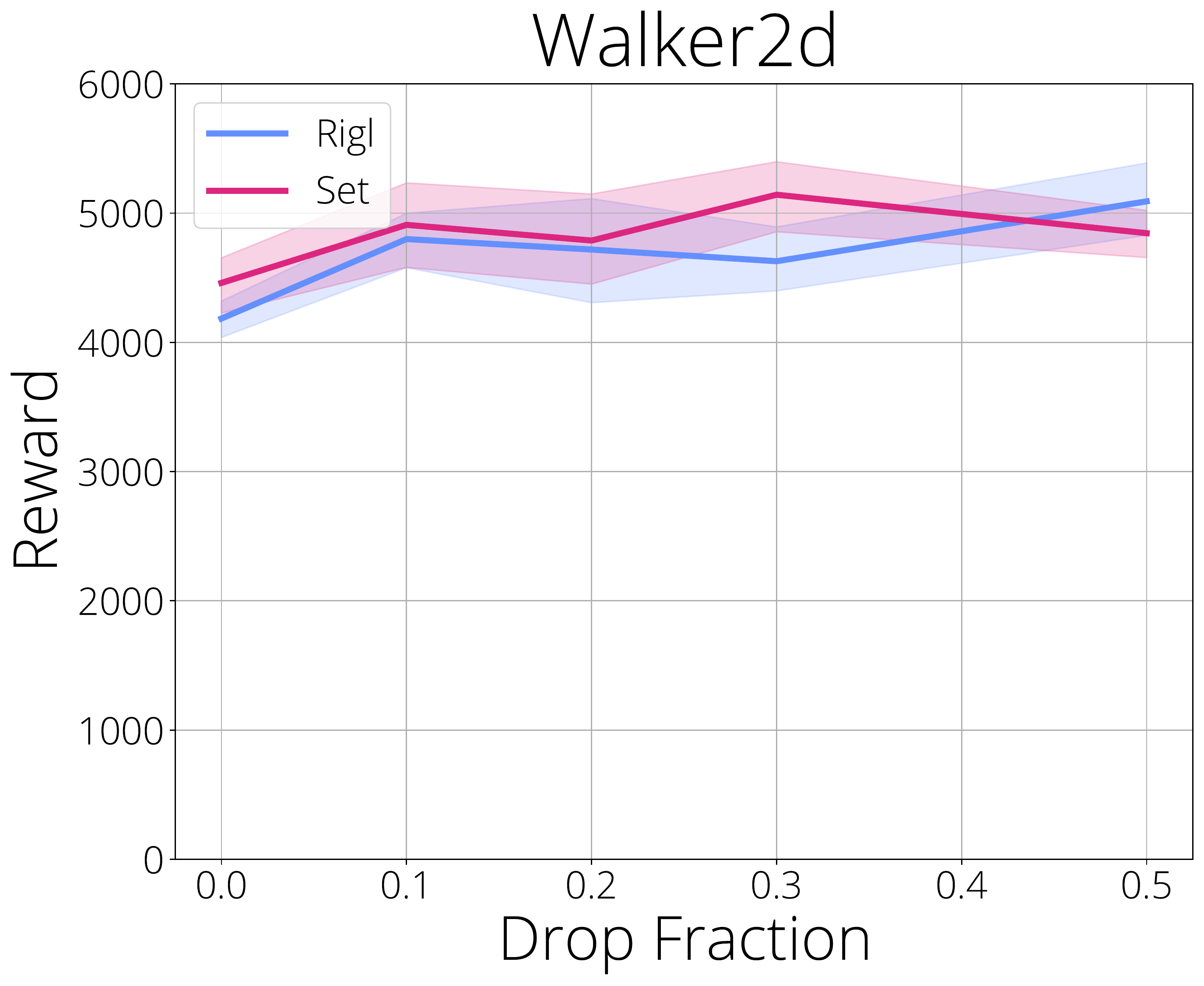}
    \includegraphics[width=0.3\textwidth]{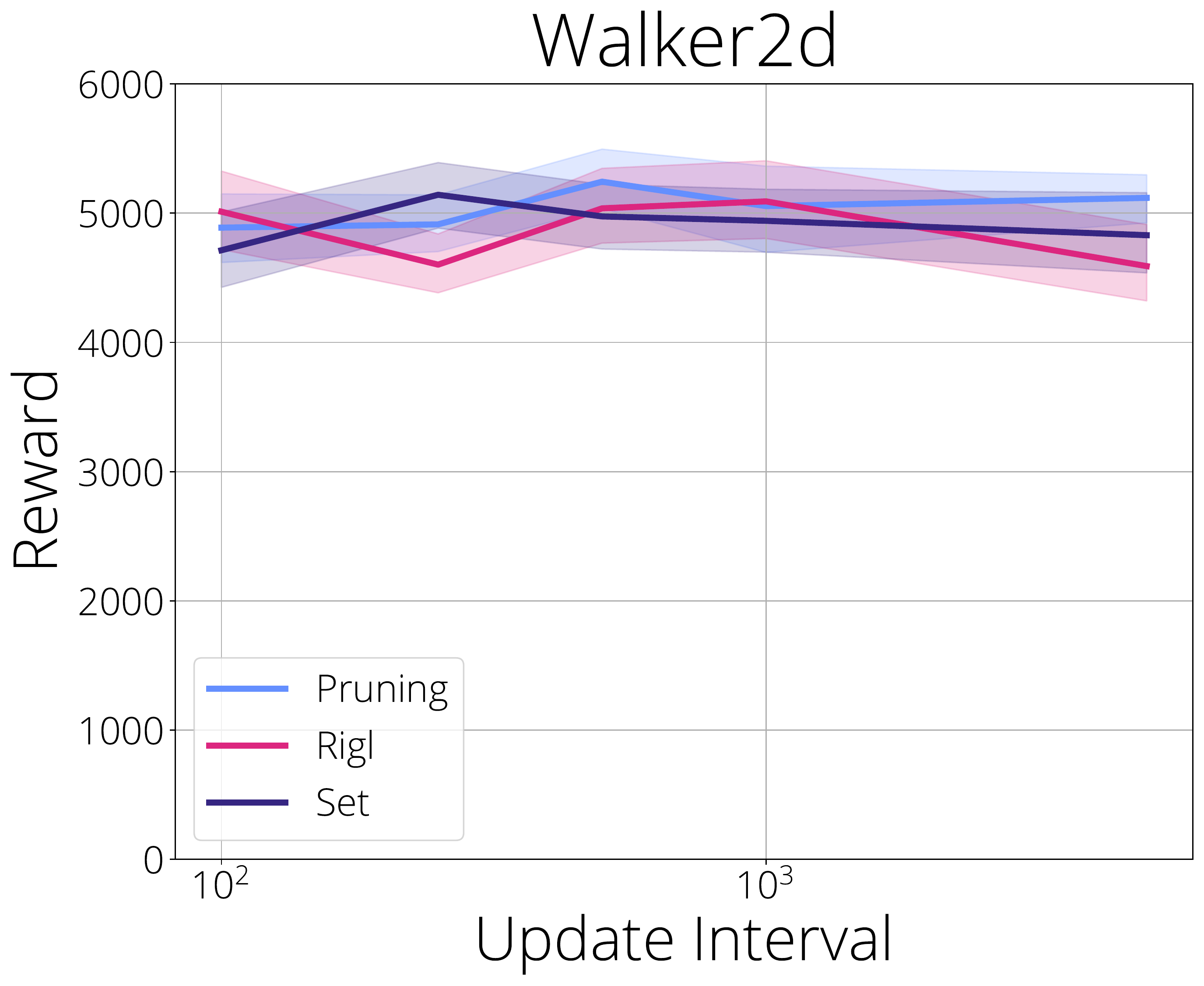}
    \includegraphics[width=0.3\textwidth]{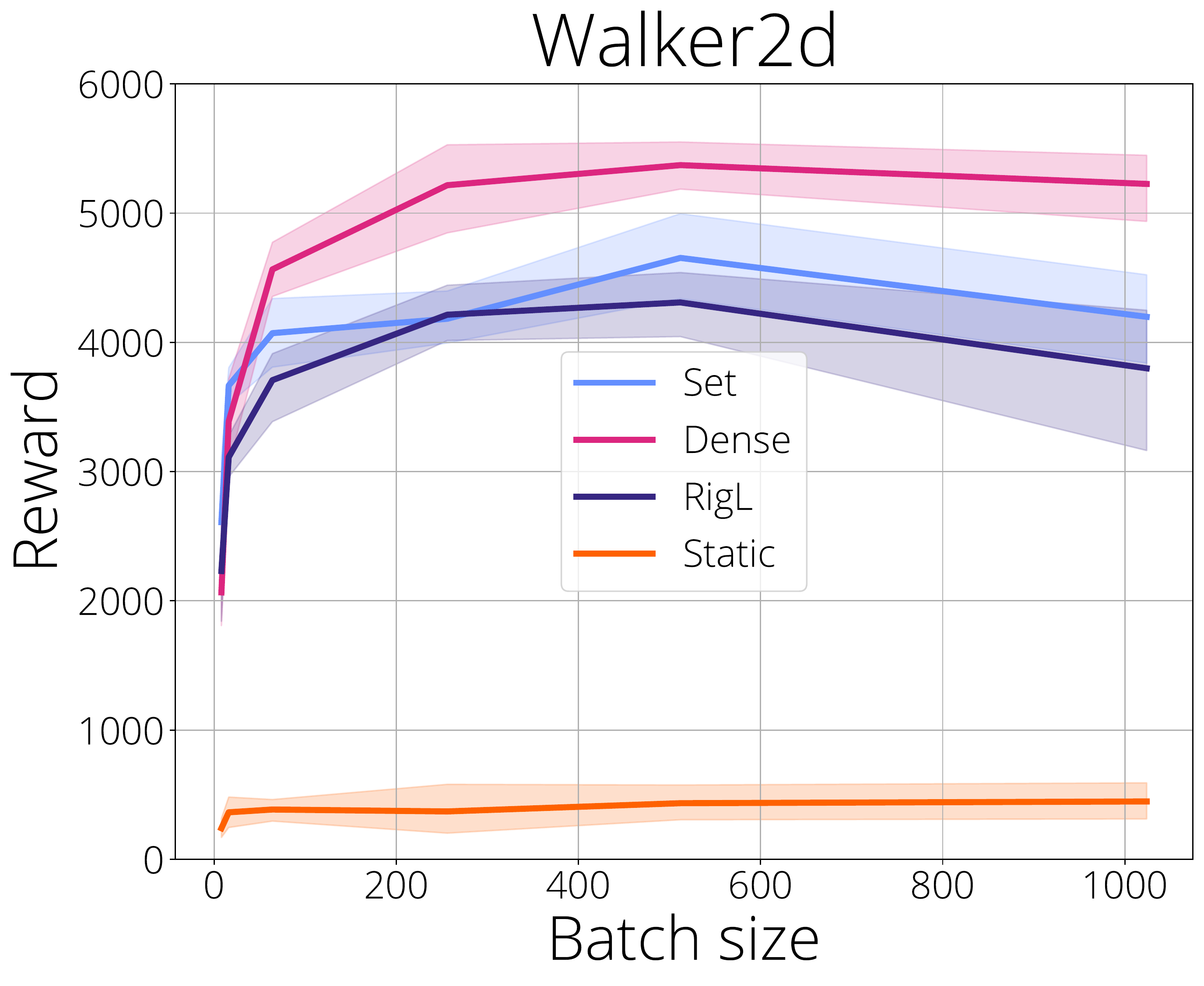}
    \caption{Sensitivity analysis: (left) Drop fraction: Comparing different drop fractions for \gls{set} and \gls{rigl} at 80\% sparsity. (center) Comparing the effect of topology update interval on pruning, static, and \gls{rigl}. (right) The effect of batch size.}
    \label{fig:app:hyper-ablations}
\end{figure*}

\paragraph{Drop fraction}
In \autoref{fig:app:hyper-ablations} (left) we evaluate the effect of drop fraction and observe that drop fractions $>0$ yield a small performance improvement over drop fraction $=0$, which is equivalent to static sparse training. This indicates that changing the network topology during training helps performance. Surprisingly training does not appear particularly sensitive to the drop fraction chosen, with values from 10 - 50\% yielding approximately equivalent performance. It is possible that this is because the environment is relatively easy, thus leading to little separation between different settings. The large improvement that \gls{rigl} and \gls{set} give over static (drop fraction = 0) in the Humanoid and Atari environments is one reason to suspect this.

\textit{Findings:} Drop fractions of 10 - 50\% worked well for \gls{rigl}, whilst a drop fraction of 30\% worked best for \gls{set}. 30\% therefore appears to be a reasonable default for dynamic sparse training. However, the uniformity of performance merits investigation on a harder environment in which more separation between drop fractions may be observed.

\paragraph{Topology update interval}
In \autoref{fig:app:hyper-ablations} (center) we evaluate the effect of topology update interval on pruning, \gls{rigl}  and \gls{set} and find that training is not particularly sensitive to it.

\textit{Findings:} Updating the network every 1000 environment steps appears to be a reasonable default.

\paragraph{Batch size}
In \autoref{fig:app:hyper-ablations} (right) we evaluate the effect of batch size. Batch size is critical for obtaining a good estimate for the gradients during training for all methods, however for \gls{rigl} it is also used when selecting new connections. We observe performance degradation for all methods when smaller batch sizes are used, with no particular additional effect on \gls{rigl}. 

\textit{Findings:} Sparse networks seems to have similar sensitivity to the batch size as the dense networks.

\begin{figure*}[!h]
    \centering
    \includegraphics[width=0.3\textwidth]{figs/sdist_sac.pdf}
    \includegraphics[width=0.3\textwidth]{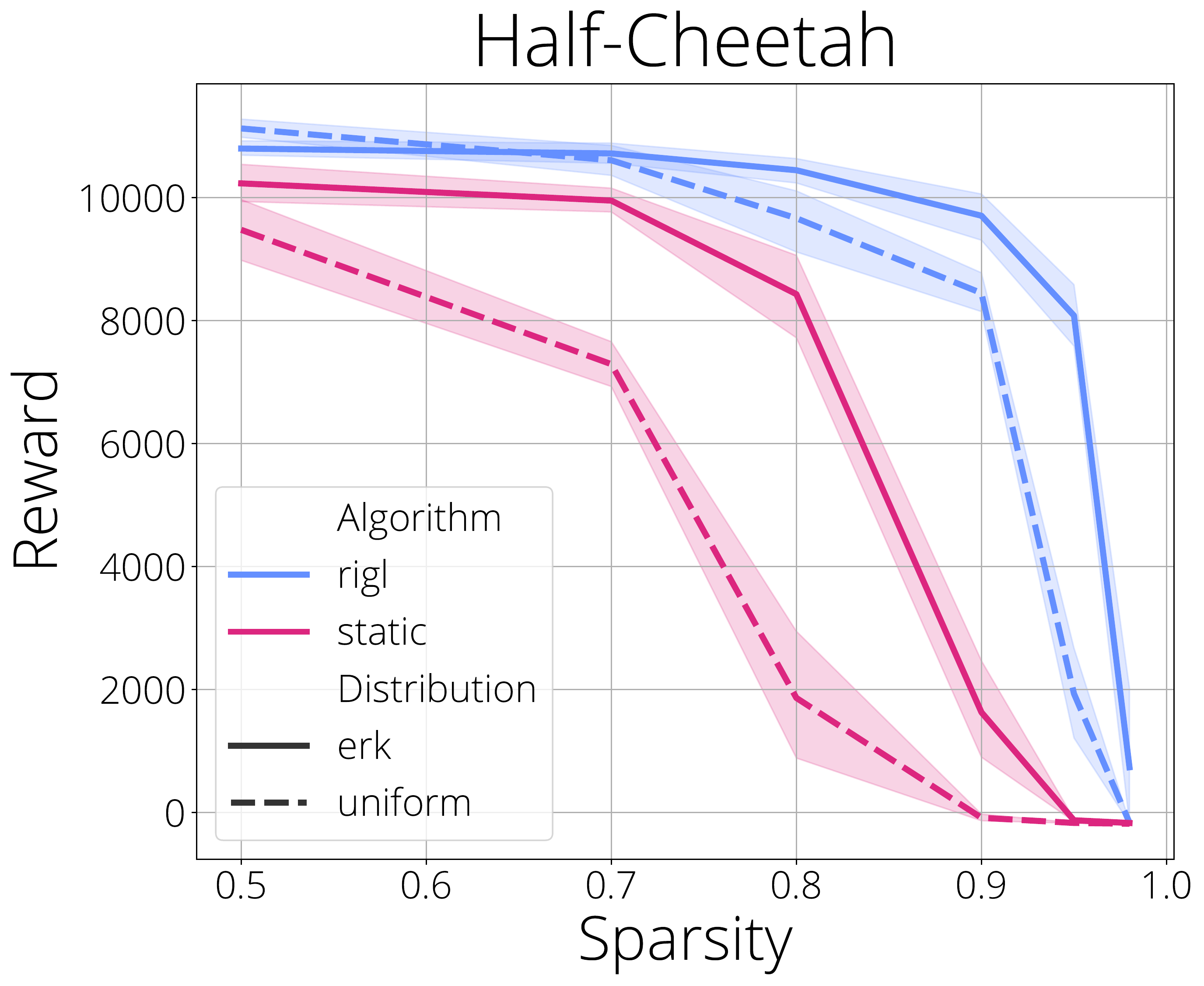}
    \includegraphics[width=0.3\textwidth]{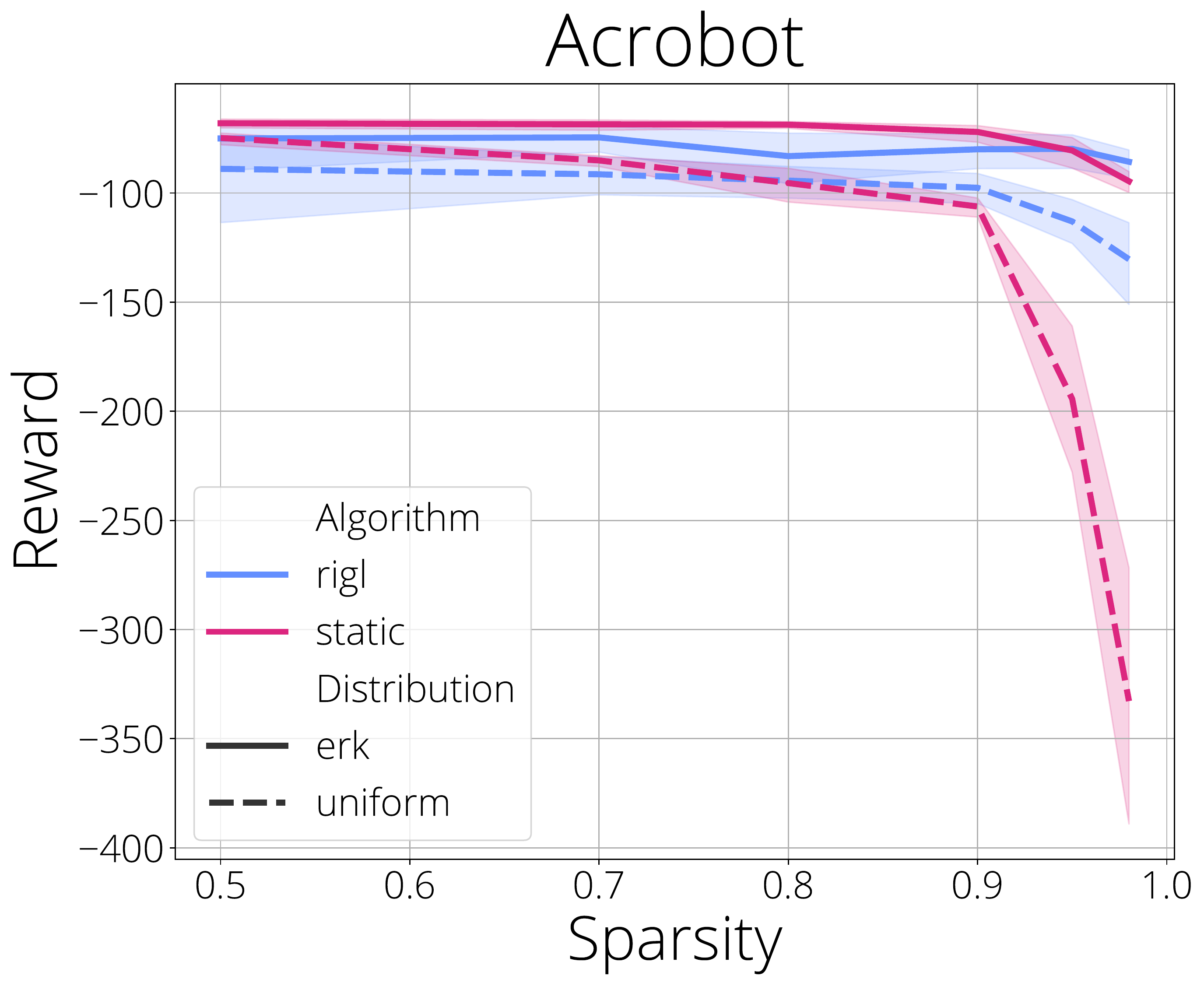}
    \caption{Evaluating non uniform sparsities on HalfCheetah using SAC when all layers pruned (left) and last layer is kept dense (center). On the right we prune all layers of the DQN MLP network on the Acrobot task.}
    \label{fig:app:withinNetDist}
\end{figure*}

\paragraph{Within network sparsity} In \autoref{fig:app:withinNetDist} we share additional analysis on within network sparsity for SAC and also on DQN in the Acrobot environment.

\section{Additional Interquartile Mean (IQM) Plots}
\label{sec:iqm_extended}
\begin{figure*}[!h]
    \centering
    \includegraphics[width=0.48\textwidth]{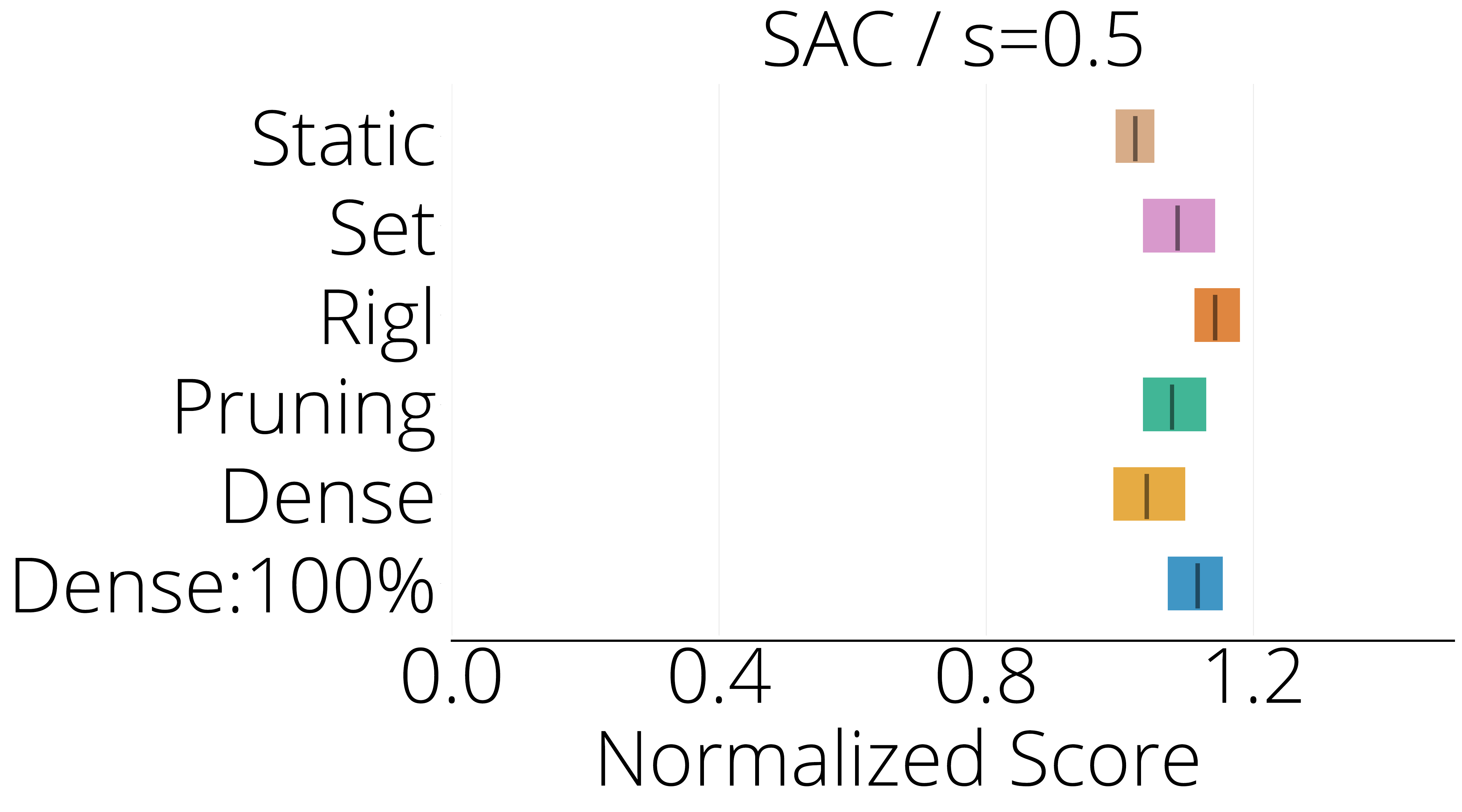}
    \includegraphics[width=0.48\textwidth]{figs/rliable_IQM_mujoco_sac_erk_sparsity_0.9.pdf}
    \includegraphics[width=0.48\textwidth]{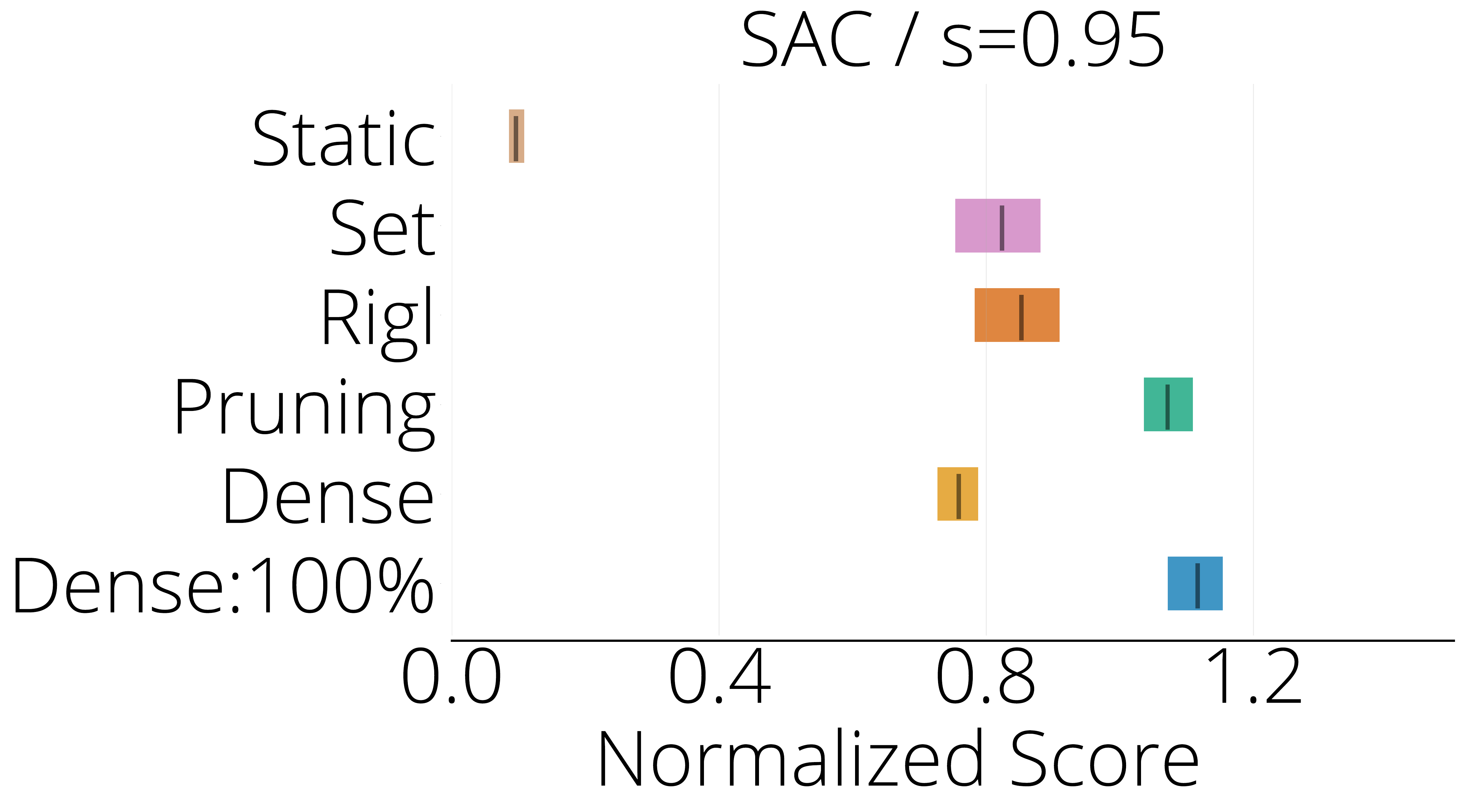}
    \includegraphics[width=0.48\textwidth]{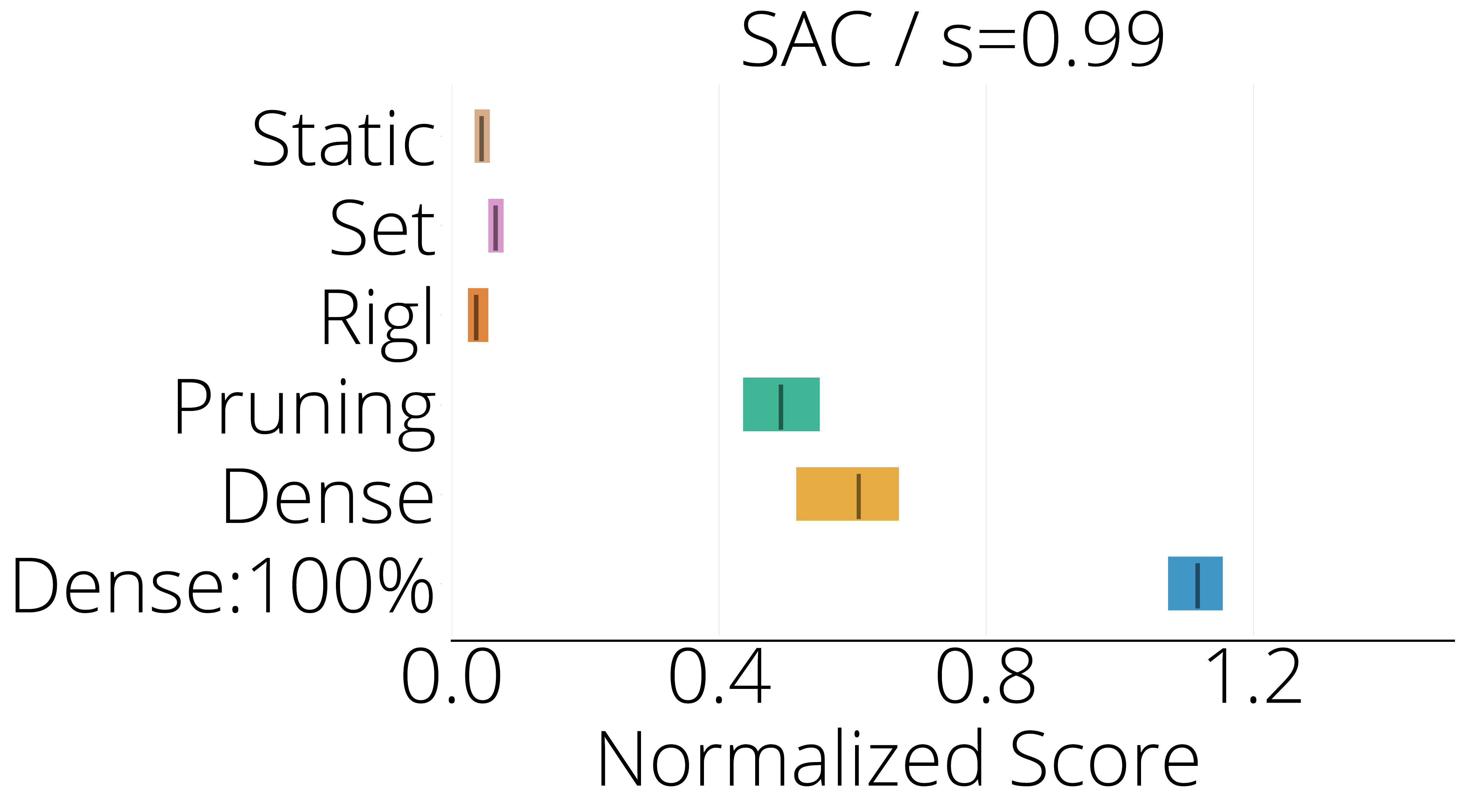}
    \caption{SAC: IQM plots calculated over five Mujoco games (Ant, HalfCheetah, Hopper, Humanoid, Walker2d), with 10 seeds per game. Except for sparsity = 0.99 which only includes results from HalfCheetah, Hopper, and Walker2d, each with 10 seeds.}
    \label{fig:mujoco-sac}
\end{figure*}

\paragraph{Mujoco SAC} In Figure \ref{fig:mujoco-sac} we present the interquartile mean (IQM) calculated over five Mujoco environments for SAC at four different sparsities, 50\%, 90\%, 95\% and 99\%, or the networks with the equivalent number of parameters in the case of Dense training. Note that for 99\% sparsity the IQM is only calculated over three Mujoco environments.

\newpage

\begin{figure*}[!h]
    \centering
    \includegraphics[width=0.48\textwidth]{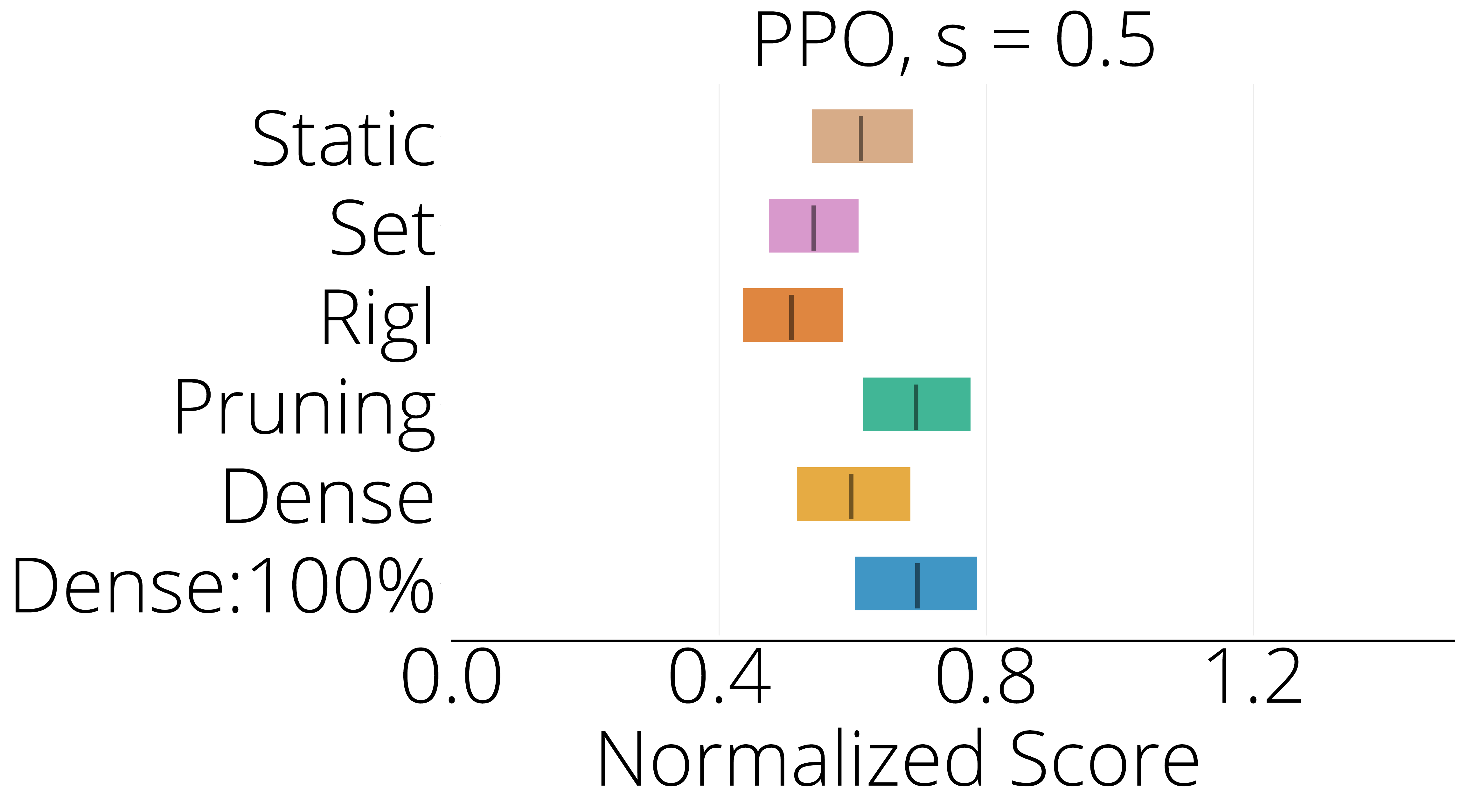}
    \includegraphics[width=0.48\textwidth]{figs/rliable_IQM_mujoco_ppo_erk_sparsity_0.9.pdf}
    \includegraphics[width=0.48\textwidth]{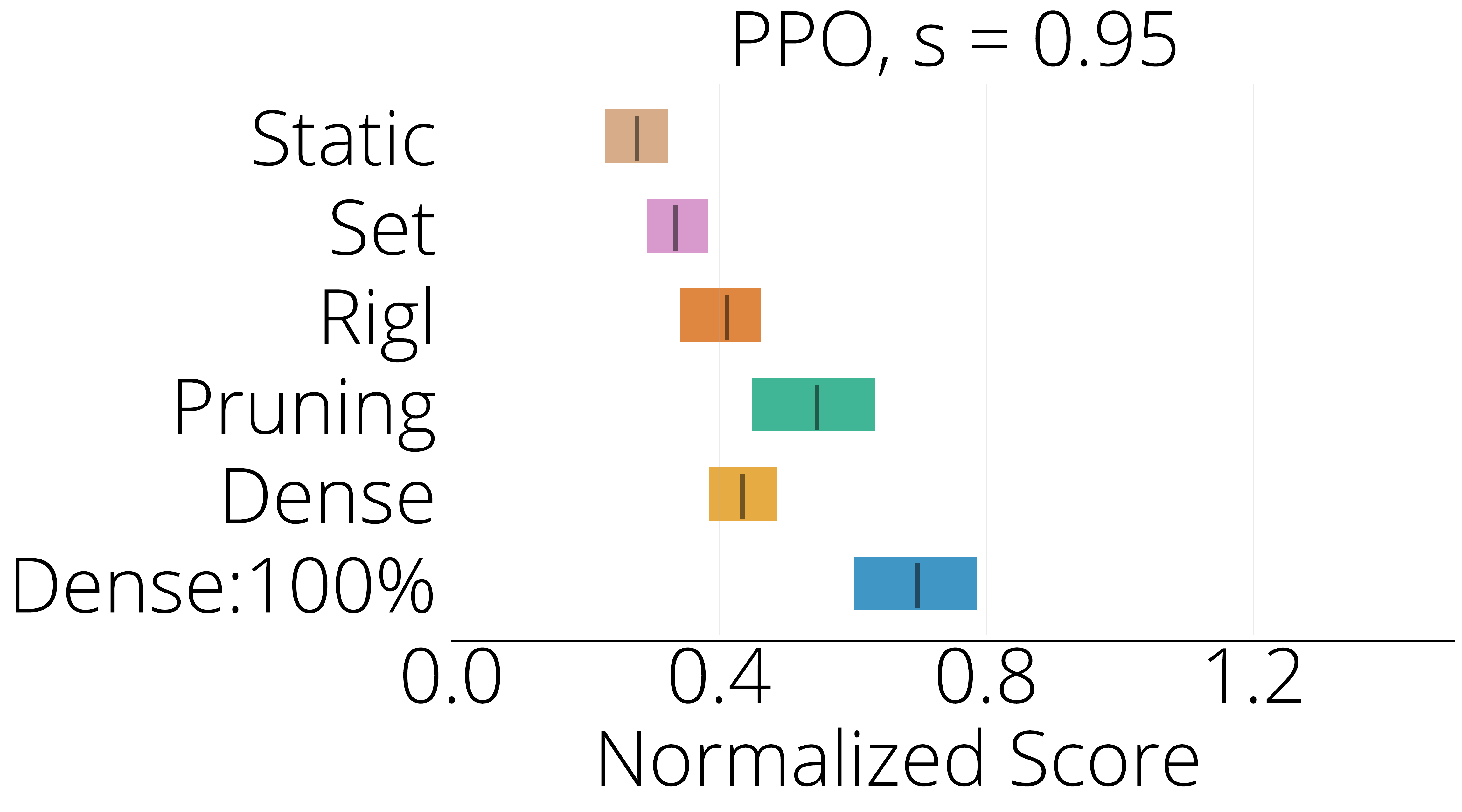}
    \includegraphics[width=0.48\textwidth]{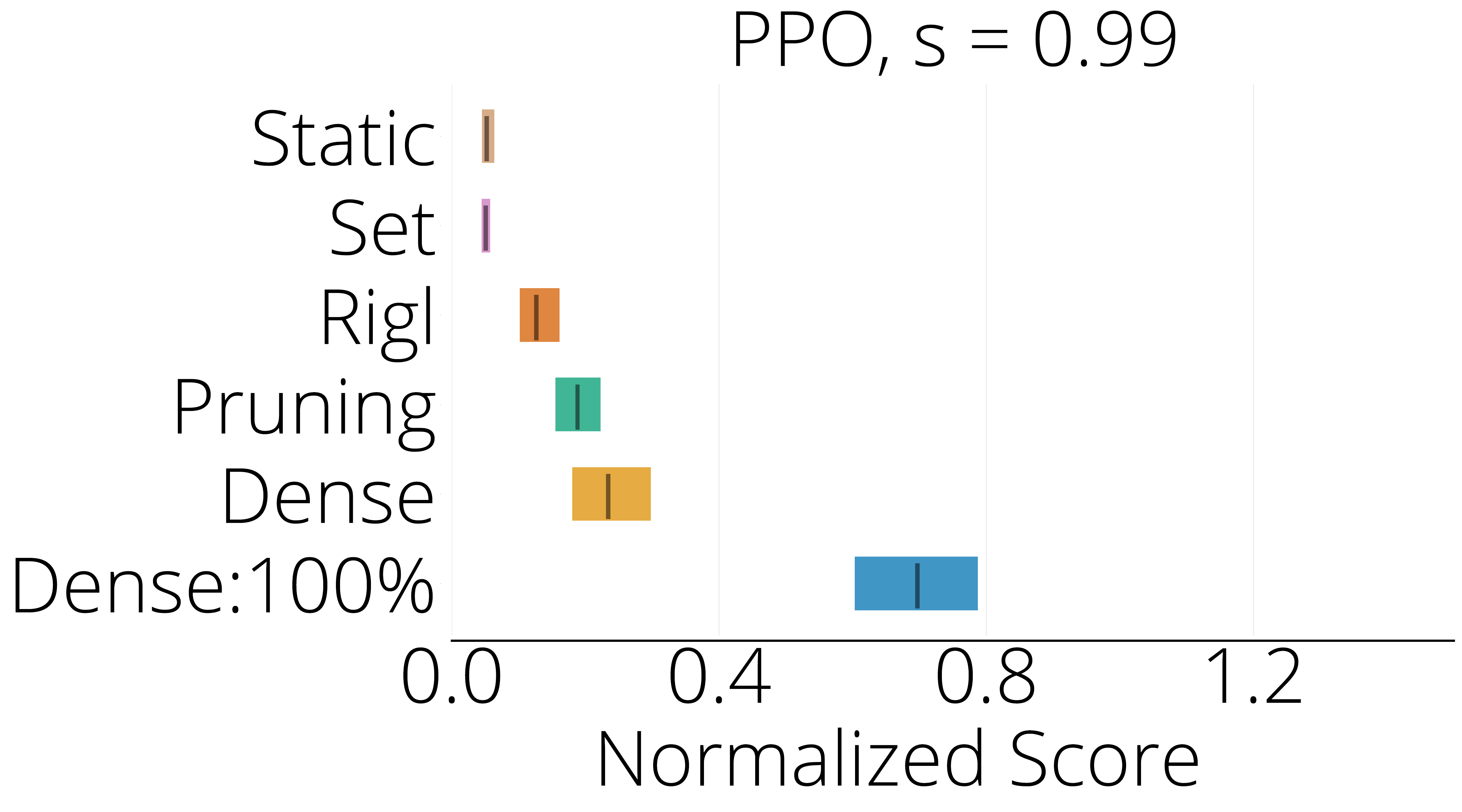}
    \caption{SAC: IQM plots calculated over multiple Mujoco games with 9 seeds per game. 50 - 90\% sparsity include five games (Ant, HalfCheetah, Hopper, Humanoid, Walker2d), 95\% sparsity includes Ant, HalfCheetah, Hopper, Walker2d, and 99\% sparsity HalfCheetah, Hopper, Walker2d.}
    \label{fig:mujoco-ppo}
\end{figure*}

\paragraph{Mujoco PPO} In Figure \ref{fig:mujoco-ppo} we present the interquartile mean (IQM) calculated over five Mujoco environments for PPO at four different sparsities, 50\%, 90\%, 95\% and 99\%, or the networks with the equivalent number of parameters in the case of Dense training. Note that for 95\% sparsity the IQM is calculated over four environments and for 99\% sparsity the IQM is only calculated over three Mujoco environments.

\begin{figure*}[!h]
    \centering
    \includegraphics[width=0.48\textwidth]{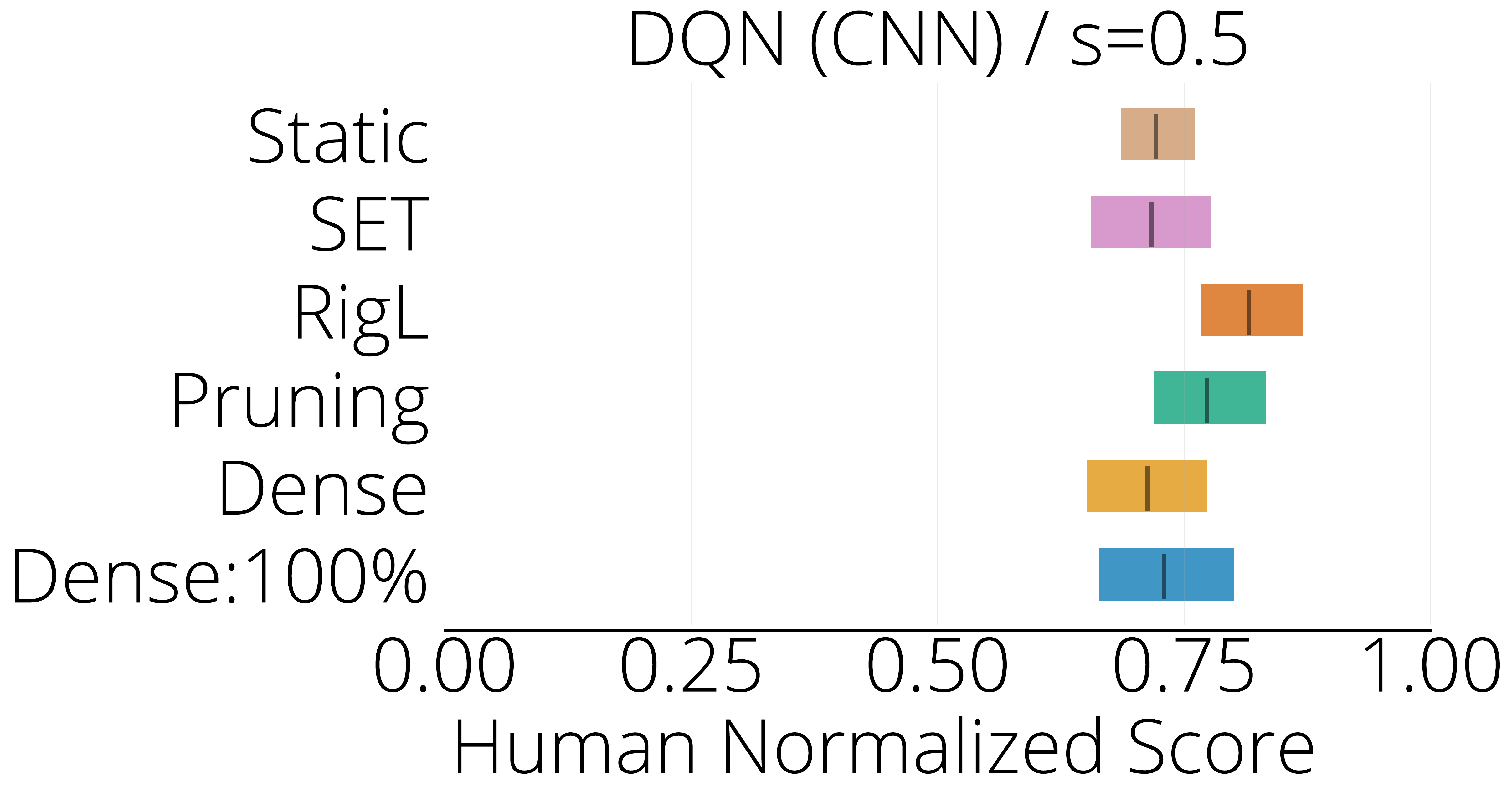}
    \includegraphics[width=0.48\textwidth]{figs/rliable_IQM_atari_dqn_erk_sparsity_0.9.pdf}
    \includegraphics[width=0.48\textwidth]{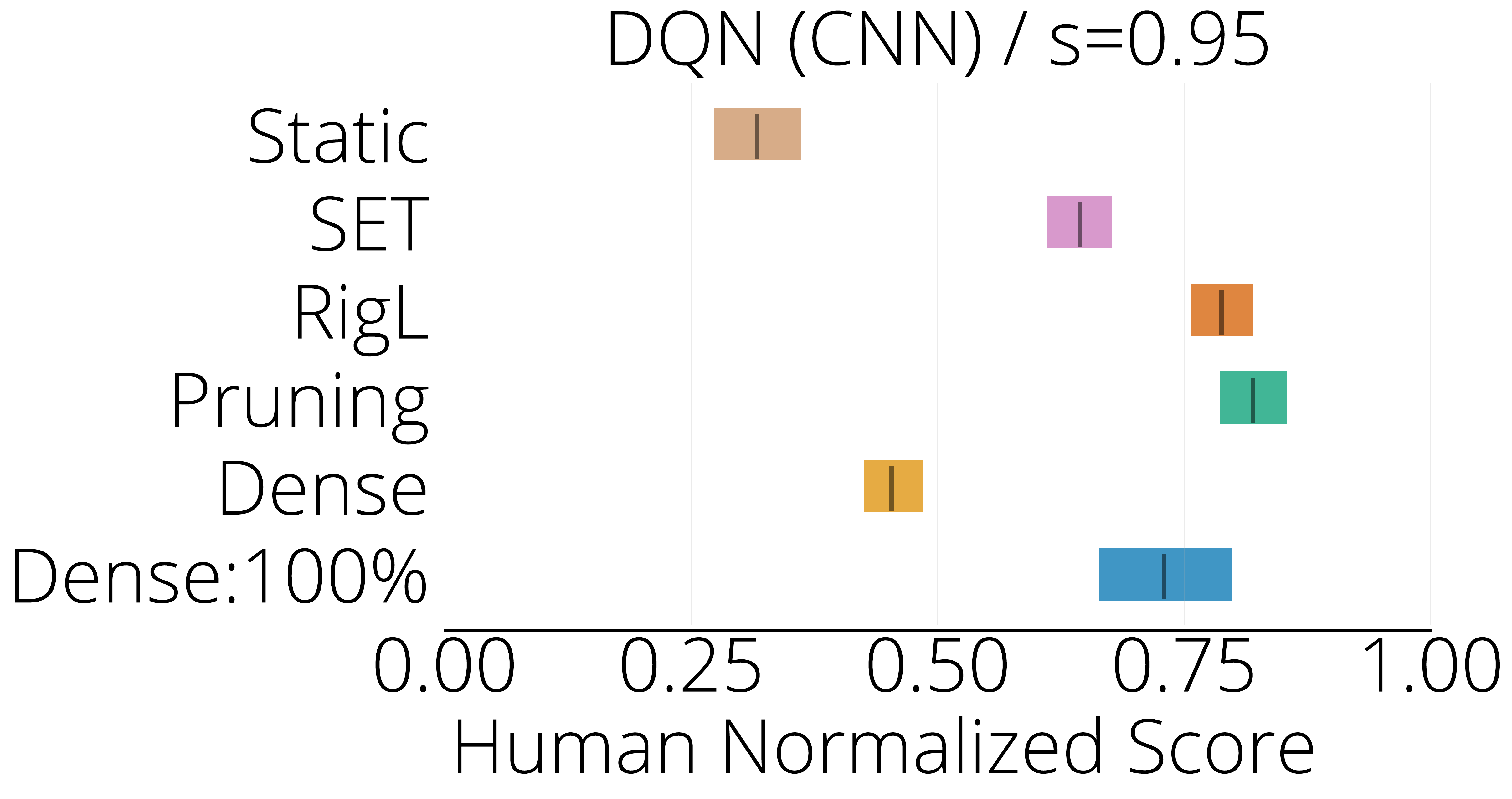}
    \includegraphics[width=0.48\textwidth]{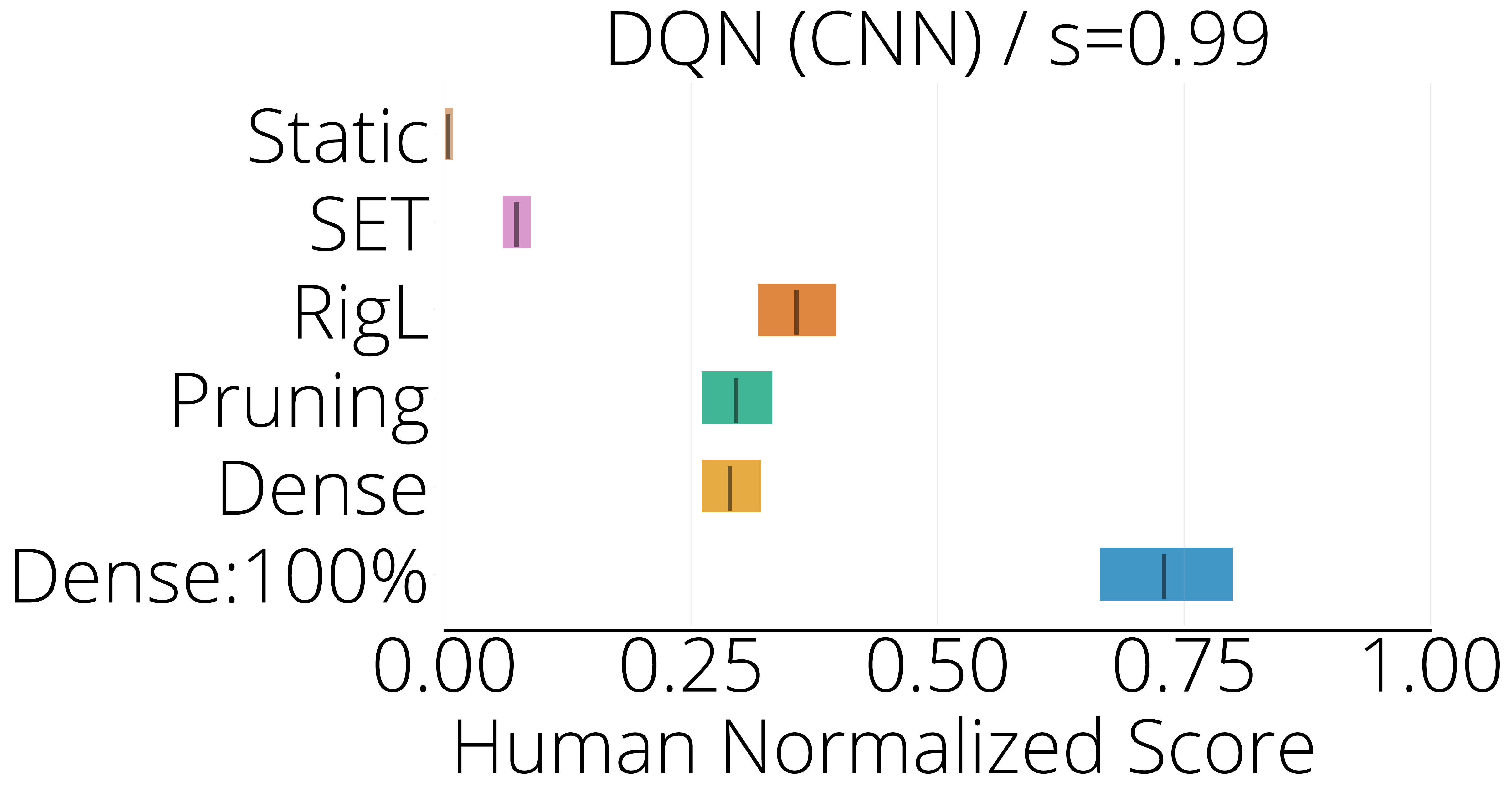}
    \caption{CNN: IQM plots calculated over 15 Atari Games, with 9 seeds per game.}
    \label{fig:atari-naturedqn-iqm}
\end{figure*}

\begin{figure*}[!h]
    \centering
    \includegraphics[width=0.48\textwidth]{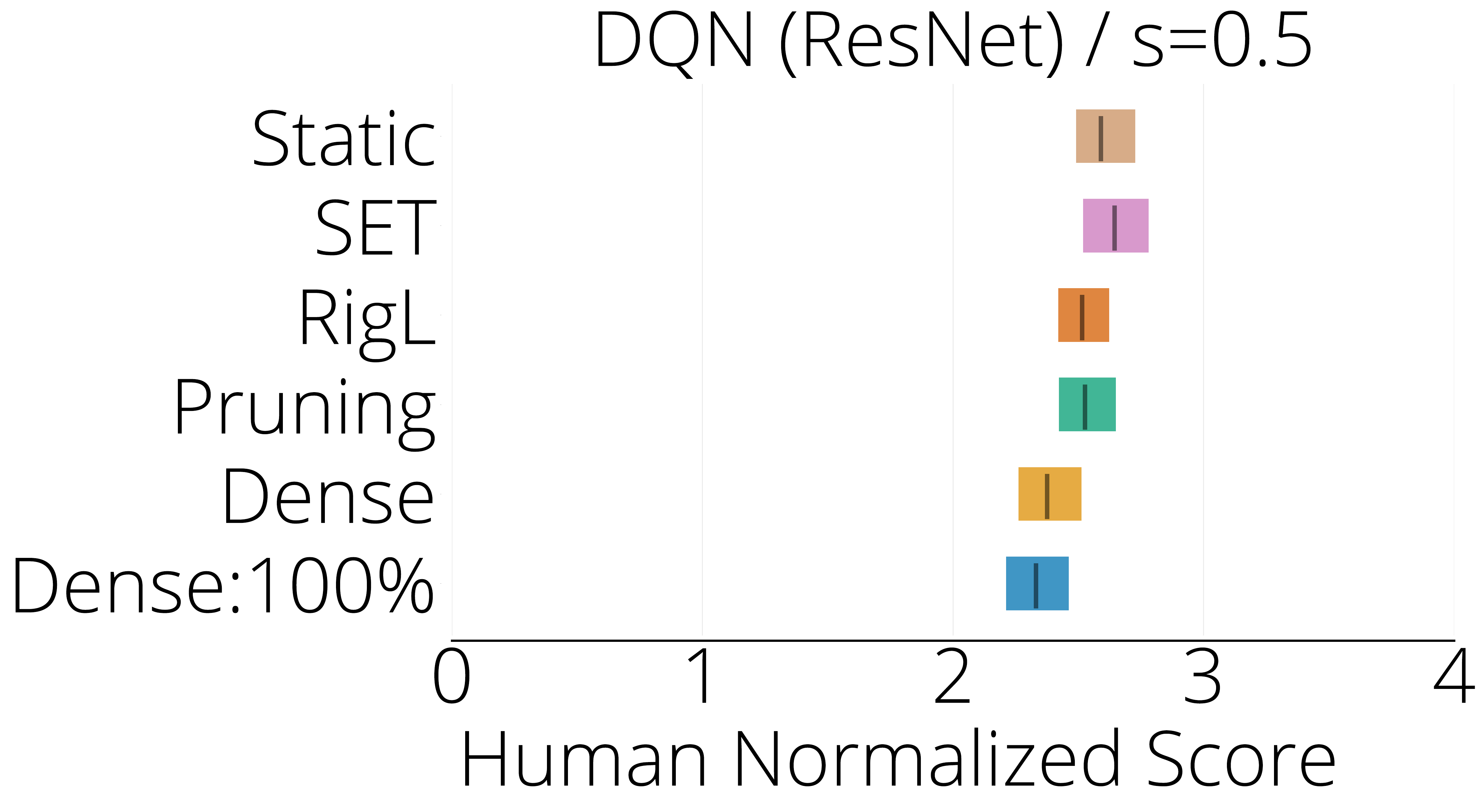}
    \includegraphics[width=0.48\textwidth]{figs/rliable_IQM_atari_dqn_resnet_erk_sparsity_0.9.pdf}
    \includegraphics[width=0.48\textwidth]{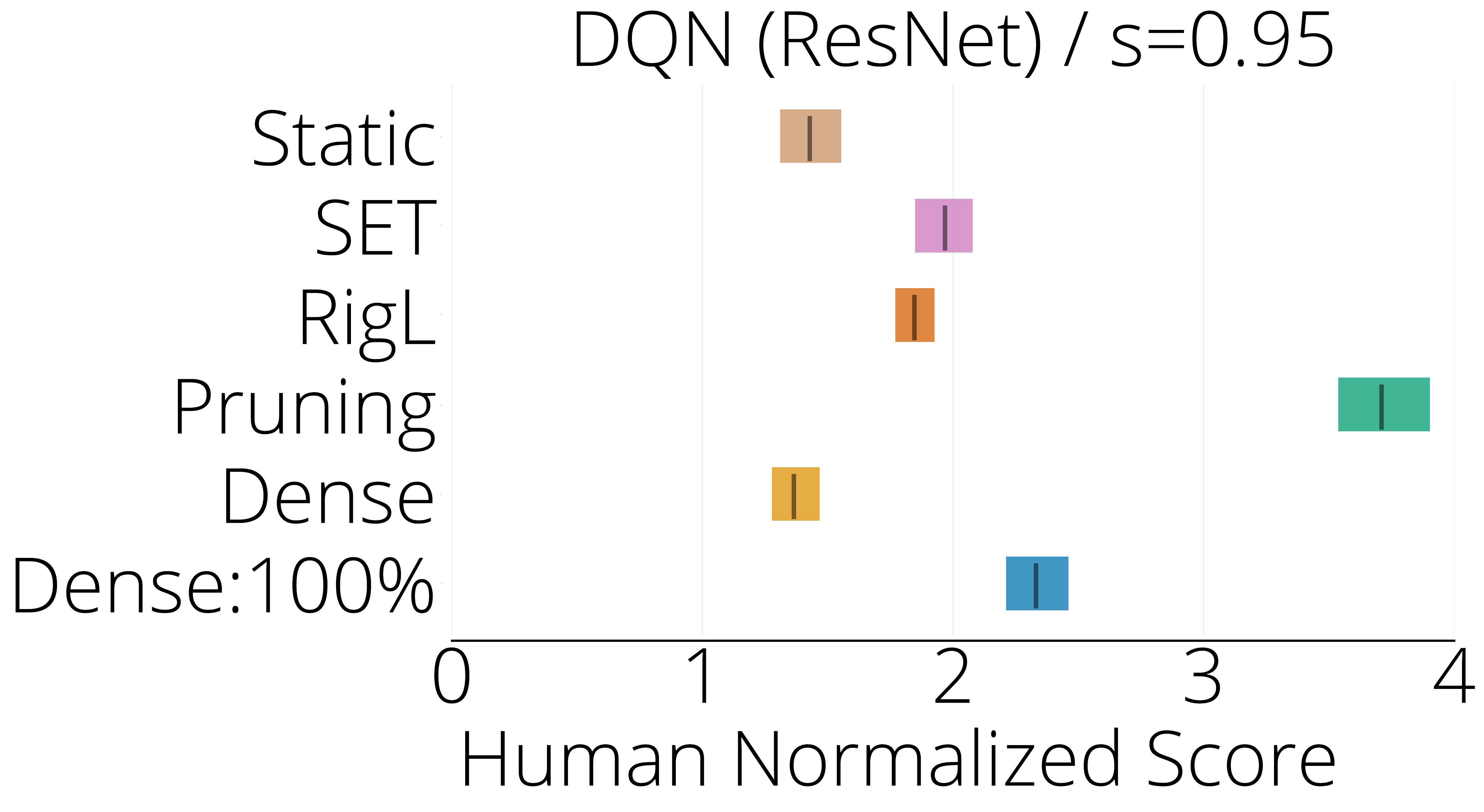}
    \includegraphics[width=0.48\textwidth]{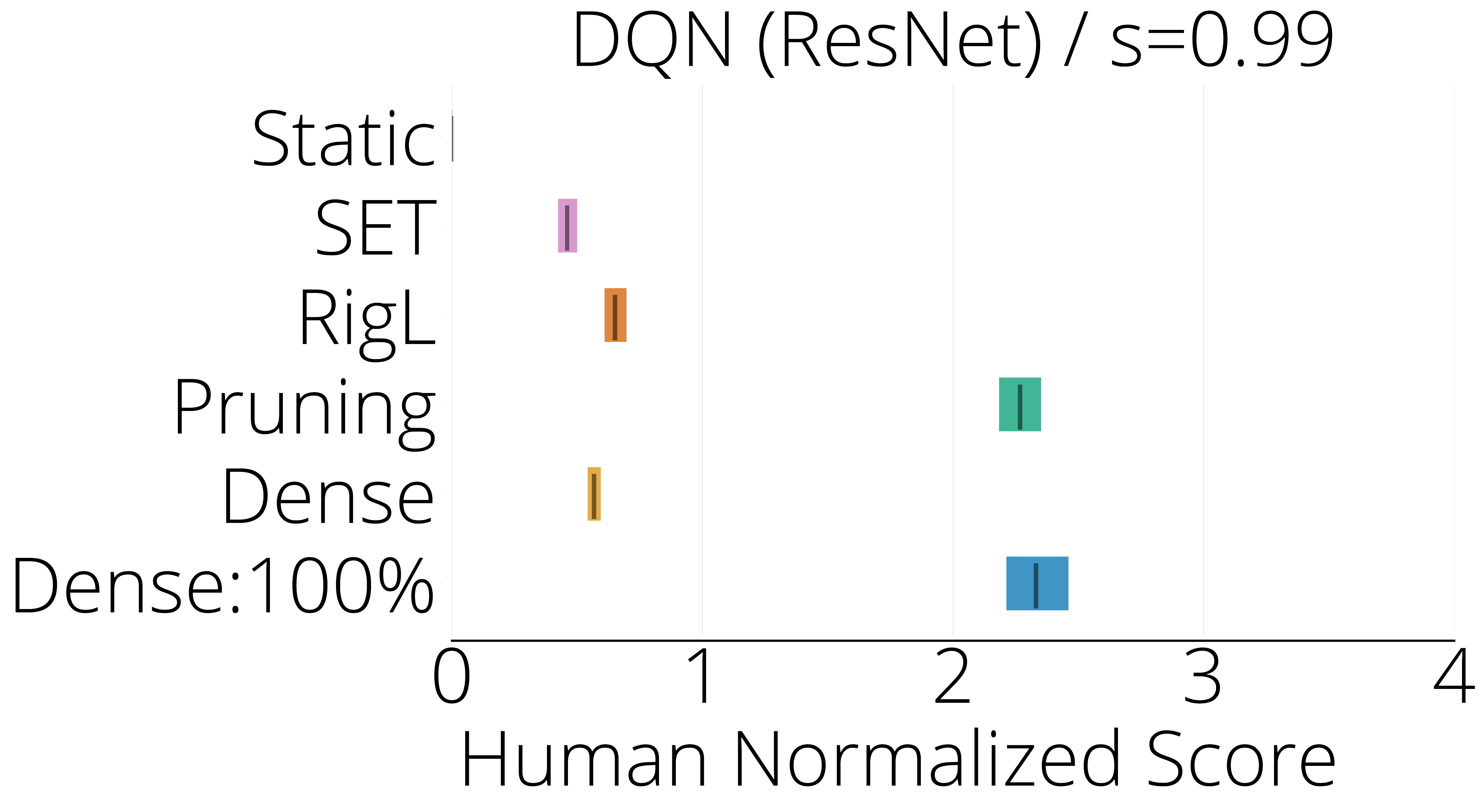}
    \caption{ResNet: IQM plots calculated over 15 Atari Games, with 9 seeds per game.}
    \label{fig:atari-resnet-iqm}
\end{figure*}

\paragraph{Atari DQN} Figure \ref{fig:atari-naturedqn-iqm} presents IQM plots calculated over 15 Atari games for the standard CNN network architecture and Figure \ref{fig:atari-resnet-iqm} presents IQM plots calculated over the same set of games for a ResNet architecture with an approximately equivalent number of parameters as the standard CNN ($\approx$4M).

\end{document}

%% file: tables/dqn_hyper_sweep.tex
\begin{tabular}{lcccc}
\toprule
Algorithm &  Weight Decay &  Update Interval &  Drop Fraction & Sparsity-aware init \\
\midrule
Dense & $1 \cdot 10^{-6}$  & - & -  & - \\
Pruning & $1 \cdot 10^{-6}$  & 1,000 & -  & True \\
Static & $1 \cdot 10^{-6}$  & - & -  & True \\
\gls{rigl} & $1 \cdot 10^{-6}$  & 1,000 & 0.5  & True \\
\gls{set} & $1 \cdot 10^{-6}$  & 1,000 & 0.5  & True \\
\bottomrule
\end{tabular}

%% file: tables/sac_hyper_sweep_80.tex
\begin{tabular}{lcccc}
\toprule
Algorithm &  Weight Decay &  Update Interval &  Drop Fraction & Sparsity-aware init \\
\midrule
Dense & $1 \cdot 10^{-4}$  & - & -  & - \\
Pruning & $1 \cdot 10^{-4}$  & 1,000 & -  & True \\
Static & $1 \cdot 10^{-4}$  & - & -  & False \\
\gls{rigl} & $1 \cdot 10^{-6}$  & 1,000 & 0.5  & True \\
\gls{set} & $1 \cdot 10^{-4}$  & 250 & 0.3  & True \\
\bottomrule
\end{tabular}

%% file: tables/ppo_hyper_sweep.tex
\begin{tabular}{lcccc}
\toprule
Algorithm &  Weight Decay &  Update Interval &  Drop Fraction & Sparsity-aware init \\
\midrule
Dense & $1 \cdot 10^{-6}$  & - & -  & - \\
Pruning & $1 \cdot 10^{-6}$  & 500 & -  & False \\
Static &  0  & - & -  & True \\
\gls{rigl} & $1 \cdot 10^{-4}$  & 250 & 0.3  & True \\
\gls{set} & $1 \cdot 10^{-6}$  & 100 & 0  & True \\
\bottomrule
\end{tabular}

%% file: tables/dqn_atari_cnn_hypers.tex
\begin{tabular}{lccccc}
\toprule
Algorithm &  Weight Decay & Learning Rate & Update Interval &  Drop Fraction & Sparsity-aware init \\
\midrule
Dense & 0 & $2.5 \cdot 10^{-4}$ & - & -  & - \\
Pruning & 0& $2.5 \cdot 10^{-4}$ &  5,000 & -  & False \\
Static & 0 & $2.5 \cdot 10^{-4}$ &  - & -  & True \\
\gls{rigl} & 0 & $2.5 \cdot 10^{-4}$ &  5,000 & 0.3  & True \\
\gls{set} & 0 & $2.5 \cdot 10^{-4}$ & 5,000 & 0.3  & True \\
\bottomrule
\end{tabular}

%% file: tables/dqn_atari_resnet_hypers.tex
\begin{tabular}{lccccc}
\toprule
Algorithm &  Weight Decay & Learning Rate & Update Interval &  Drop Fraction & Sparsity-aware init \\
\midrule
Dense & $1 \cdot 10^{-5}$ & $1 \cdot 10^{-4}$ &  - & -  & - \\
Pruning & $1 \cdot 10^{-5}$& $1 \cdot 10^{-4}$ &  5,000 & -  & False \\
Static & $1 \cdot 10^{-5}$ & $1 \cdot 10^{-4}$ &  - & -  & True \\
\gls{rigl} & $1 \cdot 10^{-5}$& $1 \cdot 10^{-4}$ &  5,000 & 0.3  & True \\
\gls{set} & $1 \cdot 10^{-5}$ & $1 \cdot 10^{-4}$ &  5,000 & 0.3  & True \\
\bottomrule
\end{tabular}

%% file: tables/dqn_hypers.tex
\begin{tabular}{l l| l }
\toprule
\multicolumn{2}{l|}{Parameter} &  Value\\
\midrule
\multicolumn{2}{l|}{\it{Shared}}& \\
& optimizer &Adam \citep{adam-kingma}\\
& discount ($\gamma$) &  0.99\\
& nonlinearity & ReLU\\
& target smoothing coefficient ($\tau$)& 1.0 \\
& gradient steps per training step & 1\\
& exploration policy & epsilon greedy\\
& epsilon decay period (env steps) & $2.5 \cdot 10^{4}$ \\
\midrule
\multicolumn{2}{l|}{\it{Classic Control}}& \\
& replay buffer size & $10^5$\\
& learning rate & $1 \cdot 10^{-3}$\\
& initial collect steps & 1,000\\
& target update interval & 100\\
& reward scale factor & 1.0\\
& gradient steps every k env steps, k = & 1\\
& final epsilon & 0.1\\
& eval epsilon & 0.1\\
& number of samples per minibatch & 128\\
& network type & MLP\\
& number of hidden dense layers & 2\\
& number of hidden units per layer & 512\\
\midrule
\multicolumn{2}{l|}{\it{Atari}}& \\
& replay buffer size & $10^6$\\
& initial collect steps & 20,000\\
& target update interval & 8000\\
& reward scale factor & 1.0\\
& gradient steps every k env steps, k = & 4\\
& final epsilon & 0.01\\
& eval epsilon & 0.001\\
& number of samples per minibatch & 32\\
& network type & CNN or ResNet\\
\bottomrule
\end{tabular}

%% file: tables/dqn_network_architectures.tex
\begin{tabular}{l l| l }
\toprule
\multicolumn{2}{l|}{Parameter} &  Value\\
\midrule
\multicolumn{2}{l|}{\it{CNN}}& \\
& learning rate & $2.5 \cdot 10^{-4}$ \\
& Adam optimizer, epsilon &  $1 \cdot 10^{-8}$\\
\multicolumn{2}{l|}{\it{CNN Architecture}}& \\
& number of hidden CNN layers & 3\\
& number of hidden dense layers & 1\\
& number of hidden units per dense layer & 512\\
\multicolumn{2}{l|}{\it{CNN params per layer (filters, kernel, stride)}}& \\
& layer 1 (filters, kernel, stride) & 32, 8, 4\\
& layer 2 (filters, kernel, stride) & 64, 4, 2\\
& layer 3 (filters, kernel, stride) & 64, 3, 1\\
\midrule
\multicolumn{2}{l|}{\it{ResNet}}& \\
& learning rate & $1 \cdot 10^{-4}$ \\
& Adam optimizer, epsilon &  $3.125 \cdot 10^{-4}$\\
\multicolumn{2}{l|}{\it{ResNet Architecture}}& \\
& number of stacks & 3\\
& number of hidden dense layers & 1\\
& number of hidden units per dense layer & 512\\
& use batch norm & False \\
\multicolumn{2}{l|}{\it{ResNet stack layers}}& \\
& num CNN layers & 1 \\
& num max pooling layers & 1 \\
& num residual-CNN layers & 2 \\
\multicolumn{2}{l|}{\it{ResNet params per layer (filters, kernel, stride)}}& \\
& stack 1 (filters, kernel, stride) & 32, 3, 1\\
& stack 2 (filters, kernel, stride) & 64, 3, 1\\
& stack 3 (filters, kernel, stride) & 64, 3, 1\\
\bottomrule
\end{tabular}

%% file: tables/sac_hypers.tex
\begin{tabular}{l l| l }
\toprule
\multicolumn{2}{l|}{Parameter} &  Value\\
\midrule
& optimizer &Adam \citep{adam-kingma}\\
& learning rate & $3 \cdot 10^{-4}$\\
& discount ($\gamma$) &  0.99\\
& replay buffer size & $10^6$\\
& number of hidden layers (all networks) & 2\\
& number of hidden units per layer & 256\\
& number of samples per minibatch & 256\\
& nonlinearity & ReLU\\
& target smoothing coefficient ($\tau$)& 0.005\\
& target update interval & 1\\
& train every k env steps, k = & 1\\
& gradient steps per training step = & 1\\
\midrule
\multicolumn{2}{l|}{\it{Hopper, Walker, Humanoid}}& \\
& initial collect steps & 1,000\\
\midrule
\multicolumn{2}{l|}{\it{HalfCheetah, Ant}}& \\
& initial collect steps & 10,000\\
\bottomrule
\end{tabular}

%% file: tables/ppo_hypers.tex
\begin{tabular}{l l| l }
\toprule
\multicolumn{2}{l|}{Parameter} &  Value\\
\midrule
& optimizer &Adam \citep{adam-kingma}\\
& learning rate & $3 \cdot 10^{-4}$\\
& discount ($\gamma$) &  0.99\\
& shared / separate networks & separate\\ 
& number of hidden layers (all networks) & 2\\
& number of hidden units per layer & 64\\
& collect sequence length (batch size) & 2048\\
& minibatch size & 64\\
& num epochs & 10\\
& importance ratio clipping & 0.2\\
& use GAE \citep{gae-iclr-2016} & True\\
& $\lambda$ (GAE) & 0.95\\
& entropy regularization  & 0\\
& value loss coeff & 0.5\\
& gradient clipping & 0.5\\
\bottomrule
\end{tabular}